% PT.tex (RMCG20140803)

\let\dvips=\relax       %% uncomment if using dvips
% RCstyle.tex (20141204)

\catcode`\^^c2=13
\ifx\files\undefined \input index \fi

\catcode`\@=11
\let\utf=\relax
\let\utf@Ch=\relax
\let\stringate=\relax
\let\dohigh=\relax
\let\doaccents=\relax
\let\dosymbols=\relax
\let\deactivate=\relax
\def\stringaccents{\def\'{\string\'}\def\~{\string\~}%
 \def\"{\string\"}\def\`{\string\`}\def\^{\string\^}}
\catcode`\@=12

\files

% Fonts

\ifx\loadfont\undefined \input fonts \fi
\xiifonts \xiititles \rm

\font\sfit=cmssi12

\catcode`\@=11

\font\xiibb=msbm10 at12pt
\font\ixbb=msbm9
\font\vibb=msbm6
\newfam\bbfam % fam 8
\textfont\bbfam=\xiibb \scriptfont\bbfam=\ixbb
\scriptscriptfont\bbfam=\vibb
\def\bb{\fam\bbfam\xiibb}
%\mathchardef\beth="0869
\mathchardef\subsetneq="3828

\font\xiifrak=eufm10 at12pt
\font\ixfrak=eufm9
\newfam\frakfam \textfont\frakfam=\xiifrak \scriptfont\frakfam=\ixfrak
\def\frak{\fam\frakfam\xiifrak}

\mathcode`?="603F % ? works as punctuation
\def\ifmath$#1${\relax\ifmmode #1\else$#1$\fi}
\def\QED{\ifmath$\diamond$}

% Layout

\ifcase\pdfoutput
 
\else
 \pdfcompresslevel=9
 \pdfdecimaldigits=4
 \pdfhorigin=1truein
 \pdfvorigin=1truein
 \pdfpageheight=297truemm
 \pdfpagewidth=210truemm
\fi
\hsize=15.92truecm \vsize=24.62truecm % DIN A4
\advance\vsize -30pt

\def\twodigits#1{\ifnum #1<10 0\fi \number#1}

\let\version=\todayiso
\def\Folio{\ifnum\pageno<0
 \uppercase\expandafter{\romannumeral-\pageno}\else\number\pageno\fi}

\headline={\vrule height10.2pt depth4.2pt width0pt
 {\xiitt www.ramoncasares.com}\quad{\xiirm\version}\hfil
 \quad{\xiibf\jobname\quad\Folio}}%
\def\makeheadline{\vbox to 30pt{\colorblack\line{\the\headline}%
  \kern 1pt \hrule height 1pt\vfil\endcolor}\nointerlineskip}
\nopagenumbers

% Sectioning

\newcount\secno
\newcount\ssecno
\newcount\thno
\newcount\parno
\let\presec=\empty
\def\presec{\S} % comment out for an empty one

\parskip=0pt plus 1pt
\newdimen\oldparindent \oldparindent=20pt \parindent=0pt
\def\hang{\hangindent\oldparindent}

\def\numberedpars{\global\advance\parno1 %\dest
 \noindent\hbox to\oldparindent{{\xiiscriptsy\char123
  \xiiscriptrm\number\parno}\hfil$\cdot$\hfil}\ignorespaces}

\outer\def\section#1{\vskip0pt plus 90pt\penalty-500\vskip0pt plus-90pt
 \everypar{}\advance\secno1
 \ssecno=0 \thno=0 \parno=0
 \goodbreak\vskip 2pc plus 1pc
 \def\secid{\presec\the\secno}
 \noindent{\fonttwo\secid\ #1\dest\toc1{#1}\lbl{#1}{\secid}}\par
 \everypar{\numberedpars}}
\outer\def\xsection#1{\vskip0pt plus 90pt\penalty-500\vskip0pt plus-90pt
 \everypar{}\parno=0
 \goodbreak\vskip 2pc plus 1pc
 \def\secid{}
 \noindent{\fonttwo#1\dest\toc1{#1}\lbl{#1}{}}\par
 \everypar{\numberedpars}}

%\outer\def\annex#1{\def\presec##1##2{#1}} % no longer used

\outer\def\subsection#1{\vskip0pt plus 60pt\penalty-250\vskip0pt plus-60pt
 \everypar{}\advance\ssecno1 \par
 \ifnum\parno=0 \vskip 0.5pc plus 6pt \else\vskip 1pc plus6pt\fi
 \thno=0 \parno=0
 \def\secid{\presec\the\secno.\the\ssecno}%
 \noindent{\fontthree\secid\ #1\dest\toc2{#1}\lbl{#1}{\secid}}\par
 \everypar{\numberedpars}}
\outer\def\clause#1{\vskip0pt plus 30pt\penalty-150\vskip0pt plus-30pt
 \everypar{}\medskip\hang\noindent
 \advance\thno1 \def\secid{\presec\the\secno.\the\ssecno.\the\thno}%
 {\sc\secid\ #1}\quad\ignorespaces}
\outer\def\comment#1{\everypar{}\par
 \hang\noindent{\sc #1}\quad\ignorespaces}

\def\note#1{\ifvmode
 \vtop to0pt{\vss\rlap{\color{White}\xiiscriptrm #1\endcolor}\kern-2pt\null}%
\else
 \vadjust{\vtop to0pt{\vss
 \rlap{\color{White}\xiiscriptrm #1\endcolor}\kern-2pt\null}}\fi}
\def\label#1{\note{#1}\dest\lbl{#1}{\secid}\ignorespaces}

% Openings

\def\title#1{\def\titledoc{#1}}
\def\author#1{\def\authordoc{#1}}
\def\contact#1{\def\contactdoc{#1}}
\def\keywords#1{\def\keywordsdoc{#1}}
\def\subject#1{\def\subjectdoc{#1}}

\def\beginnote{\insert\footins\bgroup\strut\colorblack}
\def\endnote{\endcolor\egroup}

\def\maketitle{\par
 \begingroup
 \def\'##1{\if##1o\string^^f3\else ##1\fi}% hack \'o -> f3
 \ifcase\pdfoutput
  \ifx\dvips\undefined
  \else
   \special{ps: [
    /Title (\titledoc) /Author (\authordoc)
    /Keywords (\keywordsdoc) /Subject (\subjectdoc)
    /DOCINFO pdfmark }%
  \fi
 \else
  \pdfinfo{/Title (\titledoc) /Author (\authordoc)
    /Keywords (\keywordsdoc) /Subject (\subjectdoc)}\fi
 \endgroup
 \null\vskip 3pc plus 4pc minus 1pc
 \def\secid{}\dest\toc0{\titledoc}\lbl{\titledoc}{}
 \centerline{\fontzero\titledoc}
 \vskip 1pc plus 1pc
 \centerline{\fontone\authordoc}
 \ifx\contactdoc\undefined\else
  \vskip 6pt minus 3pt
  \centerline{\rm\contactdoc}\fi
 \vskip 2pc plus 1pc minus 1pc\relax}

\def\beginabstract{\begingroup \parindent=55pt\narrower \sl\parindent=20pt\noindent}
\def\endabstract{\par \ifx\keywordsdoc\undefined\else
 \smallskip\sfit\setbox0=\hbox{Keywords:\quad}\hangindent\wd0
 \noindent\box0\keywordsdoc\par\fi \smallskip\endgroup}

% References

\def\cite #1 (#2){\ref{#1#2}}

%% \cite Post (1944) -> Post (1944)
%% \cite *Post (1944) -> Post 1944
\def\chreference #1 (#2){\everypar{}\par
 \vskip0pt plus 2\baselineskip\penalty-43
 \vskip0pt plus-2\baselineskip
 \noindent\hangindent20pt\relax
 #1\ (#2)\dest\lbl{#1#2}{#1\ (#2)}\lbl{*#1#2}{#1\ #2}}

\def\brreference #1 (#2){\everypar{}\par
 \vskip0pt plus 2\baselineskip\penalty-43
 \vskip0pt plus-2\baselineskip
 \noindent\hangindent20pt\relax
 [#1 #2]\dest\lbl{#1#2}{[#1 #2]}}

\let\reference=\chreference

\def\book#1{{\it#1\/}}
\def\periodical#1{{\it#1\/}}
\def\ISBN{{\sc isbn: }}
\def\DOI#1{{\sc doi: }\URL{#1}<http://dx.doi.org/#1>}

% Table of contents

\def\tocline#1#2#3#4#5{\par \ifcase#1
 \toclinezero{#2}{#3}{#4}{#5}\or
 \toclineone{#2}{#3}{#4}{#5}\or
 \toclinetwo{#2}{#3}{#4}{#5}\else
 \toclinethree{#2}{#3}{#4}{#5}\fi}

\def\gobblefour#1#2#3#4{\relax}
\def\leaderfill{\leaders\hbox to\baselineskip{\hss.\hss}\hfill}
\def\toclinezero#1#2#3#4{\centerline{\goto{\fonttwo #1}{#2}}\bigskip}
\def\toclineone#1#2#3#4{\def\3{#3}%
 \line{\fontthree \ifx\3\empty\else #3 \fi#1\leaderfill \goto{#4}{#2}}}
\def\toclinetwo#1#2#3#4{\line{\qquad #3 #1\leaderfill \goto{#4}{#2}}}
\let\toclinethree=\gobblefour

% Differences

% The TeX Book Exercise 14.28
\def\strutdepth{\dp\strutbox}
\def\marginalstar{\strut\vadjust{\kern-\strutdepth\specialstar}}
\def\specialstar{\vtop to \strutdepth{
 \baselineskip\strutdepth
 \vss\llap{\truecolor{Red}\rm*\endcolor\quad}\null}}

\def\new/{\truecolor{Red}}
\def\wen/{\endcolor\ifhmode\marginalstar\fi}

\def\uncatcodespecials{\def\do##1{\catcode`##1=12 }\dospecials}
\def\del/{\begingroup\uncatcodespecials
 \ifvmode \let\next=\DELv \else \let\next=\DELh \fi\next}
{\catcode`\|=0 \catcode`\\=12
 |long|gdef|DELv#1\led/{|endgroup}
 |long|gdef|DELh#1\led/{|marginalstar|endgroup|ignorespaces}}
\def\led/{\errmessage{Error! Unnested led.}}

% Misc

\def\newline{\ifvmode\null\else\null\hfil\break\fi}
\let\\=\newline

\def\needspace#1{\vskip0pt plus #1\penalty-250\vskip0pt plus -#1\relax}

\def\definition#1{{\sl #1\/}}
\def\latin#1{{\it #1\/}}

\def\URL{\leavevmode\begingroup\catcode`\#=12\catcode`\_=12\relax\@RL}
\def\@RL#1<#2>{\def\1{#1}\ifx\1\empty\def\2{#2}\else\def\2{#1}\fi
\ifcase\pdfoutput
 \ifx\dvips\undefined
  {\color{Red}\2\endcolor}\else
  {\color{Red}\2\endcolor}%
\fi
\else
 \pdfstartlink attr{/Border [0 0 0]}
  user{/Subtype /Link /A << /Type /Action
  /S /URI /URI (#2) >>}{\color{Red}\2\endcolor}\pdfendlink
\fi\endgroup\relax}

\catcode`\@=12

\def\version{20160902}  %% uncomment to fix version

\darkcolors              %% comment for black & white
\def\note#1{\relax}      %% comment to see destinations

\hyphenation{meta-problem}
\def\T#1{\buildrel\scriptscriptstyle\ast\over#1} % rel in a Turing universe
\def\numeq#1#2{\buildrel\scriptscriptstyle#1\over#2}
\def\TM#1<#2>{\def\1{#1}\ifx\empty\1\else{\cal #1}\fi\langle{\frak #2}\rangle}
\def\case{\par\everypar{}\noindent\hskip\oldparindent\hangindent2\oldparindent}
\def\into#1{\setbox0=\hbox{$\scriptstyle#1$}\dimen0=\wd0\advance\dimen0 12pt
 \mathop{\hbox to\dimen0{\rightarrowfill}}\limits^{\box0}}
\def\generic#1{\{\cdot\,#1\,\cdot\}}

\ifx\Cap\undefined
\def\Cap{\mathbin{\ooalign{\hfil$\scriptstyle\cap$\hfil\cr$\cap$\cr}}}
\fi

%%%%%%%%%%%%%%%%%%%%%%%%%%%%%%%%%%%%%%%%%%%%%%%%

\title{Problem Theory}
\author{Ram\'on Casares}
\contact{{\sc orcid:}
\URL 0000-0003-4973-3128<http://orcid.org/0000-0003-4973-3128>}
\subject{Cognition, Computing, Mathematics}

\beginnote
This is
{\tt\URL arXiv:1412.1044<http://arxiv.org/abs/1412.1044>}
version {\tt\version},
and it is licensed as {\tt cc-by}.\\
Any comments on it to
{\tt\URL papa@ramoncasares.com<mailto:papa@ramoncasares.com>}
are welcome.
\endnote

\maketitle

% \beginabstract % descriptive short
% We define a problem theory from first principles.
% We investigate the objects of this theory:
% problems, resolutions, and solutions.
% We relate problem theory with
% set theory and with computing theory.
% We find taxonomies for resolutions and for problems.
% We build a hierarchy of resolvers:
% mechanism, adapter, perceiver, learner, and subject.
% We show that the problem theory is complete, that is,
% that there are just three ways to resolve any problem:
% routine, trial, and analogy.
% Finally, we propose a thesis:
% We are Turing complete subjects because
% we are the result of an evolution of resolvers
% of the survival problem.
% \endabstract

\beginabstract % informative long
%
% Motivation & Problem Statement
%
The Turing machine,
 as it was presented by Turing himself,
models the calculations done by a person.
This means that we can compute
whatever any Turing machine can compute,
and therefore we are Turing complete.
The question addressed here is why,
{\it Why are we Turing complete?}
%
% Approach
%
Being Turing complete also means that
somehow our brain implements the function that
a universal Turing machine implements.
The point is that evolution achieved
Turing completeness, and then
the explanation should be evolutionary,
but our explanation is mathematical.
The trick is to introduce a
mathematical theory of problems,
under the basic assumption that
solving more problems provides more
survival opportunities.
%
% Results
%
So we build a problem theory by fusing
set and computing theories.
Then we construct a series of resolvers,
where each resolver is defined
by its computing capacity,
that exhibits the following property:
all problems solved by a resolver
are also solved by the next resolver in the series
if certain condition is satisfied.
The last of the conditions is to be Turing complete.
%
% Conclusions
%
This series defines a resolvers hierarchy that
could be seen as a framework for
the evolution of cognition.
Then the answer to our question would be:
{\it to solve most problems}.
By the way, the problem theory
 defines adaptation, perception, and learning,
 and it shows that
 there are just three ways to resolve any problem:
 routine, trial, and analogy.
And, most importantly, this theory demonstrates
how problems can be used
to found mathematics and computing on biology.
\endabstract

\vfill\break
\null\vfill
\input auxiliar.toc
\vfill\break

\section{Introduction}

\begingroup\medskip\everypar{}\it
Devoid of problems, thinking is useless.
\smallskip\endgroup

\comment{Warning} This paper does not explain
how to solve, nor how to resolve, any problem.

\subsection{Object}

The object of this paper is to present
a mathematical theory of problems.
The resulting problem theory
provides meaning to set theory
and to computing theory.

Problems are nearly everywhere.
We can say that
 mathematics is all about mathematical problems,
but also that
 physics is all about physical problems, and
 philosophy is all about philosophical problems.
I said nearly because there are not problems in a river;
 a river just flows.
So, where are problems?

This problem theory gives an answer:
There are problems where there is freedom.
Determinists will surely object,
but they should note that if there were only uncertainty,
and not the possibility of doing otherwise,
then problem resolving would be purposeless and absurd.
Nevertheless, in this theory
freedom cannot exist by itself,
but freedom is always limited by a condition and
both together, freedom and a condition, are a problem.
In fact, the resolution of any problem is
the process of spending all of its freedom
while still satisfying the condition.
So resolving is fighting freedom away.
And, if people fight for freedom,
it is because we want problems;
in fact, not having any problem is boring.
But I would say more,
we are devices exquisitely selected
to resolve problems,
because surviving is literally the
problem of being dead or alive:
``To be, or not to be---that is the question.''

I am digressing, sorry!
The point is that problems are related to sets
at the very bottom: for each problem there is a condition
that determines if anything is a solution to it or not,
so for each problem there is a set,
the set of its solutions,
and the condition is its characteristic function.
This means that problems and sets are just two names for
the same thing.
So problem theory, being just a rewording of set theory,
would be a better foundation for mathematics than set theory,
because problems are more related to thinking than sets are.

We have just seen how problems and solutions fit with sets,
but we have seen nothing about resolutions, that is,
the ways to go from a problem to its solutions.
It is a fact that computing is helping us in
resolving many problems. Perhaps too many:
How our modern society would subsist without computers?
I am digressing again, sorry! The right question is:
What is the relation between problem resolving and computing?

Computing is the mechanical manipulation of
strings of symbols.
Mechanical in the sense that the manipulations
do not take into account the meaning of the symbols,
but they just obey blindly a finite set of well-defined rules.
Being meaningless, what could be the purpose of computing?
Historically, computing resulted from two
apparently different pursuits:
the foundation of mathematics, and
the enhancement of calculating machines.
The second,
the development of mechanical aids to calculation,
is easier to understand.
When we learn the algorithm for division
we readily appreciate that those fixed rules
can be better applied by a machine than by a person.
This explains why an arithmetic calculator
comes handy when resolving a problem that
requires performing a numerical division.
And it could also help us to understand
why computation was seen as
the ideal for mathematical rigor, and then
how computing relates to the foundations of mathematics.

But, again, what is the purpose of
 mathematical formalization?
Is it true that a complete formalization of
mathematics would render it meaningless?
What would be the use of something meaningless?
And again, the arithmetic calculator,
dividing for us, answers the three questions:
formalization prevents mistakes and
assures that nothing has been taken for granted,
 and,
 while it is literally meaningless,
it is not useless if
the formalism helps us in resolving problems.
Though pending on an if,
formalism is not yet lost.
This paper cuts that Gordian knot by
showing that problem resolving is computing.

In fact, that `resolving is computing' comes from
the founding paper of computing.
\cite Turing (1936) proved that the
{\it Entscheidungsproblem},
which is the German word for `decision problem',
is unsolvable,
because it cannot be solved by any Turing machine.
For this proof to be valid,
`solved by a Turing machine' has to be equal to `solved',
and therefore `resolved by computing' has to be redundant.

Summarizing,
a problem is a set, and resolving is computing.
This is how this problem theory relates
to set and computing theories at the highest level
of abstraction. For a more detailed view
you should continue reading this paper.

\subsection{Contents}

The object of this paper is to introduce
a mathematical theory of problems.
Because our approach is minimalist,
aiming to keep only what is essential,
we will define a problem theory from first principles.
Section~\ref{Theory} contains
this problem theory, including its
eight concepts: problem, with freedom and condition;
resolution, with routine, trial and analogy; and solution.
Some care is advisable
to distinguish `solution' from `resolution',
because while they are usually considered 
synonyms, they are very distinct concepts in this theory:
a solution is a state,
and a resolution is a transition.
Then that `a problem is resolved unsolvable' achieves
a very precise meaning.

Section \ref{Sets} translates the problem theory to set theory.
Subsection~\ref{Problems} defines
what a problem is, and what is the set of its solutions
is defined in Subsection~\ref{Solutions}.
Then, in Subsection~\ref{Routines and Trials},
we develop the first two ways to resolve a problem,
by routine and by trial,
while we devote Subsection~\ref{Analogies} to the third way,
by analogy.
The conclusion of these two subsections is that
there is a general form that includes the three forms.
Then we observe that looking for a resolution to
a problem is also a problem, the metaproblem,
so Subsection~\ref{Metaproblems} deals with metaproblems.
The last subsection of this section,
 Subsection~\ref{Resolution Typology},
shows that there is only one level
of problem meta-ness and that
 there are five types of resolution.

The next section, Section~\ref{Computers},
is about computing.
In Subsection~\ref{Turing Machine} we present the
Turing machine, concluding that all computing
is inside countable sets.
In Subsection~\ref{Turing Completeness} we deal with
universal computers and Turing completeness.
Then, in Subsection~\ref{Turing's Thesis}, we explain that
Turing's thesis implies that everything is an expression,
that resolving is computing, and
that all problem sets are countable.
In Subsection~\ref{Full Resolution Machine},
we introduce the full resolution machine,
and we show some equivalences
between problem theory and computing theory.
In the last subsection of this section,
 Subsection~\ref{Problem Topology},
we show that there are five types of problem.

Section~\ref{Resolvers} is about resolvers,
that is, devices that resolve problems.
In the first subsection, \ref{Semantics and Syntax},
we present the practical scenario,
where functions are not solutions,
so we distinguish the semantics of solutions
 from the syntax of functions,
and we define the range of a resolver as
 the set of problems that the resolver solves,
and the power of a resolver as
 the set of problems that the resolver resolves.
Then we construct a series of five resolvers:
\begingroup\everypar{}\parindent=20pt\parskip=0pt
 \item{$\circ$} Mechanism, Subsection~\ref{Mechanism},
  is any device that implements a semantic unconditional computation.
  We show that mechanisms can resolve problems by routine.
 \item{$\circ$} Adapter, in~\ref{Adapter},
  is any device that implements a semantic conditional computation.
  We show that adapters can resolve problems by trial.
 \item{$\circ$} Perceiver, in \ref{Perceiver},
  is any device that implements a semantic functional computation
  or a syntactic unconditional computation.
  We show that perceivers can resolve problems by analogy
  and metaproblems by routine.
 \item{$\circ$} Learner, in~\ref{Learner},
  is any device that
   implements a syntactic conditional computation.
  We show that learners can resolve metaproblems by trial.
 \item{$\circ$} Subject, in~\ref{Subject},
  is any device that implements
  a syntactic functional computation.
  We show that subjects can resolve metaproblems by analogy.
\par\noindent\endgroup
In addition, we show that
the range and power of each resolver in the series includes
the range and power of the previous one,
provided that a specific condition is satisfied.
The last of these conditions
requires the subject to be Turing complete.
So, in the last subsection of this section, \ref{Resolvers Hierarchy},
we summarize the findings of the section:
we show that
 there is a hierarchy of five types of resolver,
and that
 the problem theory is complete.
The theory is complete because
Turing completeness is the maximum computing capacity,
and this means that there are
exactly three ways to resolve any problem:
routine, trial, and analogy.
Finally,
we argue that we are the Turing complete subjects
that have resulted from an evolution
of resolvers of the survival problem.

The paper finishes with some conclusions,
in Section~\ref{Conclusion}.
In the first subsection, \ref{Purpose},
we explain how problem theory provides purpose and meaning
to set theory and to computing theory.
In the next subsection, \ref{Countability},
we argue that countableness is the golden mean
that keeps paradoxes under control.
And in the last subsection, \ref{Intuition},
we explore what would be the implications of
non-computable ways of resolving, as intuition.

\section{Theory}

\subsection{Problem}

Every \definition{problem} is made up of freedom and of a condition.
There have to be possibilities and \definition{freedom} to choose among them, because
if there is only necessity and fatality, then there is neither a
problem nor is there a decision to make. The different possible options
could work, or not, as solutions to the problem, so that in every
problem a certain \definition{condition} that will determine if an option is valid or
not as a solution to the problem must exist.
$$
\centerline{$\hbox{\rm Problem}
 \left\lbrace\vcenter{\hbox{Freedom}\hbox{Condition}}\right.$}
$$

\subsection{Solution}

A fundamental distinction that we must make is between the solution and
the resolution of a problem. Resolving is to searching as solving is to
finding, and please note that one can search for something that does not
exist.
$$\vbox{\halign{\strut\hss#& \hss# $\cdot$ \hss& #\hss\crcr
 Resolving& &Searching\cr
 Solving& &Finding\cr
}}$$
Thus, \definition{resolution} is the process that attempts to reach the
solutions to the problem, while a \definition{solution} of the problem is any use
of freedom that satisfies the condition.
In the state-transition jargon:
a problem is a state of ignorance,
a solution is a state of satisfaction, and
a resolution is a transition from uncertainty to certainty.
$$
 \hbox{\rm Problem}
 \into{\hbox{\quad\rm Resolution\quad}}
 \hbox{\rm Solution}
$$

We can explain this with another analogy. The problem is defined by the
tension that exists between two opposites: freedom, free from any
limits, and the condition, which is pure limit. This tension is the
cause of the resolution process. But once the condition is fulfilled and
freedom is exhausted, the solution annihilates the problem. The
resolution is, then, a process of annihilation that eliminates freedom
as well as the condition of the problem, in order to produce the
solution.
$$
\hbox{$\underbrace{\vcenter{
  \halign{\strut\hfil\rm#\crcr Freedom\cr Condition\cr}}
    }_{\hbox{\rm\strut Problem}} \, \Bigr\rbrace
 \into{\hbox{\quad\rm Resolution\quad}}
 \hbox{\rm Solution}$}
$$

A mathematical example may also be useful in order to
distinguish resolution from solution. In a problem of
arithmetical calculation, the solution is a number and
the resolution is an algorithm such as the algorithm for division,
for example.

\subsection{Resolution}

There are three ways to resolve a problem:
 routine, trial, and analogy.
$$\hbox{Resolution}\left\lbrace\vcenter{
 \hbox{Routine}\hbox{Trial}\hbox{Analogy}}\right.$$

To resolve a problem by \definition{routine},
that is, by knowing or without reasoning,
it is necessary to know the solutions, and
it is necessary to know that they solve that problem.

If the solutions to a problem are not known,
but it is known a set of possible solutions,
then we can use a trial and error procedure,
that is, we can try the possible solutions.
To resolve by \definition{trial} is
to test each possible solution
until the set of possible solutions is exhausted or
      a halting condition is met.
There are two tasks when we try:
to test if a particular possibility satisfies the problem condition,
and to govern the process determining
the order of the tests and when to halt.
There are several ways to govern the process, that is,
there is some freedom in governing the trial, and so,
if we also put a condition on it, for example a temporal milestone,
then governing is a problem. And
there are three ways to resolve a problem (\latin{da capo}).

By \definition{analogy} we mean to transform a problem into a different one,
called question, which is usually composed of several subproblems.
This works well if the subproblems are easier to resolve than the
original problem.
There are usually several ways to transform any problem
(there is freedom), but only those transformations
that result in questions that can be resolved are valid
(which is a condition),
so applying an analogy to a problem is a problem.
There are three ways to resolve the analogy, the question,
and each of its subproblems: routine, trial, and analogy (\latin{da capo}).
If we could translate a problem into an analogue question,
and we could find a solution to that question, called answer,
and we could perform the inverse translation on it,
then we would have found a solution to the original problem.
$$\vbox{\halign{\strut\hss#\hss& \hss#\hss& \hss#\hss\crcr
 Problem && Solution\cr
 $\downarrow$&&$\uparrow$\cr
 Question &$\longrightarrow$ & Answer\cr
 }}$$

\subsection{Eight Concepts}

Lastly we are ready to list the eight concepts of the problem theory.
They are: problem, with freedom and condition;
resolution, with routine, trial, and analogy;
and solution.
$$\hbox{Problem Theory}\left\lbrace\vcenter{
   \hbox{Problem $\left\lbrace\vcenter{\hbox{Freedom}\hbox{Condition}}\right.$}\null
   \hbox{Resolution $\left\lbrace\vcenter{\hbox{Routine}\hbox{Trial}\hbox{Analogy}}\right.$}
   \hbox{Solution}\kern6pt}\right.$$

\section{Sets}

\subsection{Problems}

\clause{Notation}
We will refer to the set of problems as $\bb P$.
We will refer to the set of resolutions as $\bb R$.
We will refer to the set of solutions as $\bb S$.

\comment{Definition}
A resolution takes a problem and returns
the set of the solutions to the problem.
Then resolutions are ${\bb R} = {\bb P} \to 2^{\bb S}$,
where $2^{\bb S}$ is the powerset,
 or the set of the subsets, of $\bb S$.

\clause{Notation}
$\top$ stands for `true', and $\bot$ for `false'.
We will refer to the set of these Boolean values as $\bb B$.
${\bb B} = \{\top, \bot\}$.

\comment{Comment}
$\top = \neg\bot$ and $\bot = \neg\top$. Also
$[P = \top] = P$ and $[P = \bot] = \neg P$.

\clause{Notation}\label{functionset}
Given $s \in S \subseteq {\bb S}$
and $f \in F \subseteq ({\bb S} \to {\bb S})$,
so $f: {\bb S} \to {\bb S}$ and $f(s) \in {\bb S}$,
we will use the following rewriting rules:\\
$f(S) = \{\, f(s) \mid s \in S \,\}$,
$F(s) = \{\, f(s) \mid f \in F \,\}$, and
$F(S) = \{\, f(s) \mid s \!\in\! S \times f \!\in\! F \,\}$.

\comment{Comment}
As $f(s) \in {\bb S}$, then
$f(S) \in 2^{\bb S}$, $F(s) \in 2^{\bb S}$, and $F(S) \in 2^{\bb S}$.

\comment{Proposition}
If $s \in S$ and $f \in F$, then
 $f(S) \subseteq F(S)$ and $F(s) \subseteq F(S)$.

\clause{Definition}
Problem $\pi$ is $x? P_\pi(x)$,
where $P_\pi$ is any predicate, or Boolean-valued function,
on ${\bb S}$;
so $P_\pi: {\bb S} \to {\bb B}$, where
$P_\pi(x)=\top$ means that $x$ is a solution of $\pi$, and
$P_\pi(x)=\bot$ means that $x$ is not a solution of $\pi$.

\comment{Comment}
A problem $\pi = x? P_\pi(x)$ is made up
 of freedom and of a condition,
as defined in Section~\ref{Theory}.
The condition is $P_\pi$, and
freedom is represented by the free variable $x$,
 which is free to take any value in $\bb S$, $x\in{\bb S}$.

\clause{Definition}\label{effectivelycalculable}
A function $^*\!f$ is effectively calculable
if there is a purely mechanical process
to find $^*\!f(s)$ for any $s$.

\comment{Comment}
This definition of effective calculability 
was stated by \cite Turing (1938), \S2.

\comment{Comment}
If the result of the calculation is finite,
then an effective calculation has to complete it.
If the result of the calculation is infinite,
then an effective calculation has to proceed forever
towards the result.

\comment{Notation}
We will refer to
the set of effectively calculable functions as $^*\bb F$.

\clause{Definition}\label{expressible}
A problem $\pi$ is expressible
if its condition $P_\pi$ is an effectively calculable
function.

\comment{Comment}
The result of a condition is in set ${\bb B} = \{\top, \bot\}$,
so it is always finite.
Therefore a problem is not expressible
if for some $x$ we cannot calculate
whether $x$ is a solution or not
in a finite time.

\clause{Definition}\label{conditionisomorphism}
The condition isomorphism 
is the natural isomorphism that relates
each problem $\pi$ with its condition $P_\pi$:
for each predicate $P$
 there is a problem, $x? P(x)$, and
for each problem, $\pi = x? P_\pi(x)$
 there is a predicate, $P_\pi$.
That is,
${\bb P} \Leftrightarrow ({\bb S} \to {\bb B}):
 x? P_\pi(x) \leftrightarrow P_\pi$.

\comment{Comment}
Using the condition isomorphism,
two problems are equal if they have the same condition,
that is, $\pi = \rho \Leftrightarrow P_\pi = P_\rho$.

\comment{Comment}
The condition isomorphism abstracts freedom away.

\clause{Theorem}\label{conditionequality}
The set of problems is the set of predicates,
that is, ${\bb P} = {\bb S} \to {\bb B}$.

\comment{Proof}
${\bb P} \cong {\bb S} \to {\bb B}$,
by the condition isomorphism, see \ref{conditionisomorphism},
and, abstracting freedom, ${\bb P} = {\bb S} \to {\bb B}$.
But freedom has to be abstracted away from mathematics
because freedom is free of form and
it cannot be counted nor measured.
{\QED}

\comment{Comment}
Although in mathematics we cannot deal with freedom,
it is an essential part of problems, see \ref{Problem}.
In any case,
what defines problem $\pi$ is its condition $P_\pi$.

\clause{Lemma}
The name of the free variable is not important,
it can be replaced: $x? P(x) = y? P(y)$.

\comment{Proof}
By the condition isomorphism, and \ref{conditionequality},
both problems, $x? P(x)$ and $y? P(y)$,
are equal, $x? P(x) = y? P(y)$,
because they have the same condition, $P$.
{\QED}

\comment{Comment}
This means that the rule of $\alpha$-conversion
stands for problem expressions.
See \cite Curry \& Feys (1958), Section~3D.

\clause{Definition}\label{problemcomposition}
Let $\pi$ and $\rho$ be two problems.
Then $\pi \land \rho = x? P_\pi(x) \land P_\rho(x)$,
 and $\pi \lor \rho = x? P_\pi(x) \lor P_\rho(x)$,
 and $\bar{\pi} = x? \neg P_\pi(x)$.

\comment{Comment}
In other words,
$P_{\pi\land\rho}(x) = P_\pi(x) \land P_\rho(x)$,
$P_{\pi\lor\rho}(x) = P_\pi(x) \lor P_\rho(x)$, and
$P_{\bar\pi}(x) = \neg P_\pi(x)$.

\comment{Comment}
This provides a way to compose, or decompose, problems.

\clause{Definition}\label{tautologycontradiction}
A problem $\tau$ is tautological if its condition is a tautology;
$P_\tau$ is a tautology, if $\forall x,\, P_\tau(x) = \top$.
A problem $\bar\tau$ is contradictory
 if its condition is a contradiction;
$P_{\bar{\tau}}$ is a contradiction,
 if $\forall x,\, P_{\bar{\tau}}(x) = \bot$.

\comment{Lemma}
Both $\tau$ and $\bar\tau$ are expressible.

\comment{Proof}
Because $P_\tau$ and $P_{\bar\tau}$ are effectively calculable,
 see \ref{expressible} and \ref{effectivelycalculable}.
{\QED}

\clause{Theorem}
$\left< {\bb P}, \lor, \land, \neg, \bar\tau, \tau \right>$
is a Boolean algebra,
 where $\bar\tau$ is the neutral for $\lor$,
 and $\tau$ is the neutral for $\land$.

\comment{Proof}
Because $P_\pi(x) \in {\bb B}$. In detail,
$\forall \pi,\rho,\sigma \in {\bb P}$:
\case 1o. $(\pi\lor\rho)\lor\sigma=
 x? P_{\pi\lor\rho}(x) \lor P_\sigma(x) =
 x? (P_\pi(x)\lor P_\rho(x)) \lor P_\sigma(x) =\hfil\break\null\hfill
 x? P_\pi(x) \lor (P_\rho(x)\lor P_\sigma(x)) =
 x? P_\pi(x) \lor P_{\rho\lor\sigma}(x) =
 \pi\lor(\rho\lor\sigma)$.
\case 1a. $(\pi\land\rho)\land\sigma=
 x? P_{\pi\land\rho}(x) \land P_\sigma(x) =
 x? (P_\pi(x)\land P_\rho(x)) \land P_\sigma(x) =\hfil\break\null\hfill
 x? P_\pi(x) \land (P_\rho(x)\land P_\sigma(x)) =
 x? P_\pi(x) \land P_{\rho\land\sigma}(x) =
 \pi\land(\rho\land\sigma)$.
\case 2o. $\pi\lor\rho = x? P_\pi(x) \lor P_\rho(x) =
 x? P_\rho(x) \lor P_\pi(x) =
 \rho\lor\pi$.
\case 2a. $\pi\land\rho = x? P_\pi(x) \land P_\rho(x) =
 x? P_\rho(x) \land P_\pi(x) =
 \rho\land\pi$.
\case 3o. $\pi \lor \bar\tau = x? P_\pi(x) \lor P_{\bar\tau}(x) =
 x? P_\pi(x) \lor \bot = x? P_\pi(x) = \pi$.
\case 3a. $\pi \land \tau = x? P_\pi(x) \land P_\tau(x) =
 x? P_\pi(x) \land \top = x? P_\pi(x) = \pi$.
\case 4o. $\pi \lor \bar\pi = x? P_\pi(x) \lor P_{\bar\pi}(x) =
 x? P_\pi(x) \lor \neg P_\pi(x) = x? \top = x? P_{\tau}(x) =
 \tau$.
\case 4a. $\pi \land \bar\pi = x? P_\pi(x) \land P_{\bar\pi}(x) =
 x? P_\pi(x) \land \neg P_\pi(x) = x? \bot = x? P_{\bar\tau}(x) =
 \bar\tau$.
\case 5o. $\pi\lor(\rho\land\sigma)=
 x? P_\pi(x)\lor P_{\rho\land\sigma}(x) =
 x? P_\pi(x)\lor (P_\rho(x) \land P_\sigma(x)) =\hfil\break\null\hfill
 x? (P_\pi(x)\lor P_\rho(x)) \land (P_\pi(x) \lor P_\sigma(x)) =
 x? P_{\pi\lor\rho}(x) \land P_{\pi\lor\sigma}(x) =
 (\pi\lor\rho)\land(\pi\lor\sigma)$.
\case 5a. $\pi\land(\rho\lor\sigma)=
 x? P_\pi(x)\land P_{\rho\lor\sigma}(x) =
 x? P_\pi(x)\land (P_\rho(x) \lor P_\sigma(x)) =\hfil\break\null\hfill
 x? (P_\pi(x)\land P_\rho(x)) \lor (P_\pi(x) \land P_\sigma(x)) =
 x? P_{\pi\land\rho}(x) \lor P_{\pi\land\sigma}(x) =
 (\pi\land\rho)\lor(\pi\land\sigma)$.
\case {\QED}

\subsection{Solutions}

\clause{Theorem}\label{everythingisin}
Everything is in $\bb S$.
In other words, $\bb S$ is the set of everything.

\comment{Proof}
Anything, let us call it $s$, is a solution to problem
$x? [x = s]$,
because equality is reflexive, and therefore
everything satisfies the condition of being equal to itself.
{\QED}

\comment{Comment}
Freedom is complete,
because $x$ is free to take any value;
$x\in{\bb S}$ is not a restriction.
And $P_\pi: {\bb S} \to {\bb B}$
is a predicate on everything.

\comment{Comment}
Some paradoxes derive from this theorem,
see \ref{SsubsetP}.
For a constructive vision of ${\bb S}$,
see Section~\ref{Resolvers}.
See also Subsection~\ref{Countability}.

\comment{Corollary}
${\bb P} \subset \bb S$ and ${\bb R} \subset {\bb S}$.
Even ${\bb B} \subset {\bb S}$.

\comment{Comment}
If you are a teacher looking for a problem to ask
in an exam, then your solution is a problem,
so ${\bb P} \subset {\bb S}$ makes sense.
And if you are a mathematician looking
for an algorithm to resolve some kind of problems,
then your solution is a resolution,
so ${\bb R} \subset {\bb S}$ makes sense.
There are many yes-or-no questions,
so ${\bb B} \subset {\bb S}$ makes sense.

\clause{Notation}\label{setofsolutions}
Let $\Sigma_\pi$ be the (possibly infinite) set of
all the solutions to problem $\pi$.
So $\Sigma_\pi \subseteq {\bb S}$, or $\Sigma_\pi \in 2^{\bb S}$,
and $\Sigma_\pi = \{\, s \mid P_\pi(s) \,\}$.

\comment{Comment}
A solution of the problem is
 any use of freedom that satisfies the condition,
see Section~\ref{Theory},
so $s$ is a solution of problem $\pi$, if $P_\pi(s)$ stands.

\comment{Comment}
The condition of the problem $\pi$ is
the characteristic function of 
its set of solutions, that is,
$P_\pi$ is the characteristic function of $\Sigma_\pi$.

\clause{Lemma}\label{operations}
$\Sigma_{\pi\lor\rho} = \Sigma_\pi \cup \Sigma_\rho$, and
$\Sigma_{\pi\land\rho} = \Sigma_\pi \cap \Sigma_\rho$, and
$\Sigma_{\bar{\pi}} = \overline{\Sigma_\pi}$.

\comment{Proof}
Just apply the definitions in \ref{problemcomposition}:\\
$\Sigma_{\pi\lor\rho} = \{\, s \mid P_{\pi\lor\rho}(s) \,\} =
           \{\, s \mid P_\pi(s) \lor P_\rho(s) \,\} =
           \{\, s \mid s \in \Sigma_\pi \lor s \in \Sigma_\rho \,\} =
           \Sigma_\pi \cup \Sigma_\rho$.\hfil\break
$\Sigma_{\pi\land\rho} = \{\, s \mid P_{\pi\land\rho}(s) \,\} =
           \{\, s \mid P_\pi(s) \land P_\rho(s) \,\} =
           \{\, s \mid s \in \Sigma_\pi \land s \in \Sigma_\rho \,\} =
           \Sigma_\pi \cap \Sigma_\rho$.\hfil\break
$\Sigma_{\bar{\pi}} = \{\, s \mid P_{\bar\pi}(s) \,\} =
           \{\, s \mid \neg P_\pi(s) \,\} =
           \{\, s \mid s \notin \Sigma_\pi \,\} = \overline{\Sigma_\pi}$.
{\QED}

\clause{Lemma}\label{neutrals}
For a tautological problem, $x? P_\tau(x)$,
everything is a solution, $\Sigma_\tau = {\bb S}$.
For a contradictory problem, $x? P_{\bar{\tau}}(x)$,
nothing is a solution, $\Sigma_{\bar \tau} = \emptyset$.

\comment{Proof}
$\Sigma_\tau = \{\, s \mid P_\tau(s) \,\} = \{\, s \mid \top \,\}
 = {\bb S}$.
$\Sigma_{\bar\tau} = \{\, s \mid P_{\bar\tau}(s) \,\} =
 \{\, s \mid \bot \,\} = \{\} = \emptyset$.
{\QED}

\clause{Lemma}\label{binarypartition}
$\Sigma_\pi \cup \Sigma_{\bar{\pi}} = {\bb S}$ and
$\Sigma_\pi \cap \Sigma_{\bar{\pi}} = \emptyset$.

\comment{Proof}
$\Sigma_\pi \cup \Sigma_{\bar{\pi}} = 
\{\, s \mid P_\pi(s) \,\} \cup \{\, s \mid \neg P_\pi(s) \,\}  =
 \{\, s \mid P_\pi(s) \lor \neg P_\pi(s) \,\} =
 \{\, s \mid \top \,\} = {\bb S}$.
$\Sigma_\pi \cap \Sigma_{\bar{\pi}} =
 \{\, s \mid P_\pi(s) \,\} \cap \{\, s \mid \neg P_\pi(s) \,\}  =
 \{\, s \mid P_\pi(s) \land \neg P_\pi(s) \,\} =
 \{\, s \mid \bot \,\} = \emptyset$.
{\QED}

\clause{Lemma}
The solutions of $\pi\land\rho$ are solutions of $\pi$ and of $\rho$.

\comment{Proof}
$\forall s \in {\bb S};\;
 s \in \Sigma_{\pi\land\rho} \;\Leftrightarrow\;
 s \in \Sigma_\pi \cap \Sigma_\rho \;\Leftrightarrow\;
 s \in \Sigma_\pi \land s \in \Sigma_\rho$.
{\QED}

\comment{Comment}
The reader is free to explore this Boolean landscape,
but here we will close with the following theorems.

\clause{Theorem}\label{powerSisalgebra}
$\langle 2^{\bb S}, \cup, \cap, -, \emptyset, {\bb S} \rangle$
is a Boolean algebra,
 where $\emptyset$ is the neutral for $\cup$,
 and $\bb S$ is the neutral for $\cap$.

\comment{Proof}
The powerset of a set $M$, with the operations of
 union $\cup$, intersection $\cap$, and complement with respect to set $M$,
 noted $\overline Q$,
is a typical example of a Boolean algebra. 
 In detail, $\forall Q,R,S \in 2^{\bb S}$:
\case 1o. $(Q\cup R)\cup S = Q\cup(R\cup S)$.\hfill
      1a. $(Q\cap R)\cap S = Q\cap(R\cap S)$.
\case 2o. $Q\cup R = R\cup Q$.\hfill
      2a. $Q\cap R = R\cap Q$.
\case 3o. $Q\cup\emptyset = Q$.\hfill
      3a. $Q\cap {\bb S} = Q$.
\case 4o. $Q\cup \overline Q = {\bb S}$.\hfill
      4a. $Q\cap \overline Q = \emptyset$.
\case 5o. $Q\cup(R\cap S) = (Q\cup R)\cap(Q\cup S)$.\hfill
      5a. $Q\cap(R\cup S) = (Q\cap R)\cup(Q\cap S)$.
{\QED}

\clause{Theorem}\label{setisomorphism}
$\left< {\bb P}, \lor, \land, \neg, \bar\tau, \tau \right>$
is isomorphic to
$\langle 2^{\bb S}, \cup, \cap, -, \emptyset, {\bb S} \rangle$,
that is,
${\bb P} \cong 2^{\bb S}$.

\comment{Proof}
We define the bijection $\Sigma$ that relates
each problem $\pi$ with the set of its solutions $\Sigma_\pi$:
for every problem $\pi \in {\bb P}$ there is a set,
 the set of its solutions, $\Sigma_\pi \in 2^{\bb S}$, and
for every set $S \in 2^{\bb S}$ there is a problem,
 $\pi_S =  x? [x \in S]$, where $\pi_S \in {\bb P}$.
Now, by Lemma~\ref{operations},
the bijection translates properly all three operations,
$\lor \leftrightarrow \cup$,
$\land \leftrightarrow \cap$,
$\neg \leftrightarrow -$,
and, by Lemma~\ref{neutrals}, also the two neutrals,
$\bar\tau \leftrightarrow \emptyset$,
$\tau \leftrightarrow {\bb S}$.
{\QED}

\comment{Comment}
We will call ${\bb P} \cong 2^{\bb S}$ the set isomorphism. 
That is, ${\bb P} \Leftrightarrow 2^{\bb S}:
 \pi \leftrightarrow \Sigma_\pi$.

\comment{Comment}
Using the set isomorphism,
two problems are equal if they have the same solutions,
that is, $\pi = \rho \Leftrightarrow \Sigma_\pi = \Sigma_\rho$.

\clause{Theorem}\label{setequality}
The set of problems is equal to the powerset of the solutions,
 that is, ${\bb P} = 2^{\bb S}$.

\comment{Proof}
The equality ${\bb P} = 2^{\bb S}$ derives directly
from the set isomorphism ${\bb P} \cong 2^{\bb S}$,
 see \ref{setisomorphism},
because no property was abstracted out.
{\QED}

\clause{Definition}
The set of singletons is:
${\bb S}^1 = \{\, S \in 2^{\bb S} \mid [\,|S| = 1\,] \,\}$.

\comment{Proposition}
${\bb S}^1 \subset 2^{\bb S}$, because
$\forall S \in {\bb S}^1$, $S \in 2^{\bb S}$,
but $\emptyset \in 2^{\bb S}$ and $\emptyset \notin{\bb S}^1$.

\clause{Definition}\label{singletonisomorphism}
The singleton isomorphism is the isomorphism
between ${\bb S}$ and ${\bb S}^1$ that
relates each $s\in{\bb S}$ to the set $\{s\} \in{\bb S}^1$,
and the converse. That is, ${\bb S} \cong {\bb S}^1$, and
${\bb S} \Leftrightarrow {\bb S}^1: s \leftrightarrow \{s\}$.

\comment{Comment}
We can extend any operation on $\bb S$ to ${\bb S}^1$.
For example, for any binary operation $*$ on $\bb S$,
we define $\{a\} * \{b\} = \{a*b\}$.

\comment{Comment}
From the singleton isomorphism:
${\bb S} \cong {\bb S}^1 \subset 2^{\bb S}$.

\clause{Lemma}\label{SsubsetP}
The set of solutions ${\bb S}$ is a proper subset
of the set of problems ${\bb P}$, that is,
${\bb S} \subset {\bb P}$.

\comment{Proof}
By the singleton isomorphism, see \ref{singletonisomorphism},
${\bb S} \cong {\bb S}^1$,
and, by the set isomorphism, see \ref{setisomorphism},
 for each singleton there is a problem, so
${\bb S}^1 \subset {\bb P}$, and then
 ${\bb S} \cong {\bb S}^1 \subset {\bb P}$.
{\QED}

\comment{Paradox}
We have both, ${\bb S} \subset {\bb P}$ and,
 by \ref{everythingisin},
${\bb P} \subset {\bb S}$.

\comment{Comment}
If we only accept computable functions and computable sets,
then ${\bb S}^* \not\subset {\bb P}^*$,
see Subsection~\ref{Countability}.

\clause{Definition}\label{solved}
A problem $\pi$ is solved if a solution of $\pi$ is known.

\comment{Comment}
To solve a problem,
given the set of its solutions $\Sigma_\pi$,
a choice function $f_c: 2^{\bb S} \setminus \emptyset \to {\bb S}$ is needed. 

\clause{Definition}\label{solvable}
A problem is unsolvable if $\Sigma_\pi = \{\} = \emptyset$,
that is, if $|\Sigma_\pi| = 0$.
A problem is solvable if $|\Sigma_\pi| > 0$.

\comment{Comment}
If a problem has not any solution, then it is unsolvable.
If a problem has a solution, then it can be solved.
A problem is solvable if it can be solved.

\comment{Comment}
Solved implies solvable, but not the converse:
$\hbox{Solved} \Rightarrow \hbox{Solvable}$.

\subsection{Routines and Trials}

\clause{Definition}\label{routine}
We will refer to the routine of problem $\pi$ as $R_\pi$.
The routine is the set of the solutions to the problem,
a set that is known, see \ref{Resolution}.
Then $R_\pi = \Sigma_\pi$.

\comment{Comment}
The routine of problem $\pi$, $R_\pi$,
is then, or
an extensive definition of $\Sigma_\pi$,
$\Sigma_\pi = \{s_1,\dots,s_n\}$, or
a procedure $\cal P$ that generates all problem $\pi$ solutions
and then halts.
If the number of solutions is infinite, $|\Sigma_\pi|\geq\aleph_0$,
then $R_\pi$ has to be a procedure $\cal P$ that 
keeps generating solutions forever.

\clause{Definition}
A trial on problem $\pi$ 
over the set of possible solutions $S$,
written $T_\pi(S)$,
returns the set of those elements in $S$
that satisfy the problem condition $P_\pi$,
 see \ref{Resolution}.
Then $T_\pi(S) = \{\, s \in S \mid P_\pi(s) \,\}$.

\comment{Comment}
Mathematically we will ignore the
practical problem of governing the trial.
Practically we will need a halt condition
to truncate the calculations that are too long (or infinite),
and some ordering on the tests
to fit the execution of the tests to the available calculating machinery.

\clause{Definition}
To test if a possible solution $s \in S$
 is a solution to problem $\pi$,
is to replace the free variable with $s$.
So, being $\pi = x? P_\pi(x)$, then
to test if $s$ is a solution
is to calculate $P_\pi(s)$.

\comment{Comment}
Testing is a calculation ${\bb S} \to {\bb B}$.

\clause{Remark}
Replacing variables in expressions
requires not confusing free with bound variables,
nor bound with free variables.

\comment{Comment}
This means that the rule of $\beta$-conversion
and the rules $\gamma$ for substitution
stand for testing.
See \cite Curry \& Feys (1958),
 Section~3D for $\beta$-conversion, and
 Section~3E for substitution (the rules $\gamma$).

\clause{Theorem}\label{trialisintersection}
A trial on problem $\pi$ over the set $S$
is equal to the intersection of $S$ with
the set of the solutions $\Sigma_\pi$, that is,
$T_\pi(S) = S \cap \Sigma_\pi$. 

\comment{Proof}
$T_\pi(S) = \{\, s \in S \mid P_\pi(s) \,\} =
          \{\, s \mid s \in S \land P_\pi(s) \,\} =
          \{\, s \mid s \in S \land s \in \Sigma_\pi \,\} =\\
          \{\, s \mid s \in S \,\} \cap \{\, s \mid s \in \Sigma_\pi \,\} =
          S \cap \Sigma_\pi$.
{\QED}

\comment{Corollary}
Any trial is a subset of the set of solutions,
$T_\pi(S) \subseteq \Sigma_\pi$.

\comment{Proof}
$T_\pi(S) = S \cap \Sigma_\pi \subseteq \Sigma_\pi$.
{\QED}

\comment{Corollary}
Any trial is a subset of the routine, that is,
$T_\pi(S) \subseteq \Sigma_\pi = R_\pi$.

\clause{Lemma}\label{trialonsuperset}
If $S$ is a superset of $\Sigma_\pi$,
then a trial on problem $\pi$ over $S$
is equal to $\Sigma_\pi$, and the converse, that is,
$\Sigma_\pi \subseteq S \Leftrightarrow T_\pi(S) = \Sigma_\pi$.

\comment{Proof}
$\Sigma_\pi \subseteq S \Leftrightarrow S \cap \Sigma_\pi =
 \Sigma_\pi \Leftrightarrow T_\pi(S) = \Sigma_\pi$,
using Theorem~\ref{trialisintersection}.
{\QED}

\comment{Corollary}
If $S$ is a superset of $\Sigma_\pi$,
then a trial on problem $\pi$ over $S$
is equal to the routine of $\pi$, and the converse, that is,
$\Sigma_\pi \subseteq S \Leftrightarrow T_\pi(S) = R_\pi$.

\comment{Proof}
$\Sigma_\pi \subseteq S \Leftrightarrow S \cap \Sigma_\pi =
 \Sigma_\pi \Leftrightarrow T_\pi(S) = R_\pi$.
{\QED}

\comment{Corollary}
A trial on problem $\pi$ over the whole ${\bb S}$
 is equal to $\Sigma_\pi$,
that is, $T_\pi({\bb S}) = \Sigma_\pi$.

\comment{Proof}
Because $\Sigma_\pi \subseteq {\bb S}$.
{\QED}

\comment{Comment}
$T_\pi({\bb S})$ is an exhaustive search.

\clause{Theorem}\label{routineastrial}
The routine is a trial over all the solutions, that is,
$R_\pi = T_\pi(\Sigma_\pi)$.

\comment{Proof}
By Theorem~\ref{trialisintersection},
$T_\pi(\Sigma_\pi) = \Sigma_\pi \cap \Sigma_\pi =
 \Sigma_\pi = R_\pi$.
{\QED}

\comment{Comment}
$T_\pi(R_\pi) = T_\pi(\Sigma_\pi) = \Sigma_\pi = R_\pi$.

\subsection{Analogies}

\clause{Definition}
If $A$ is an analogy, and $\pi = x? P_\pi(x)$ is a problem,
then $A\pi$ is another problem $A\pi = x? P_{\!A\pi}(x)$.
That is, $A: {\bb P} \to {\bb P}$.

\comment{Comment}
So analogies transform a condition into a condition,
$P_\pi$ into $P_{\!A\pi}$ in this example.

\comment{Comment}
Taking advantage of problem decomposition, see \ref{problemcomposition},
the result of an analogy, $A\pi$, can be a composition of problems
that are easier to resolve than the original problem, $\pi$,
see \ref{Resolution}.

\clause{Definition}\label{conservativeanalogy}
If $\Sigma_\pi = \Sigma_{A\pi}$,
then we say that the analogy is conservative.

\comment{Comment}
If an analogy is not conservative, then
a function ${\cal T}_A$ to translate $\Sigma_{A\pi}$ to $\Sigma_\pi$
is required, because otherwise the analogy would be useless.

\clause{Notation}
We will call function ${\cal T}_A$
the translating function of analogy $A$.\\
${\cal T}_A: 2^{\bb S} \to 2^{\bb S}$ and
${\cal T}_A(\Sigma_{A\pi}) = \Sigma_\pi$.

\clause{Lemma}\label{analogyanalogy}
An analogy followed by another one is an analogy.

\comment{Proof}
Because any analogy transforms a problem into a problem:
${\bb P} \to {\bb P}$.
{\QED}

\comment{Corollary}
Analogies can be chained.

\clause{Lemma}\label{analogiesonly}
Using only analogies we cannot resolve any problem.

\comment{Proof}
Because using analogies we only get problems.
{\QED}

\comment{Comment}
While routines $R$ and trials $T(S)$ are
 functions that return a set, ${\bb P} \to 2^{\bb S}$,
analogies $A$ are functions that return a function,
${\bb P} \to {\bb P}$.

\clause{Notation}
We will write $A\circ T$ to express the composition
of functions, where $A$ is applied first and then $T$.

\comment{Comment}
$[A\circ T](x) = T(A(x))$. Diagram:
$x \into{A} A(x) \into{T} T(A(x))$.

\comment{Comment}
If $A_1$ and $A_2$ are analogies,
then $A_1\circ A_2$ is also an analogy, by Lemma~\ref{analogyanalogy}.

\clause{Definition}
To resolve a problem by analogy $A$ is to
compose $A \circ \Re \circ {\cal T}_A$,
 where $\Re$ is any resolution,
 and ${\cal T}_A$ is the translating function of $A$. Diagrams:
$${\pi} \into A {A\pi} \into\Re {\Sigma_{A\pi}} \into{{\cal T}_A} {\Sigma_\pi}
\qquad\hbox{or}\qquad
  {\bb P} \into A {\bb P} \into\Re 2^{\bb S} \into{{\cal T}_A} 2^{\bb S}
\;.$$

\comment{Comment}
Analogy $A$ is a translation from some original problem domain
to some analogue problem domain.
Then, by Lemma~\ref{analogiesonly},
 we need a resolution $\Re$ to resolve the analogue problem.
And, finally, we need to translate the solutions back to the 
original domain.

\clause{Lemma}\label{translatingofcomposition}
The translating function of the composition $A\circ A'$
is ${\cal T}_{A'} \circ {\cal T}_{A}$.

\comment{Proof}
If $\Re = A' \circ \Re' \circ {\cal T}_{A'}$ then we get
$A \circ ( A' \circ  \Re' \circ {\cal T}_{A'} ) \circ {\cal T}_A = 
A \circ A' \circ  \Re' \circ {\cal T}_{A'} \circ {\cal T}_A =
(A \circ A') \circ  \Re' \circ ({\cal T}_{A'} \circ {\cal T}_A) $,
because function composition is associative.
Diagram:
$${\bb P} \into{A}
  \underbrace{\strut
    {\bb P}\into{A'}{\bb P}\into{\Re'}2^{\bb S}\into{{\cal T}_{A'}}2^{\bb S}
  }_\Re
      \into{{\cal T}_{A}} 2^{\bb S} .\quad{\QED}$$

\comment{Corollary}
The translating function of the composition $A_1\circ A_2 \dots \circ A_n$
is ${\cal T}_{A_n} \circ \dots \circ {\cal T}_{A_2} \circ {\cal T}_{A_1}$.
That is: ${\cal T}_{A_1\circ A_2 \dots \circ A_n} =
{\cal T}_{A_n} \circ \dots \circ {\cal T}_{A_2} \circ {\cal T}_{A_1}$.

\comment{Comment}
This is how analogies can be chained.

\clause{Definition}\label{identityfunction}
The identity function, written $I$, transforms
anything into itself: $\forall x,\;I(x) = x$.

\comment{Comment}
The identity function $I$
is an effectively calculable function,
see \ref{effectivelycalculable}.
It is $\lambda$-definable;
in $\lambda$-calculus, $I=(\lambda x.x)$.

\comment{Comment}
Identity $I$ transforms $\pi$ into $\pi$, $I(\pi)=\pi$, 
and $P_\pi$ into $P_\pi$, $I(P_\pi)=P_\pi$.

\comment{Comment}
Identity $I$ can work as an analogy: $I\pi = I(\pi) = \pi$.

\clause{Lemma}
The translating function of the identity analogy
is the identity function: ${\cal T}_I = I$.

\comment{Proof}
Because $I(\Sigma_\pi) = \Sigma_\pi$. Diagram:
$\pi \into{I} \pi \into\Re \Sigma_\pi \into{I} \Sigma_\pi$.
{\QED}

\comment{Comment}
The identity analogy is conservative,
 see \ref{conservativeanalogy}.

\clause{Lemma}
The identity $I$ followed by any function $f$,
or any function $f$ followed by identity $I$,
is equal to the function: $\forall f,\; I\circ f = f = f\circ I$.

\comment{Proof}
$\forall f,\forall x,\; [I\circ f](x) = f(I(x)) = f(x) = I(f(x)) = [f\circ I](x)$.
{\QED}

\comment{Comment}
$I\circ \Re(I\pi)\circ {\cal T}_I=I\circ \Re(\pi)\circ I=\Re(\pi)$.

\clause{Theorem}\label{generalform}
$A\circ {T\!}_{A\pi}(S)\circ {\cal T}_A$,
 where $A$ is an analogy,
 ${T\!}_{A\pi}(S)$ is a trial, and
 ${\cal T}_A$ is the translating function of $A$,
 is the general form of a resolution.

\comment{Proof}
If the analogy is the identity $I$,
then the general form is reduced to
$T_\pi(S)$, because ${\cal T}_I = I$, $I\pi = \pi$, so
$I\circ T_{I\pi}(S)\circ I = T_\pi(S)$, which is a trial.
By Theorem~\ref{routineastrial},
a routine is a specific trial, $R_\pi = T_\pi(R_\pi)$,
so $I\circ T_\pi(R_\pi)\circ I= T_\pi(R_\pi)=R_\pi$
reduces the general form to the routine.
Resolving by analogy is, by definition,
$A\circ \Re \circ {\cal T}_A$, and analogies can be chained,
by Lemma~\ref{analogyanalogy},
 so a chain of analogies is an analogy, $A_1\circ A_2\circ \dots\circ A_n = A$,
 and by Lemma~\ref{translatingofcomposition},
 ${\cal T}_A = {\cal T}_{A_1\circ A_2\circ \dots\circ A_n} =
 {\cal T}_{A_n}\circ \dots\circ {\cal T}_{A_2}\circ {\cal T}_{A_1}$.
Then 
$A_1 \circ A_2 \circ \dots\circ A_n \circ
 {T\!}_{A\pi}(S) \circ
 {\cal T}_{A_n} \circ \dots \circ {\cal T}_{A_2} \circ {\cal T}_{A_1} =
 A \circ {T\!}_{A\pi}(S) \circ {\cal T}_A$.
{\QED}

\comment{Summary}
There are three ways to resolve a problem:
routine $R_\pi = I\circ T_\pi(R_\pi)\circ I$,
trial $T_\pi(S) = I\circ T_\pi(S)\circ I$, and
analogy $ A_1 \circ \dots \circ A_n \circ
 {T\!}_{A\pi}(S) \circ
 {\cal T}_{A_n} \circ \dots \circ {\cal T}_{A_1} =
 A \circ {T\!}_{A\pi}(S) \circ {\cal T}_A$.

\subsection{Metaproblems}

\clause{Definition}\label{validresolution}
A resolution $\Re: {\bb P}\to2^{\bb S}$
is a valid resolution for a problem $\pi$
if it finds all the solutions of problem $\pi$ and then halts.
In other words,
$\Re$ is a valid resolution for $\pi$ if it satisfies two conditions:
that $\Re(\pi)$ is effectively calculable, and
that $\Re$ fits problem $\pi$, that is, that $\Re(\pi) = \Sigma_\pi$.

\comment{Comment}
If $\Sigma_\pi$ is infinite, $|\Sigma_\pi|\geq\aleph_0$,
then a valid $\Re(\pi)$ does not halt,
but it keeps building $\Sigma_\pi$ forever.

\clause{Definition}\label{resolved}
A problem $\pi$ is resolved if a valid resolution for $\pi$ is known.

\comment{Comment}
To solve a problem we have to find one solution,
 see \ref{solved}.
To resolve a problem we have to find all the solutions.
To resolve a problem is to exhaust the problem.

\clause{Lemma}
Once a problem is resolved,
we can thereafter resolve it by routine.

\comment{Proof}
Once a problem is resolved,
we know all of its solutions, $\Sigma_\pi$,
and knowing $\Sigma_\pi$, we know its routine resolution,
because $R_\pi = \Sigma_\pi$, see \ref{routine}.
{\QED}

\comment{Proposition}
If $\pi\land\rho$ is solvable, then
by resolving $\pi\land\rho$ both $\pi$ and $\rho$ are solved.

% \comment{Proof}
% If $\pi\land\rho$ is solvable, see \ref{solvable}, then
% $\pi\land\rho$ has, at least, one solution, let us call it $s$,
% which is a solution of $\pi$ and it is also a solution of $\rho$.
% By resolving $\pi\land\rho$, see \ref{resolved},
% we get all the solutions of $\pi\land\rho$, including $s$.
% By knowing $s$, both $\pi$ and $\rho$ get solved,
% see \ref{solved}.
% {\QED}

\clause{Definition}\label{resolvable}
A problem is resolvable
if there is a valid resolution for the problem,
see \ref{validresolution},
that is, if there is a resolution $\Re$
such that $\Re(\pi)$ is effectively calculable,
      and $\Re(\pi) = \Sigma_\pi$.
Otherwise, the problem is unresolvable.

\comment{Comment}
A problem is resolvable if it can be resolved.

\comment{Comment}
Resolved implies resolvable, but not the converse:
$\hbox{Resolved} \Rightarrow \hbox{Resolvable}$.

\clause{Definition}
For any Boolean-valued function $P: {\bb S} \to {\bb B}$,
we define the function $\check P: {\bb B} \to 2^{\bb S}$,
called the inverse of condition $P$,
as follows:
$$\eqalign{
 \check P(\top) &= \{\, x \mid [P(x) = \top] \,\} ,\cr
 \check P(\bot) &= \{\, x \mid [P(x) = \bot] \,\} .\cr}$$

\clause{Lemma}\label{inversecondition}
If $P_\pi(x)$ is the condition of a problem $\pi$,
 then $\check P_\pi(\top) = \Sigma_\pi$ and\\
 $\check P_\pi(\bot) = \overline{\Sigma_\pi} = \Sigma_{\bar \pi}$.

\comment{Proof}
Because 
$\check P_\pi(\top) = \{\, x \mid [P_\pi(x)=\top] \,\} =
     \{\, x \mid P_\pi(x) \,\} = \Sigma_\pi$, and\\
$\check P_\pi(\bot) = \{\, x \mid [P_\pi(x)=\bot] \,\} =
     \{\, x \mid \neg P_\pi(x) \,\} = \overline{\Sigma_\pi} =
  \Sigma_{\bar \pi}$, by Lemma~\ref{operations}.
{\QED}

\clause{Theorem} The inverse of the condition of a problem,
provided it is an effectively calculable function,
resolves the problem and its complementary by routine.

\comment{Proof}
By \ref{inversecondition} and \ref{routine},
% By Lemma \ref{inversecondition} and Definition \ref{routine},
$\check P_\pi(\top) = \Sigma_\pi = R_\pi$,
then $\check P_\pi(\top)$
is the routine resolution of $\pi$,
if $\check P_\pi(\top)$ is effectively calculable,
see \ref{effectivelycalculable}.
And if $\check P_\pi(\bot)$ is effectively calculable,
then it resolves the complementary problem by routine,
$\check P_\pi(\bot) = \Sigma_{\bar\pi} = R_{\bar\pi}$.
{\QED}

\comment{Comment}
It is a nice theorem,
but how can we find the inverse of a condition?

\clause{Definition}\label{metaproblem}
The metaproblem of a problem, written $\Pi\pi$, is
the problem of finding the valid resolutions for problem $\pi$.
In other words, if $\pi = x? P_\pi(x)$,
then $\Pi\pi = \Re? [\Re(\pi) = \Sigma_\pi]$.

\comment{Comment}
The solutions of the metaproblems are the resolutions,
$\Pi{\bb S} = {\bb R}$.

\comment{Comment}
The condition of the metaproblem, $P_{\Pi\pi}$, is
$[\Re(\pi) = \Sigma_\pi]$, that is,
$P_{\Pi\pi}(\Re) = [\Re(\pi) = \Sigma_\pi]$,
 or using an $\alpha$-conversion,
$P_{\Pi\pi}(x) = [x(\pi) = \Sigma_\pi]$.

\clause{Lemma}\label{metaproblemisproblem}
A metaproblem is a problem,
that is, $\Pi{\bb P} \subset {\bb P}$.

\comment{Proof}
Because $\Pi\pi = x? P_{\Pi\pi}(x)$,
but some problems are not metaproblems.
{\QED}

\comment{Comment}
A metaproblem is a problem because it has its two ingredients:
there are several ways to resolve a problem,
so there is freedom, but
only the valid resolutions resolve the problem,
so there is a condition.

\clause{Definition}\label{metacondition}
The metacondition $P_\Pi$ is
$P_\Pi(p,r)= [r(p) = \Sigma_p]$,
for any problem $p\in{\bb P}$, and
for any resolution $r\in{\bb R}$.

\comment{Comment}
Using another $\alpha$-conversion,
$P_\Pi(\pi,x) = [x(\pi) = \Sigma_\pi] = P_{\Pi\pi}(x)$.

\comment{Comment}
$\Pi\pi = x? P_\Pi(\pi,x)$.

\clause{Definition}\label{metaresolving}
Metaresolving is resolving the metaproblem to resolve the problem.

\comment{Comment}
Metaresolving is a kind of analogy. Diagram:
$${\pi} \into\Pi
  {\Pi\pi} \into{\Pi\Re}
  {\Sigma_{\Pi\pi}}=\{\,\Re \mid [\Re(\pi)=\Sigma_\pi] \,\}
  \into{f_c} \Re_c \into{(\pi)} \Re_c(\pi)=\Sigma_\pi
\;.
$$
Function $f_c$ is a choice function,
and the last calculation, noted $(\pi)$,
means to apply $\pi$ as the argument, not as the function.
If you only metasolve, then you don't need to choose.
In any case, the translating function of metaresolving is
${\cal T}_\Pi = f_c \circ (\pi)$.
Then we can draw the following diagrams:
$${\pi} \into\Pi {\Pi\pi}
        \into{\Pi\Re} {\Sigma_{\Pi\pi}}
        \into{{\cal T}_\Pi} {\Sigma_\pi}
\qquad\hbox{or}\qquad
{\bb P} \into\Pi \Pi{\bb P} 
        \into{\Pi\Re} 2^{\Pi\bb S} = 2^{\bb R}
        \into{{\cal T}_\Pi} 2^{\bb S}
\;.$$

\clause{Lemma}\label{solvemetaisresolve}
The metaproblem $\Pi\pi$ of some problem $\pi$ is solvable if,
and only if, the problem $\pi$ is resolvable, that is,
$\Pi\pi \hbox{ is solvable} \Leftrightarrow
 \pi \hbox{ is resolvable}$.

\comment{Proof}
If $\Pi\pi$ is solvable, then there is a solution to it,
 see \ref{solvable},
and that solution is a valid resolution for $\pi$,
 see \ref{metaproblem},
and then $\pi$ is resolvable,
 see \ref{resolvable}.
If $\pi$ is resolvable, then there is a valid resolution for it,
 see \ref{resolvable},
and that resolution is a solution of its metaproblem $\Pi\pi$,
 see \ref{metaproblem},
and then $\Pi\pi$ is solvable,
 see \ref{solvable}.
{\QED}

\comment{Corollary}
To solve the metaproblem $\Pi\pi$ of problem $\pi$
is to resolve problem $\pi$.

\comment{Proof}
Because to resolve problem $\pi$ is to find
a valid resolution for $\pi$, see~\ref{resolved},
and to solve the metaproblem $\Pi\pi$ is
 to find a solution to $\Pi\pi$, see~\ref{solved},
which is also to find a valid resolution for $\pi$,
see~\ref{metaproblem}.
{\QED}

\comment{Comment}
And again, ${\bb R}=\Pi{\bb S}$.

\clause{Lemma}\label{metaproblemroutine}
The set of the valid resolutions for problem $\pi$
is the routine resolution of its metaproblem $\Pi\pi$,
that is,
$\{\, \Re \mid [\Re(\pi) = \Sigma_\pi] \,\} = R_{\Pi\pi}$.

\comment{Proof}
$R_{\Pi\pi} = \Sigma_{\Pi\pi}$, by the definition of routine,
 see~\ref{routine}.\\ And
$\Sigma_{\Pi\pi} = \{\, \Re \mid [\Re(\pi) = \Sigma_\pi] \,\}$,
 by the definition of $\Pi\pi$,
 see~\ref{metaproblem}.
{\QED}

\subsection{Resolution Typology}

\clause{Definition}
The meta$^n$-metaproblem of $\pi$, $\Pi^n\Pi\pi$,
is (the metaproblem of)$^n$ the metaproblem of $\pi$,
where $n \in {\bb N}$.

\comment{Special case}
The meta-metaproblem of $\pi$, $\Pi\Pi\pi = \Pi^1\Pi\pi$,
is the metaproblem of the metaproblem of $\pi$.

\comment{Examples}
$\Pi^0\Pi\pi = \Pi\pi$.
$\Pi^1\Pi\pi = \Pi\Pi\pi = \Pi^2\pi$.
$\Pi^2\Pi\pi = \Pi\Pi\Pi\pi = \Pi^3\pi$.

%\comment{Comment}
%The meta$^0$-metaproblem of $\pi$ is
%the metaproblem of $\pi$, $\Pi^0\Pi\pi = \Pi\pi$.
%The meta$^2$-metaproblem of $\pi$ is
%the meta-meta-metaproblem of $\pi$,
%$\Pi^2\Pi\pi = \Pi\Pi\Pi\pi = \Pi^3\pi$.

\comment{Comment}
From $\Pi{\bb S} = {\bb R}$,
we get $\Pi\Pi{\bb S} = \Pi{\bb R}$ and
$\Pi^n\Pi{\bb S} = \Pi^n{\bb R}$.

\comment{Comment}
The condition of the meta$^n$-metaproblem of $\pi$,
$P_{\Pi^n\Pi\pi}$, where $n \in {\bb N}$, is:
$P_{\Pi^n\Pi\pi}(x) = [x(\Pi^n\pi) = \Sigma_{\Pi^n\pi}]$.

\comment{Examples}
$P_{\Pi^0\Pi\pi}(x) = [x(\Pi^0\pi) = \Sigma_{\Pi^0\pi}] =
 [x(\pi) = \Sigma_\pi] = P_{\Pi\pi}(x)$.\\
$P_{\Pi^1\Pi\pi}(x) = [x(\Pi^1\pi) = \Sigma_{\Pi^1\pi}] =
 [x(\Pi\pi) = \Sigma_{\Pi\pi}] = P_{\Pi\Pi\pi}(x)$.

\clause{Lemma}\label{metametapiispi}
A meta$^n$-metaproblem is a problem,
where $n \in {\bb N}$.

\comment{Proof}
If $n>0$, then
$\Pi^n\Pi\pi  = x? P_{\Pi^n\Pi\pi}(x)$.
For $n=0$, see \ref{metaproblemisproblem}.
{\QED}

\comment{Corollary}
$\bigcup_{n\in{\bb N}}\Pi^n\Pi{\bb P} \subset {\bb P}$.

\clause{Definition}
The meta$^n$-metacondition $P_{\Pi^n\Pi}$,
 with $n \in {\bb N}$, $p \in {\bb P}$, and $r \in {\bb R}$
is:
$P_{\Pi^n\Pi}(p,r) = [r(\Pi^n p) = \Sigma_{\Pi^n p}]$.

\comment{Comment}
Using an $\alpha$-conversion,
$P_{\Pi^n\Pi}(\pi,x) = [x(\Pi^n\pi) = \Sigma_{\Pi^n\pi}] =
 P_{\Pi^n\Pi\pi}(x)$.

\comment{Example}
$P_{\Pi^1\Pi}(\pi,x) = P_{\Pi\Pi}(\pi,x) =
 [x(\Pi\pi) = \Sigma_{\Pi\pi}] =
 [x(\Pi^1\pi) = \Sigma_{\Pi^1\pi}]$.

\clause{Lemma}\label{metametacondition}
$P_{\Pi^n\Pi}(\pi,x) = P_\Pi(\Pi^n\pi,x)$, where $n \in {\bb N}$.

\comment{Proof}
By \ref{metacondition},
$P_\Pi(\Pi^n\pi,x) = [x(\Pi^n\pi) = \Sigma_{\Pi^n\pi}] =
 P_{\Pi^n\Pi}(\pi,x)$.
{\QED}

\comment{Special Case}
$P_{\Pi\Pi}(\pi,x) = P_\Pi(\Pi\pi,x)$.

% \comment{Proof}
% $P_\Pi(\Pi\pi,x) = [x(\Pi\pi) = \Sigma_{\Pi\pi}] =
%  [x(\Pi^1\pi) = \Sigma_{\Pi^1\pi}] = P_{\Pi^1\Pi}(\pi,x) =
%  P_{\Pi\Pi}(\pi,x)$.
% {\QED}

\comment{Comment}
The meta-metacondition is the metacondition of the metaproblem.

\clause{Lemma}\label{metanismeta}
A meta$^n$-metaproblem is a metaproblem, where $n \in {\bb N}$.
%%that is, $\Pi^n\Pi{\bb P} \subseteq \Pi{\bb P}$.

\comment{Proof}
If $n > 0$,
$\Pi^n\Pi\pi = x? P_\Pi(\Pi^n\pi,x)$,
and $\Pi^0\Pi\pi = \Pi\pi = x? P_\Pi(\pi,x)$.
%%$\Pi^0\Pi{\bb P} = \Pi{\bb P}$.
{\QED}

\comment{Corollary}
$\bigcup_{n\in{\bb N}}\Pi^n\Pi{\bb P} = \Pi{\bb P}$.

\clause{Lemma}\label{series}
We have the following infinite series of mathematical objects:\\
${\bb S}$,
${\bb P} =  2^{\bb S}$,
${\bb R} = \Pi{\bb S} = 2^{\bb S} \to 2^{\bb S}$,
$\Pi{\bb P} = 2^{2^{\bb S} \to 2^{\bb S}}$,
$\Pi{\bb R} = \Pi\Pi{\bb S} =
 2^{2^{\bb S} \to 2^{\bb S}} \to 2^{2^{\bb S} \to 2^{\bb S}}$,
$\dots$

\comment{Proof}
${\bb P} = {\bb S} \to {\bb B} = 2^{\bb S}$,
 by Theorems \ref{conditionequality} and \ref{setequality}.\\
$\Pi{\bb S} = {\bb R}$, by the metaproblem definition,
see \ref{metaproblem}, and
${\bb R} = {\bb P} \to 2^{\bb S} =  2^{\bb S} \to 2^{\bb S}$.\\
$\Pi{\bb P} = \Pi{\bb S} \to {\bb B} =
 {\bb R} \to {\bb B} = 2^{\bb R} =
  2^{2^{\bb S} \to 2^{\bb S}}$.\\
$\Pi{\bb R} = \Pi{\bb P} \to 2^{\Pi{\bb S}} =
 2^{\bb R} \to 2^{\bb R} =
 2^{2^{\bb S} \to 2^{\bb S}} \to 2^{2^{\bb S} \to 2^{\bb S}}$.\\
And so on.
{\QED}

\clause{Theorem}\label{onemetaness}
There is only one level of problem meta-ness.

\comment{Proof}
By Lemma \ref{metanismeta},
because  every meta$^n$-metaproblem is a metaproblem, and
every metaproblem is a meta$^0$-metaproblem, so
$\bigcup_{n\in{\bb N}}\Pi^n\Pi{\bb P} = \Pi{\bb P} \subset {\bb P}$.
{\QED}

\comment{Comment}
While a problem condition is any predicate, $P(x)$,
a metaproblem condition is a specific kind of predicate,
namely, $P_\Pi(p,r) = [r(p) = \Sigma_p]$.
And any meta$^n$-metaproblem condition, $P_{\Pi^n\Pi}$,
is the same specific predicate $P_\Pi$,
see \ref{metametacondition}.

\comment{Comment}
We are assuming that
functions are free to take functions as arguments.
See that, in predicate $P_\Pi(p,r) = [r(p) = \Sigma_p]$,
argument $r$ is a function in $\Pi^n{\bb R}$
that takes $p\in\Pi^n {\bb P}$ as argument.
Therefore, the theorem holds unconditionally for
$\lambda$-definable functions,
including predicates, see \ref{universalareequivalent}.
And then, under Church's thesis, see \ref{Turingsthesis},
the theorem is true for effectively calculable functions,
and in particular, it is true for expressible and for resolvable
problems, see \ref{expressible} and \ref{resolvable}.

\clause{Theorem}\label{fivetypesofresolution}
There are five types of resolution.

\comment{Proof}
From Theorem~\ref{generalform}
we get three types for the resolution of problems:
$R_\pi$, $T_\pi(S)$, and
$A\circ {T\!}_{A\pi}(S)\circ {\cal T}_A$.
This shows that there are several ways of resolving, 
so choosing a resolution that find solutions
to the original problem $\pi$
is another problem, the metaproblem $\Pi\pi$,
see \ref{metaproblem}.
Then we should get another three for the resolution
of the metaproblem, but,
by \ref{metaproblemroutine},
the set of the resolutions of a problem
is the routine resolution of its metaproblem,
so we only add two more for the metaproblem:
$T_{\Pi\pi}(R)$, and
${\cal A}\circ {T\!}_{{\cal A}\Pi\pi}(R)\circ {\cal T}_{\cal A}$.
Finally, by \ref{onemetaness},
we do not need to go deeper
into meta$^n$-metaproblems.
{\QED}

\comment{Comment}
We will call them:
routine $R_\pi$, trial $T_\pi(S)$,
analogy $A\circ {T\!}_{A\pi}(S)\circ {\cal T}_A$,
meta-trial $T_{\Pi\pi}(R)$, and meta-analogy
 ${\cal A}\circ {T\!}_{{\cal A}\Pi\pi}(R)\circ {\cal T}_{\cal A}$.
The first three can also be called meta-routines.

\clause{Remark}\label{metadiagrams}
The diagram for the meta-trial, or trial of the metaproblem, is:
$${\pi} \into\Pi {\Pi\pi}
  \into{T_{\Pi\pi}(R)} {\Sigma_{\Pi\pi}}
  \into{{\cal T}_\Pi} {\Sigma_\pi} \;.
$$
And the diagram for the meta-analogy, or analogy of the metaproblem, is:
$${\pi} \into\Pi {\Pi\pi} \into{\cal A} {{\cal A}\Pi\pi}
  \into{T_{{\cal A}\Pi\pi}(R)} \Sigma_{{\cal A}\Pi\pi}
  \into{{\cal T}_{\cal A}} \Sigma_{\Pi\pi} \into{{\cal T}_\Pi} \Sigma_\pi \;.
$$
See that 
${\cal A} : \Pi{\bb P} \to \Pi{\bb P} = % 2^{\bb R}\to2^{\bb R} =
 2^{2^{\bb S}\to2^{\bb S}}\to2^{2^{\bb S}\to2^{\bb S}}$
and
${\cal T}_{\cal A} : 2^{\Pi{\bb S}} \to 2^{\Pi{\bb S}} =
 2^{2^{\bb S}\to2^{\bb S}}\to2^{2^{\bb S}\to2^{\bb S}}$,
using \ref{series}.
Both are functions
   taking sets of functions on sets to sets and
returning sets of functions on sets to sets.

\section{Computers}

\subsection{Turing Machine}

\clause{Definition}
A computation is any manipulation of a string of symbols,
irrespective of the symbols meanings,
but according to a finite set of well-defined rules.

\comment{Comment}
Computing is any mechanical transformation of a string of symbols.

\clause{Definition}
A computing device, or computer,
is any mechanism that can perform computations.

\comment{Comment}
The prototype of computing device
 is a Turing machine, see \cite Turing (1936).

\clause{Notation}\label{Turingmachine}
The Turing machine has two parts:
the processor ${\cal P}$,
 which is a finite state automaton,
and an infinite tape, $\TM<\;>$,
 which in any moment contains only a finite number of symbols.

\comment{Comment}
In the case of a processor of a Turing machine,
the output alphabet $O$,
that is, the finite set of output symbols, has to be:
$O = I^+ \times \{l,h,r\}$,
where $I$ is the finite not empty input alphabet,
$I^+ = I \cup \{ b \}$, where $b \notin I$ is {\it blank},
and $l$, $h$, and $r$ mean
{\it left}, {\it halt}, and {\it right}.
Then its transition function is
${\cal T}: S \times I^+ \to S \times I^+ \times \{l,h,r\}$,
where $S$ is the finite set of internal states.
And the strings that the Turing machine transforms are
sequences of symbols taken from set $I$.

\clause{Notation}\label{computingsets}
We will refer to the set of Turing machines as $\frak T$.
We will refer to the set of the strings of symbols as $\frak E$.

\comment{Comment}
Because all Turing machines tapes are equal,
the processor defines the Turing machine, and therefore
we will refer to the Turing machine with processor $\cal P$
as the Turing machine $\cal P$, and then $\cal P \in \frak T$.
We will refer to the string of symbols written on the tape
as the expression $\frak e \in E$.

\clause{Lemma}\label{expressionsarecountable}
The set of expressions is countable, that is,
$|{\frak E}| = |{\bb N}| = {\aleph}_0$.

\comment{Proof}
Let $I$ be any finite alphabet, and $s$ its cardinality,
that is, $s$ is the number of symbols, $s = |I| > 0$.
We write $I^n$ the set of strings of length $n$,
so $|I^n| = s^n$.
Then ${\frak E} = \bigcup_{n\in{\bb N}} I^n$,
and we can define a bijection
 between ${\frak E}$ and ${\bb N}$ this way:
 it maps the empty string in $I^0$ to $0$,
 it maps the $s$ strings in $I^1$ to the next $s$ numbers,
 it maps the $s^2$ strings in $I^2$ to the next $s^2$ numbers,
 and so on.
Note that ordering the symbols in $I$,
we can order alphabetically the strings in each $I^n$.
{\QED}

\comment{Comment}
Most real numbers are not expressible.
See \cite Turing (1936) \S10 for details;
but, for example, transcendental numbers
$\pi$ and $e$ are computable, page 256.

\clause{Notation}
We will use the notation
$\TM P<e> \hookrightarrow {\frak r}$
to indicate that,
if we write the expression~${\frak e} \in {\frak E}$
on the tape of the Turing machine with processor ${\cal P}$ and
we leave it running, then when it halts we will find the
expression~$\frak r \in E$ on the tape.
If, on the contrary, the Turing machine $\cal P$ does not halt
when we write the expression~$\frak w$,
then we would say that $\frak w$ is a paradox in $\cal P$,
and we would indicate this as follows:
$\TM P<w> \hookrightarrow \infty$.

\clause{Definition}
${\frak E}^+ = {\frak E} \cup \{\infty\}$.

\comment{Comment}
Some computations do not halt,
so we need $\infty$ to refer to them.
Note that $\infty \notin {\frak E}$,
but $\infty \in {\frak E}^+$.
So ${\frak E} \subset {\frak E}^+$.

\clause{Definition}\label{Turingequivalence}
For each Turing machine ${\cal P} \in {\frak T}$ we define
a function $F_{\cal P}: {\frak E} \to {\frak E}^+$,
this way:
$$F_{\cal P}({\frak e}) = 
\cases{{\frak r} & if $\TM P<e> \hookrightarrow {\frak r}$\cr
       \infty    & if $\TM P<e> \hookrightarrow \infty$ .\cr}$$

\comment{Comment}
If $\forall{\frak e} \in {\frak E},
 F_{\cal P}({\frak e}) = F_{\cal Q}({\frak e})$,
then we say that Turing machines $\cal P$ and $\cal Q$ are
behaviorally equivalent, ${\cal P} \equiv_F {\cal Q}$,
or that $\cal P$ and $\cal Q$ implement the same function.

\clause{Definition}\label{computable}
We say that a function is computable
if there is a Turing machine that implements the function.

\clause{Lemma}
For each Turing machine we can define a unique finite string of symbols,
that is, $\exists{\frak c}: {\frak T} \to {\frak E}$ such that
$ {\cal P} = {\cal Q} \;\Leftrightarrow\; {\frak c}({\cal P}) = {\frak c}({\cal Q})$.

\comment{Proof}
Proved by \cite Turing (1936), \S5.
Turing machines are defined by their processors,
which are finite state automata.
And every finite state automaton is defined by
the table that describes its transition function $\cal T$ in full,
which is a finite table of expressions
referring to internal states, input symbols, and output symbols.
A table can be converted to a string
just using an additional symbol for the end of line,
and another symbol for the end of cell.
To assure uniqueness, we have to impose some order
on the lines and on the cells. {\QED}

\comment{Comment}
${\frak c}({\cal P}) \in {\frak E}$ is the string of symbols that
represents the Turing machine ${\cal P} \in {\frak T}$.

\clause{Notation}
We will refer to ${\frak p} = {\frak c}({\cal P})$ as a program,
 and to the set of programs as ${\frak P}$.
The set of programs is a proper subset of the set of expressions,
${\frak P} \subset {\frak E}$.

\clause{Definition}\label{programisomorphism}
The program isomorphism is the natural isomorphism that relates
each Turing machine ${\cal P} \in {\frak T}$
with the expression describing it,
 ${\frak c}({\cal P}) = {\frak p} \in {\frak P}$.
That is, ${\frak T} \Leftrightarrow {\frak P}: {\cal P} \leftrightarrow {\frak c}({\cal P})$.

\comment{Comment}
Now, ${\frak T} \cong {\frak P} \subset {\frak E}$.

\clause{Lemma}\label{programsarecountable}
The set of Turing machines is countable, that is,
$|{\frak T}| = |{\bb N}| = {\aleph}_0$.

\comment{Proof}
Proved by \cite Turing (1936), \S5.
Using the program isomorphism, see \ref{programisomorphism},
we order the Turing machines according to its
corresponding program ${\frak p} = {\frak c}({\cal P})$.
We can order the programs,
because they are finite strings of symbols,
for example first by length,
and then those of a given length by some kind of alphabetical order.
Once ordered, we can assign a natural number to each one. {\QED}

\clause{Theorem}\label{computingiscounting}
All computing sets are countable, that is,
$|{\frak T}| =  |{\frak E}| = {\aleph}_0$.

\comment{Proof}
By Lemmas \ref{expressionsarecountable} and \ref{programsarecountable}. {\QED}

\comment{Comment}
All computing is about countable sets. Computing is counting.

\subsection{Turing Completeness}

\clause{Theorem}\label{universalTuringmachine}
There is a Turing machine, called universal Turing machine,
$\cal U$,
that can compute anything that any Turing machine can compute.
That is:
$$\exists\,{\cal U} \in {\frak T} \mid
  \forall{\cal P} \in {\frak T}, \forall{\frak d} \in {\frak E},\quad
  \TM U<c({\cal P})\; d> = \TM P<d> .$$

\comment{Proof}
Proved by \cite Turing (1936), \S6 and \S7. {\QED}

\comment{Comment}
The equality means that
if $\TM P<d> \hookrightarrow {\frak r}$ then 
$\TM U<c({\cal P})\; d> \hookrightarrow {\frak r}$,
and the converse, and also that
if $\TM P<d> \hookrightarrow \infty$ then 
$\TM U<c({\cal P})\; d> \hookrightarrow \infty$,
and the converse.
That is,
$\TM U<c({\cal P})> \equiv_F {\cal P}$.
To complete the definition,
if ${\frak e} \notin {\frak P}$, then
$\TM U<e\,d> \hookrightarrow {\frak e\, d}$.

\clause{Notation}
We will refer to the set of universal Turing machines as $\frak U$.

\comment{Comment}
The set of universal Turing machines is a proper subset of
the set of Turing machines,
${\frak U} \subset {\frak T}$.

\clause{Lemma}\label{universalprogram}
For each universal Turing machine ${\cal U}$
there is a universal program ${\frak u}$.

\comment{Proof}
Universal Turing machines are Turing machines,
and ${\frak u} = {\frak c}({\cal U})$.
Then, by the program isomorphism,
see \ref{programisomorphism},
${\frak u} = {\cal U}$. {\QED}

\comment{Comment}
Given ${\frak u} = {\frak c}({\cal U})$
and ${\frak p} = {\frak c}({\cal P})$, then
$\TM U<p\, d> = \TM P<d>$ and
$\TM U<u\, p\, d> = \TM U<p\, d>$,
so $\frak u$ is the identity for programs, and
$ %\TM U<u\, u\, u\, p\, d> = 
  \TM U<u\, u\, p\, d> = 
  \TM U<u\, p\, d> = 
  \TM U<p\, d> =
  \TM P<d>$.

\clause{Definition}
The terminating condition
 $P_{\sigma}: {\frak T} \to {\bb B}$ is:
$$P_{\sigma}({\cal P}) =
  \cases{\bot & if $\exists\,{\frak w}\in{\frak E},\;
                 \TM P<w> \hookrightarrow \infty$\cr
         \top & otherwise .\cr}$$

\comment{Comment}
A terminating Turing machine always halts.
There are not paradoxes in a terminating Turing machine.
While Turing machines implement
partial functions, ${\frak E} \to {\frak E}^+$,
see \ref{Turingequivalence},
terminating Turing machines implement
total functions, ${\frak E} \to {\frak E}$.

\clause{Definition}\label{terminating}
The terminating problem is $\sigma = p? P_{\sigma}(p)$.\\
The non-terminating problem is
 $\bar{\sigma} = p?\, \neg P_{\sigma}(p)$.

\comment{Comment}
The terminating problem follows from
 the condition isomorphism of problems,
 see \ref{conditionisomorphism},
applied to the terminating condition $P_{\sigma}$.\\
The non-terminating problem is derived
 from the terminating one by negation,
 see \ref{problemcomposition}.

\comment{Comment}
The set of terminating Turing machines is $\Sigma_{\sigma}$,\\
and the set of non-terminating Turing machines is $\Sigma_{\bar{\sigma}}$.

\comment{Proposition}
 $\Sigma_{\sigma}$ and $\Sigma_{\bar{\sigma}}$ are a partition of ${\frak T}$,
because $\Sigma_{\sigma} \cap \Sigma_{\bar{\sigma}} = \emptyset$ and 
         $\Sigma_{\sigma} \cup \Sigma_{\bar{\sigma}} = {\frak T}$.

\clause{Definition}
We will call ${\frak a} = {\frak c}({\cal P}_{\!\sigma}) \in {\frak E}$,
 where ${\cal P}_{\!\sigma} \in \Sigma_{\sigma}$, an algorithm.

\comment{Comment}
$\forall{\frak d},\,
 \TM P_{\!\sigma}<d> \hookrightarrow {\frak r} \neq \infty
 \;\Leftrightarrow\;
\forall{\frak d},\,
\TM U<a \; d> \hookrightarrow {\frak r} \neq \infty$.

\comment{Comment}
An algorithm is the expression of a computation
that always halts.

\comment{Notation}
We will refer to the set of algorithms as
 ${\frak A}$.

\comment{Comment}
${\frak A} \subset {\frak P} \subset {\frak E}$.

\clause{Lemma}\label{Uisnonterminating}
Universal Turing machines are non-terminating, that is,
${\frak U} \subset  \Sigma_{\bar{\sigma}} \subset {\frak T}$.

\comment{Proof}
Because there are paradoxes in some Turing machines.
For example, for Turing machine $\cal W$,
that has not any $h$ ({\it halt\/}) in its transition table,
every expression is a paradox.
 That is,
$\exists{\cal P}\in{\frak T},\exists{\frak w}\in{\frak E},\;
  \TM P<w> \hookrightarrow \infty 
  \;\Rightarrow\; \forall{\cal U}\in{\frak U},\;
  \TM U<c({\cal P})\; w> \hookrightarrow \infty$.
{\QED}

\comment{Comment}
If expression ${\frak w}$ is a paradox in $\cal P\!$, then
expression ${\frak c({\cal P})\; w}$
is a paradox in ${\cal U}$.
Then, ${\cal U}\in\Sigma_{\bar{\sigma}}$.

\clause{Definition}\label{Turingcomplete}
A computing device is Turing complete if
it can compute whatever any Turing machine can compute.
We will call every Turing complete device
a universal computer.

\comment{Comment}
The prototype of universal computer is a universal Turing machine,
$\cal U$.

\comment{Comment}
The Turing machine, as it was presented by \cite Turing (1936),
models the calculations done by a person. This means that
we can compute whatever any Turing machine can compute
 provided we have enough time and memory,
and therefore we are Turing complete
 provided we have enough time and memory.

\clause{Theorem}\label{universalareequivalent}
All universal computers are equivalent.

\comment{Proof}
G\"odel and Herbrand recursiveness,
Church $\lambda$-definability, and
Turing computability
are equivalent, because
\cite Kleene (1936) showed that
 every recursive function is $\lambda$-definable,
 and the converse,
and then \cite Turing (1937) showed that
 every $\lambda$-definable function is computable, and that
 every computable function is recursive.
{\QED}

\comment{Comment}
A universal Turing machine is equivalent to a $\lambda$-calculus interpreter,
where a $\lambda$-calculus interpreter is a device
that can perform any $\lambda$-calculus reduction.
A  universal Turing machine is equivalent to a mathematician
calculating formally, and without errors, any recursive function.

\comment{Comment}
The universal Turing machine, the $\lambda$-calculus interpreter,
and the mathematician, who is a person, are equal in
computing power. And all of them are Turing complete.

\clause{Proviso}\label{proviso}
Whenever we apply a general statement to
a finite universal computing device, we should add a cautious
`provided it has enough time and memory'.

\comment{Comment}
Although the finite universal computer can perform
each and every step of the computation
exactly the same as the unrestricted universal computer,
 the finite universal computer could meet
 some limitations of time or memory
 that would prevent it to complete the computation.
In that case, the same finite universal computer,
provided with some additional time
 and some more memory,
would perform some more computing steps
 exactly the same as the unrestricted universal computer.
This extension procedure can be repeated as desired
to close the gap between the finite and the unrestricted
universal computer.

\comment{Comment}
We will understand that the proviso
`provided it has enough time and memory'
is implicitly stated whenever we refer to
a finite universal computing device.

\clause{Convention}
Because all universal computers are equivalent,
we can use any of them, let us call the one used $\cal U$,
and then drop every $\cal U$ from the formulas,
and just examine expressions,
that is, elements in ${\frak E}$.
In case we need to note a non-halting computation,
we will use $\infty$.

\comment{Comment}
Using the convention is as if we were always looking
inside the tape of ${\cal U}$.
Given a universal computer, $\cal U$,
computing is about expressions manipulating expressions.

\comment{Example}
Formula
 $\TM U<c({\cal P})\ d> \hookrightarrow {\frak r}$
is reduced to
 $\TM<c({\cal P})\ d> \hookrightarrow {\frak r}$,
and even to
 $\TM<p\ d> \hookrightarrow {\frak r}$,
using the rewriting rule:
 $\forall{\cal P}\in {\frak T},\; {\frak c}({\cal P}) = {\frak p}$.
If the universal computer is a $\lambda$-calculus interpreter,
then this is usually written as the $\beta$-reduction
 $({\frak p}\ {\frak d}) \to {\frak r}$,
 where the left hand side is a $\lambda$-application, and
 ${\frak p}$ is defined by some $\lambda$-abstraction.

\clause{Definition}
For each program ${\frak p} \in {\frak P}$ we define
a function ${\frak F}_{\frak p}: {\frak E} \to {\frak E}^+$,
this way:
$${\frak F}_{\frak p}({\frak e}) = 
\cases{{\frak r} & if $\TM<p\, e> \hookrightarrow {\frak r}$\cr
       \infty    & if $\TM<p\, e> \hookrightarrow \infty$ .\cr}$$

\comment{Comment}
If $\forall{\frak e} \in {\frak E}$,
${\frak F}_{\frak p}({\frak e}) = {\frak F}_{\frak q}({\frak e})$,
then we say that programs $\frak p$ and $\frak q$ are
behaviorally equivalent, ${\frak p} \equiv_{\frak F} {\frak q}$,
or that $\frak p$ and $\frak q$ implement the same function.

\clause{Theorem}
$\forall {\cal P} \in {\frak T}$,
 $F_{\cal P} = {\frak F}_{\frak p}$,
where ${\frak p} = {\frak c}({\cal P})$.

\comment{Proof}
$\forall{\frak d} \in {\frak E}$,
$\forall{\cal P} \in {\frak T}$,
${\frak F}_{\frak p}({\frak d}) = F_{\cal P}({\frak d})$,
 see \ref{Turingequivalence}, because
 $\TM U<c({\cal P})\ d> = \TM P<d>$,
by Theorem~\ref{universalTuringmachine},
and therefore $F_{\cal P} = {\frak F}_{\frak p}$ when
the universal computer is a universal Turing machine, $\cal U$.
Theorem \ref{universalareequivalent}
extends it to every universal computer.
{\QED}

\comment{Comment}
$\cal P$ and $\frak p$ implement the same function.

\comment{Comment}
This theorem is a consequence of the program isomorphism,
 see \ref{programisomorphism}. In other words,
${\frak T}\cong{\frak P}$ implies that
$\mathord{\equiv_F} \leftrightarrow \mathord{\equiv_{\frak F}}$,
so ${\cal P} \equiv_F {\cal Q} \;\Leftrightarrow\;
 {\frak p} \equiv_{\frak F} {\frak q}$.

\comment{Corollary}
$F_{\cal U} = {\frak F}_{\frak u}$,
 where ${\frak u} = {\frak c}({\cal U})$.

\subsection{Turing's Thesis}

\clause{Thesis}\label{Turingsthesis}
What is effectively calculable is computable.

\comment{Comment}
This is Church's thesis, or rather Turing's thesis,
 as it was expressed by \cite Gandy (1980).
There, `something is effectively calculable'
if its results can be found
by some purely mechanical process,
see \ref{effectivelycalculable},
and `computable' means that the same results
will be found by some Turing machine.
Then, $^*{\bb F} \subseteq {\frak T}$.

\comment{Comment}
`What is computable is effectively calculable',
or  ${\frak T} \subseteq {}^*{\bb F}$,
is the converse of Turing's thesis.
And it is obvious that if a Turing machine
can compute a function, then
the function is effectively calculable,
see \ref{effectivelycalculable},
 by a Turing machine.
Therefore, $^*{\bb F} = {\frak T}$,
and $|^*{\bb F}| = \aleph_0 $, by \ref{programsarecountable}.

\clause{Remark}
An effectively calculable function is not an input to output mapping;
it is a process to calculate the output from the input.

\comment{Example}
To multiply a number expressed in binary by two
we can append a `0' to it, which is
an effectively calculable function that we will call {\it app0}.
But the complete memoization of the same function,
which we will call {\it memoby2}, is not effectively calculable
because it would require an infinite quantity of memory.
And therefore, $\hbox{\it app0} \neq \hbox{\it memoby2}$.

\clause{Notation}\label{Turinguniverse}
We will call
every universe where the Turing's thesis is true
a Turing universe.
When we want to note that
something is true in a Turing universe,
we will use an asterisk,
so $A\T= B$ means that $A=B$ if the Turing's thesis stands.

\comment{Examples}
$^*{\bb F} \T= {\frak T}$ and $|^*{\bb F}| \T= {\aleph}_0$.

\comment{Comment}
The Turing's thesis affirms that this is a Turing universe.
In any Turing universe
the Turing's thesis is a law of nature,
as it was defended by \cite Post (1936), last paragraph.
Then a Turing universe can also be called a Post universe.

\comment{Comment}
While the Turing's thesis is true, you can ignore the asterisks.

\clause{Theorem}\label{maximumcomputing}
Universal computers are* the most capable computing devices.

\comment{Proof}
If Turing's thesis stands, see \ref{Turingsthesis}, then
anything that any mechanism can effectively calculate
can be computed by some Turing machine, and therefore,
 by Theorem~\ref{universalTuringmachine},
it can be computed by any universal Turing machine,
and finally, by Theorem~\ref{universalareequivalent},
it can be computed by any universal computer.
{\QED}

\clause{Lemma}\label{computinglimits}
There are definable functions that no Turing machine can compute.

\comment{Proof}
You can use a diagonal argument,
or work from other theorems that use the diagonal argument.
For example,
the set of Turing machines is countable,
 see \ref{programsarecountable}, $|{\frak T}| = |{\bb N}| = {\aleph}_0$, 
while the possible number of predicates
on natural numbers, that is,
the number of functions ${\bb N} \to {\bb B}$,
is $2^{|{\bb N}|} = 2^{{\aleph}_0}$, which is not countable,
$|{\frak T}| = |{\bb N}| = {\aleph}_0 <  2^{{\aleph}_0} = 2^{|{\bb N}|}$.
This uses Cantor's theorem, $|S| < |2^S|$,
with its diagonal argument.
So there are not enough Turing machines
to compute every definable function.
{\QED}

\comment{Corollary}
Universal computers cannot compute every definable function.

\comment{Comment}
If the Turing's thesis stands, see \ref{Turingsthesis},
then it follows that
there are definable functions that are not effectively calculable,
see \ref{effectivelycalculable}.

\comment{Comment}
There are* more mappings than processes.

\clause{Definition}
The identity Turing machine, $\cal I$, just halts.

\comment{Comment}
It does nearly nothing. But, wait!

\clause{Lemma}\label{identityTuringmachine}
$\forall{\frak x}\in{\frak E},\; \TM I<x> \hookrightarrow {\frak x}$,
where $\cal I$ is the identity Turing machine.

\comment{Proof}
Whatever expression ${\frak x}\in{\frak E}$
 is written on the tape of $\cal I$,
that very same expression ${\frak x}$ is written
 when $\cal I$ halts,
because halting is all what $\cal I$ does.
{\QED}

\comment{Comment}
$\cal I$ does not touch the expression.

\clause{Lemma}
The identity Turing machine is terminating,
that is, ${\cal I} \in \Sigma_{\sigma}$.

\comment{Proof}
The identity Turing machine, which just halts,
is terminating, see \ref{terminating},
because it always halts; it only halts.
{\QED}

\comment{Comment}
$\cal I$ behaves, because sometimes
`you can look, but you better not touch'.

\clause{Lemma}\label{identityequivalence}
The identity Turing machine $\cal I: {\frak E} \to {\frak E}$
is* the identity function $i: {\bb S} \to {\bb S}$
such that $\forall x\in{\bb S}$, $i(x) = x$,
that is, ${\cal I} \T= i$.

\comment{Proof}
The identity function $i$
is an effectively calculable function,
 see \ref{effectivelycalculable}.
Therefore,
if the Turing's thesis stands, see \ref{Turingsthesis},
then there is a Turing machine ${\cal J}$ such that 
$\forall{\frak x}\in{\frak E},\;
 \TM J<x> \hookrightarrow {\frak x}$.
By Lemma~\ref{identityTuringmachine},
that Turing machine ${\cal J}$ is the
identity Turing machine ${\cal I}$.
{\QED}

\comment{Comment}
If ${\frak i} = {\frak c}({\cal I})$, then
$\TM U<i\,p\,d> = \TM I<p\,d> \hookrightarrow {\frak p\,d}$, and
$\TM U<u\,p\,d> = \TM U<p\,d> = \TM P<d> \hookrightarrow {\frak r}$, or $\infty$,
see \ref{universalprogram}.
Then $\frak i$ is the literal identity for expressions, or quotation,
and $\frak u$ is the functional identity for programs, or evaluation. Both are computable,
but ${\cal I}\in\Sigma_\sigma$ and ${\cal U}\in\Sigma_{\bar\sigma}$,
see \ref{Uisnonterminating}.

\clause{Theorem}\label{allisexpression}
Everything is* an expression, that is, ${\frak E} \T= {\bb S}$.

\comment{Proof}
${\bb S}$ is the set of everything, see \ref{everythingisin}.
In computing, there are only computing devices, $\frak T$,
and expressions, $\frak E$, see \ref{computingsets}.
But then, by the program isomorphism,
see \ref{programisomorphism},
computing devices are expressions, ${\frak T} \subset {\frak E}$.
Therefore, in computing everything is an expression.
And now, 
if the Turing's thesis stands, see \ref{Turingsthesis}, then
Lemma~\ref{identityequivalence} also stands, so
$\forall x\in{\bb S},\,
 x=i(x) \T= \TM I<{\mit x}> \hookrightarrow x \in {\frak E}$.
The converse,
$\forall{\frak x}\in{\frak E},\, {\frak x}\hookleftarrow\TM I<x>=
 i({\frak x})={\frak x} \in {\bb S}$,
holds irrespective of Turing's thesis.
Therefore, ${\bb S} \T= {\frak E}$.
{\QED}

\comment{Comment}
We will write $\frak x$ to indicate a computing point
of view of $x$, but $\forall x,\,x \T= {\frak x}$.
For example, $i \T= {\frak i}$.

\clause{Lemma}\label{solutionsarecountable}
The set of solutions ${\bb S}$ is* countable,
 that is, $|{\bb S}| \T= \aleph_0$.

\comment{Proof}
${\bb S} \T= {\frak E}$, by \ref{allisexpression},  and
$|{\frak E}| = |{\bb N}| = {\aleph}_0$, by \ref{expressionsarecountable},
therefore $|{\bb S}| \T= |{\frak E}| = {\aleph}_0$.
{\QED}

\comment{Comment}
We will refer to 
the set of solutions in a Turing universe as ${\bb S}^*$. 
So we can also write this lemma as $|{\bb S}^*| = {\aleph}_0$.

\clause{Theorem}\label{resolvingiscomputing}
Resolving is* computing, that is, ${\frak T} \T= {\bb R}$.

\comment{Proof}
From Theorem~\ref{allisexpression},
everything is* an expression, %${\bb S} \T= {\frak E}$,
and taking transitions and not states,
it follows that
whatever transforms expressions in computing theory,
 that is, a Turing machine ${\cal P}$,
 or its equivalent program ${\frak p}$,
 or a $\lambda$-function of the $\lambda$-calculus,
is* equivalent to
 whatever transforms sets in set theory,
  that is, an effectively calculable function,
and it is* also equivalent to
 whatever transforms problems in problem theory,
  that is, a resolution $\Re$.
Therefore, resolving is* computing, ${\bb R} \T= {\frak T}$.
{\QED}

\comment{Comment}
${}^*\!f \T\equiv {\cal P} \equiv {\frak p} \T\equiv {\Re}$,
and ${}^*{\bb F} \T= {\frak T} \T=  {\bb R}$.

\comment{Comment}
We can define functions that are not effectively calculable,
see \ref{computinglimits}.
Those functions that cannot effectively calculate,
cannot therefore transform, and they are,
in this sense, useless;
we can define them, but we cannot use them.

\comment{Corollary}
Metasolutions are* effectively calculable functions,
that is, $\Pi{\bb S} \T= {}^*{\bb F}$.

\comment{Proof}
Because ${\bb R} = \Pi{\bb S}$,
see \ref{metaproblem}, so
$\Pi{\bb S} = {\bb R} \T= {}^*{\bb F}$.
{\QED}

\clause{Lemma}\label{resolutionsarecountable}
The set of resolutions ${\bb R}$ is* countable,
 that is, $|{\bb R}| \T= \aleph_0$.

\comment{Proof}
${\bb R}\T={\frak T}$, by \ref{resolvingiscomputing}, and
$|{\frak T}| = |{\bb N}| = {\aleph}_0$, by \ref{programsarecountable},
therefore $|{\bb R}| \T= |{\frak T}|= {\aleph}_0$.
{\QED}

\comment{Comment}
We will refer to 
the set of resolutions in a Turing universe as ${\bb R}^*$. 
So we can also write this lemma as $|{\bb R}^*| = {\aleph}_0$.

\clause{Lemma}\label{deltaiscalculable}
Predicate $P_{\delta_s}$,
 where $P_{\delta_s}\!(x) = [x=s]$,
is* effectively calculable.

\comment{Proof}
Both $s$ and $x$ are* expressions, by \ref{allisexpression},
so both are finite strings of symbols,
$s = s_1 s_2 \dots s_n$, and $x = x_1 x_2 \dots x_m$.
Then we can define a Turing machine with $n+2$ states, that
starts in state $1$, and that
when some string $x$ is written on its tape,
it scans the string $x$, symbol by symbol, from the leftest one, this way:
1) in state $i$, with $1\leq i\leq n$,
 if the read symbol is $s_i$, then
  it writes a {\it blank},
  goes to state $i+1$, and
  moves to the {\it right},
 but if the read symbol is not $s_i$, then
  it writes a {\it blank},
  goes to state $0$, and
  moves to the {\it right};
2) in state $n+1$,
 if the read symbol is {\it blank}, then
  it writes a $\top$,
  goes to state $0$, and
  {\it halt\/}s,
 but if the read symbol is not {\it blank}, then
  it writes a {\it blank},
  goes to state $0$, and
  moves to the {\it right};
3) in state $0$,
 if the read symbol is not {\it blank}, then
  it writes a {\it blank},
  goes to state $0$, and
  moves to the {\it right},
 but if the read symbol is {\it blank}, then
  it writes a $\bot$,
  goes to state $0$, and
  {\it halt\/}s.
This Turing machine implements $P_{\delta_s}$,
and therefore $P_{\delta_s}$ is computable.
{\QED}

\comment{Corollary}
Problem $\delta_s = x? [x=s]$ is* expressible.

\comment{Proof}
Because problem $\delta_s$ condition $P_{\delta_s}$
is* effectively calculable, see \ref{expressible}.
{\QED}

\comment{Comment}
Problem $\delta_s$ is used
in the proof of Theorem~\ref{everythingisin}.

\comment{Corollary}
The only solution to problem $\delta_s$ is $s$,
so $\Sigma_{\delta_s} = \{s\} \in {\bb S}^1$.

\comment{Proof}
Because
$\Sigma_{\delta_s} = \{\, x \mid P_{\delta_s}\!(x) \,\} =
\{\, x \mid [x=s] \,\} = \{s\}$.
{\QED}

\clause{Lemma}\label{problemsarecountable}
The set of problems ${\bb P}$ is* countable,
 that is, $|{\bb P}| \T= \aleph_0$.

\comment{Comment}
We will refer to 
the set of problems in a Turing universe as ${\bb P}^*$.
So we can also write this lemma as $|{\bb P}^*| = {\aleph}_0$.
If the condition of a problem is computable,
then the problem is in ${\bb P}^*$;
 $\delta_s \in {\bb P}^*$, for example.

\comment{Proof}
Problem $\delta_s$ is* expressible,
 see \ref{deltaiscalculable}. Then,
$\delta_{{\bb S}^*} =
 \{\, \delta_s \mid s\in{\bb S}^* \,\} \subseteq {\bb P}^*$
because each $\delta_s \in {\bb P}^*$,
and $|\delta_{{\bb S}^*}| = |{\bb S}^*|$ because there is a bijection
$\delta_{{\bb S}^*} \Leftrightarrow {\bb S}^* : \delta_s \leftrightarrow s$.
Also, by Theorem~\ref{allisexpression}, ${\bb P}^* \subseteq {\frak E}$.
Therefore,
$\delta_{{\bb S}^*} \subseteq {\bb P}^* \subseteq {\frak E}$, and
$|\delta_{{\bb S}^*}| = |{\bb S}^*| = {\aleph}_0 = |{\frak E}|$, 
and then, by the Cantor-Bernstein-Schr\"oder theorem,
$|{\bb P}^*| = \aleph_0$.
{\QED}

\comment{Comment}
The Cantor-Bernstein-Schr\"oder theorem is
Theorem~B of \S2, page 484, in \cite Cantor (1895).
We have really used the equivalent Theorem~C,
in the same page.

\clause{Theorem}\label{resolvingiscounting}
All problem sets are* countable, that is,
$|{\bb S}^*| = |{\bb P}^*| = |{\bb R}^*| = \aleph_0$.

\comment{Proof}
By Lemmas
\ref{solutionsarecountable},
\ref{problemsarecountable}, and
\ref{resolutionsarecountable}.
{\QED}

\subsection{Full Resolution Machine}

\clause{Definition}
A full resolution machine is a device that can execute any resolution.

\clause{Theorem}\label{syntaxengineisuniversal}
A full resolution machine is* a Turing complete device.

\comment{Proof}
By Theorem~\ref{resolvingiscomputing},
resolving is* computing, ${\Re} \T\equiv {\cal P}$.
This means that to achieve the maximum resolving power
is* to achieve the maximum computing power,
which is* the computing power of a universal computer,
by Theorem~\ref{maximumcomputing}.
Therefore, in a Turing universe
a full resolution machine has to be Turing complete.
{\QED}

\comment{Comment}
To execute any resolution ${\Re}: {\bb P} \to 2^{\bb S}$,
the full resolution machine has
to calculate functions that
can take functions and that can return functions
without limitations, as
\lower 2pt\hbox{$2^{2^{\bb S}\to2^{\bb S}}\to2^{2^{\bb S}\to2^{\bb S}}$}
for meta-analogies, see \ref{metadiagrams}. Then
a full resolution machine has to execute every possible function,
and therefore, in a Turing universe,
it has to execute every computable function, and then
it has to be a $\lambda$-calculus interpreter,
or an equivalent computing device, for example $\cal U$.

\comment{Comment}
This means that
problem resolving is* equal to computing, and then
full problem resolving is* equal to universal computing.

\comment{Corollary}
A full resolution machine* is a universal computer.

\comment{Proof}
Because a full resolution machine is* a universal computer.
{\QED}

\comment{Comment}
Now we will state two equivalences between
computing theory and problem theory concepts
that are true in any Turing universe,
and that are needed to show the limitations
of full resolution machines.

\clause{Definition}\label{recursivelyenumerable}
A set is recursively enumerable
if there is a Turing machine that
generates all of its members, and then halts.

\comment{Comment}
If the set is infinite, the Turing machine
will keep generating its members forever.

\comment{Definition} A set is computable if it is
recursively enumerable.

\clause{Theorem}\label{resolvableisrecursivelyenumerable}
Resolvable in problem theory
is* equivalent to
recursively enumerable in computing theory, that is,
$$\hbox{Resolvable}\T=\hbox{Recursively Enumerable}\;.$$

\comment{Proof}
To see that a problem is resolvable if, and only if,
the set of its solutions is recursively enumerable,
just compare the definition of
 resolvable problem, in \ref{resolvable},
with the definition of
 recursively enumerable set, in \ref{recursivelyenumerable}.
The only remaining gap is to equate
the valid resolution $\Re$ of the resolvable problem to
the Turing machine of the recursively enumerable set,
 a gap that we can bridge with the help of
 Theorem~\ref{resolvingiscomputing}.
Finally see that, by the set isomorphism, see \ref{setequality},
we can refer interchangeably
 to the problem $\pi$ or to the set of its solutions $\Sigma_\pi$.
Then we can say that a problem is recursively enumerable,
or that a set is resolvable.
{\QED}

\clause{Definition}\label{recursive}
A set is recursive if its characteristic function
can be computed by a Turing machine that always halts.

\clause{Theorem}\label{expressibleisrecursive}
Expressible in problem theory
is* equivalent to
recursive in computing theory, that is,
$$\hbox{Expressible}\T=\hbox{Recursive}\;.$$

\comment{Proof}
The condition of a problem, $P_\pi$,
is the characteristic function of the
set of its solutions,
because $\Sigma_\pi = \{\, s \mid P_\pi(s) \,\}$,
 see \ref{setofsolutions}.
Then, if the set of all the solutions to a problem 
is a recursive set, see~\ref{recursive},
then the condition $P_\pi$ can be computed by
a Turing machine that always halts.
So the condition $P_\pi$ is an effectively calculable function,
and therefore the problem is expressible,
see \ref{expressible}.
If  Turing's thesis, \ref{Turingsthesis}, is true,
then the converse is also true;
just go backwards from expressible to recursive.
Finally, by the set isomorphism, see \ref{setequality},
we can refer interchangeably
 to the problem $\pi$ or to the set of its solutions $\Sigma_\pi$.
Then, we can say that a problem is recursive,
or that a set is expressible.
{\QED}

\clause{Lemma}
The limitations of full resolution machines
are* the limitations of universal computers.

\comment{Proof}
Because a full resolution machine is* a universal computer,
see \ref{syntaxengineisuniversal}.
{\QED}

\comment{Comment}
Even if universal computers are the most capable computers,
they cannot compute everything,
see \ref{computinglimits}.
Now we will present three limits related to problems.

\clause{Lemma}\label{cannotexpress}
A full resolution machine can* execute any resolution,
but it cannot* express some problems.

\comment{Proof}
There is a recursively enumerable set that is not recursive;
this is the last theorem in \cite Post (1944) \S1.
Translating,
 by Theorems \ref{resolvableisrecursivelyenumerable}
 and \ref{expressibleisrecursive},
 to problem theory:
there is* a resolvable problem that is* not expressible.
{\QED}

\comment{Comment}
This is the problem limit of full resolution machines*.

\comment{Comment}
That last theorem in \cite Post (1944) \S1, page 291,
is an abstract form of G\"odel's incompleteness theorem,
see \cite Post (1944) \S2.

\clause{Lemma}\label{cannotresolve}
A full resolution machine can* execute any resolution,
but it cannot* resolve some problems.

\comment{Proof}
Let us call $\kappa$ some problem that
 is resolvable but not expressible, see \ref{cannotexpress}.
This means that
$\exists\Re \mid \Re(\kappa)= \Sigma_{\kappa}$, but
$\mathord{\!\not\,}\exists P_{\kappa} \mid
 P_{\kappa}(x) = [x \in \Sigma_{\kappa}]$.
Note that $|\Sigma_{\kappa}| \geq \aleph_0$,
 because otherwise $\exists P_{\kappa}$.
Then its metaproblem $\Pi\kappa$ is solvable but not resolvable.
$\Pi\kappa$ is solvable because $\kappa$ is resolvable,
 see \ref{solvemetaisresolve},
 or, easier, because $\Re$ is a solution to $\Pi\kappa$.
For $\Pi\kappa$ to be resolvable there should be a
resolution that would find `all the solutions of $\Pi\kappa$',
that is, `all the valid resolutions for $\kappa$'.
But, whenever a possible valid resolution for $\kappa$,
 let us call it $\Re'$,
generates a value not yet generated by $\Re$, let us call it $z$,
we cannot decide whether $z \in \Sigma_{\kappa}$
 and it will be eventually generated by $\Re$,
or if $z \notin \Sigma_{\kappa}$
 and it will never be  generated by $\Re$;
remember that $\kappa$ is not expressible,
 $\mathord{\!\not\,}\exists P_{\kappa}$.
And, not being able to decide on $z$, we cannot decide
 whether $\Re'$ is a valid resolution for $\kappa$ or not.
{\QED}

\comment{Comment}
This is the resolution limit of full resolution machines*.

\comment{Comment}
Problem $\kappa$ is named after the complete set $K$
of \cite Post (1944), \S3.

\clause{Lemma}\label{cannotsolve}
A full resolution machine can execute any resolution,
but it cannot solve some problems.

\comment{Proof}
Simply because some problems have not any solution,
$\Sigma_\pi=\{\}=\emptyset$.
{\QED}

\comment{Comment}
This is the solution limit of full resolution machines,
which also applies to full resolution machines*.

\comment{Comment}
An unsolvable problem can be resolved
by showing that it has not any solution.
For example,
the decision problem of the halting problem, $\Delta\eta$,
see \ref{decisionhaltingproblem} below,
was resolved unsolvable by \cite Turing (1936), \S8.

\clause{Theorem}\label{threelimits}
A full resolution machine* can execute any resolution,
but it cannot express some problems (problem limit), and
it cannot resolve some problems (resolution limit), and
it cannot solve some problems (solution limit).

\comment{Proof}
By Lemmas \ref{cannotexpress},
\ref{cannotresolve}, and
\ref{cannotsolve}.
{\QED}

\comment{Comment}
Full resolution machines* have limitations on
each of the three main concepts
of the problem theory.
$$ \hbox{Problem} \,\into{\hbox{\quad\rm Resolution\quad}}\,
   \{\,\hbox{Solution}\,\}$$

\subsection{Problem Topology}

\clause{Definition}\label{decisionproblem}
The decision problem of a problem $\pi = x? P_\pi(x)$,
written $\Delta\pi$, is:
$$\Delta\pi = {\cal P}?\; [{\cal P} \in \Sigma_{\sigma}] \land
 [\forall({\frak x} \T= x),\, \TM P<x> \T= P_\pi(x)] .$$

\comment{Comment}
A solution to the decision problem $\Delta\pi$
 of some original problem $\pi$
is a Turing machine $\cal P$ that always halts and that
computes the original problem condition $P_\pi$ for any input.
Decision problems are only defined in Turing universes,
where ${\frak x} \T= x$ by Theorem~\ref{allisexpression}.

\comment{Comment}
This definition follows \cite Post (1944), page 287.
% By the {\it decision problem\/} of a given set of positive integers
% we mean the problem of effectively determining for an arbitrarily
% given positive integer whether it is, or is not, in the set.

\clause{Definition}
The halting condition
 $P_{\eta}: {\frak T} \times {\frak E} \to {\bb B}$ is:
$$%\belowdisplayskip=6pt minus 3pt
 P_{\eta}({\cal P},{\frak d}) =
  \cases{\bot & if $\TM P<d> \hookrightarrow \infty$\cr
         \top & otherwise .\cr}$$

\clause{Definition}
The halting problem is $\eta = (p,d)? P_{\eta}(p,d)$.

\comment{Comment}
The halting problem $\eta$ corresponds to the halting condition $P_\eta$
by the condition isomorphism of problems,
 see \ref{conditionisomorphism}.

\comment{Comment}
$P_\sigma(p) = \bigwedge_{d\in{\frak E}} P_{\eta}(p,d)$,
 see \ref{terminating},
so $\sigma = \bigwedge_{d\in{\frak E}} \eta$,
 by \ref{conditionisomorphism} and \ref{problemcomposition}.

\clause{Definition}\label{decisionhaltingproblem}
The decision problem of the halting problem, $\Delta\eta$, is:
$$%\abovedisplayskip=0pt \belowdisplayskip=6pt minus 3pt
 \Delta\eta =
 {\cal H}?\; [{\cal H} \in \Sigma_{\sigma}] \land
   [\forall{\cal P} \in {\frak T},\forall{\frak d} \in {\frak E},\,
   \TM H<c({\cal P})\ d> = P_{\eta}({\cal P},{\frak d})] .$$

\clause{Theorem}\label{decisionhaltingisunsolvable}
The decision problem of the halting problem $\Delta\eta$
has not any solution.

\comment{Proof}
\cite Turing (1936), \S8, resolved that $\Delta\eta$ is unsolvable.
{\QED}

\comment{Comment}
There is not any Turing machine that always halts
and that compute $P_{\eta}$ for each possible input.
There is not any algorithm ${\frak a} \in {\frak A}$
that would compute $P_{\eta}({\frak p},{\frak d})$ for every pair
$({\frak p},{\frak d}) \in {\frak P} \times {\frak E}$.

\clause{Lemma}\label{Turingsolvableisexpressible}
The decision problem $\Delta\pi$ of some problem $\pi$ is solvable if,
and only if, the problem $\pi$ is expressible*,
that is,
$\Delta\pi \hbox{ is solvable} \Leftrightarrow 
 \pi \hbox{ is expressible*}$.

\comment{Proof}
From solvable to expressible.
That the decision problem $\Delta\pi$ is solvable,
see \ref{decisionproblem},
means that there is a Turing machine that
always halts, and that computes $P_\pi$ for each
possible input.
Therefore, $P_\pi$ is effectively calculable,
see \ref{effectivelycalculable}, by a Turing machine,
and then the problem $\pi$ is expressible, see \ref{expressible},
and then it is also expressible*.
Now from expressible to solvable.
If a problem $\pi$ es expressible,
then its condition $P_\pi$ is an effectively calculable function,
see \ref{expressible}.
Then, if the Turing's thesis stands, see \ref{Turingsthesis},
that is, if it is expressible*,
then there is a Turing machine $\cal P$ that can compute $P_\pi$
exactly as the effectively calculable function.
$\cal P$ always halts, because $P_\pi$ is a condition,
so its result is finite.
Therefore, the decision problem $\Delta\pi$ of the problem
has a solution, $\cal P$, and then $\Delta\pi$ is solvable,
see \ref{solvable}.
{\QED}

\comment{Corollary}
The halting problem $\eta$ is not expressible*.

\comment{Proof}
The decision problem of the halting problem, $\Delta\eta$,
is not solvable, see \ref{decisionhaltingisunsolvable},
and then the halting problem $\eta$ is not expressible*.
{\QED}

\comment{Comment}
The halting problem $\eta$ is inexpressible*, but solvable.
While the decision problem of the halting problem $\Delta\eta$ is unsolvable,
the halting problem $\eta$ has many solutions.

\clause{Theorem}
The following equivalences stand:
$$\openup-3pt\eqalign{
 \Delta\pi \hbox{ is solvable} \T\Leftrightarrow&\; \pi \hbox{ is expressible},\cr
    \Pi\pi \hbox{ is solvable}   \Leftrightarrow&\; \pi \hbox{ is resolvable},\cr
       \pi \hbox{ is solvable}   \Leftrightarrow&\; \pi \hbox{ is solvable}.\cr}$$

\comment{Proof}
The last one is trivial, and the other two equivalences
were already proved by Lemmas \ref{Turingsolvableisexpressible}
 and \ref{solvemetaisresolve}.
{\QED}

\clause{Definition}\label{problemsets}
A problem $\pi$ can be:
expressible* ($\cal E$) or not expressible* ($\overline{\cal E}$),
resolvable* ($\cal R$) or not resolvable* ($\overline{\cal R}$), and
solvable ($\cal S$) or not solvable ($\overline{\cal S}$).

\comment{Comment}
An expressible problem is* equivalent to a recursive set,
by Theorem~\ref{expressibleisrecursive},
a resolvable problem is* equivalent to a recursively enumerable set,
by Theorem~\ref{resolvableisrecursivelyenumerable}, and
an unsolvable problem is equivalent to an empty set.

\comment{Comment}
Then $\cal R$ is the set of computable sets,
see \ref{recursivelyenumerable}. 

\comment{Comment}
Not every combination is possible.

\clause{Lemma}\label{ifexpressibleresolvable}
If a problem is expressible*,
then it is resolvable*,
that is, ${\cal E}\subset{\cal R}$.

\comment{Proof}
Because every recursive set is recursively enumerable,
 ${\cal E}\subseteq{\cal R}$.
This is a corollary to the first theorem in \cite Post (1944) \S1.
And ${\cal E}\neq{\cal R}$, see the proof of Lemma~\ref{cannotexpress}.
To translate between sets and problems we use
Theorems \ref{resolvableisrecursivelyenumerable} and
 \ref{expressibleisrecursive}.
{\QED}

\comment{Comment}
The first theorem in \cite Post (1944) \S1, page 290,
states that a set $M$ is recursive if and only if
both the set $M$ and its complement $\overline{M}$ are
recursively enumerable.

\clause{Lemma}\label{ifnotsolvableexpressible}
If a problem is not solvable,
then it is expressible*,
that is, $\overline{\cal S} \subset {\cal E}$.

\comment{Proof}
If a problem $\nu$ is not solvable,
 $\nu \in \overline{\cal S}$,
then $\Sigma_\nu = \{\}$,
see \ref{solvable}.
So $\nu$ is a contradictory problem,
see \ref{tautologycontradiction},
and its condition $P_\nu$ is
 the contradiction $P_{\bar\tau}$, that is,
$\forall x, P_\nu(x) = P_{\bar\tau}(x) = \bot$.
So $P_\nu = P_{\bar\tau}$
is an effectively calculable function,
see \ref{effectivelycalculable},
and therefore $\nu$ is expressible,
see \ref{expressible},
and then expressible*.
And $\overline{\cal S} \neq {\cal E}$,
because $(x? [2x=x^2]) \in {\cal S}\cap{\cal E}$.
{\QED}

\comment{Comment}
Being expressible*, by Lemma~\ref{ifexpressibleresolvable},
$\nu$ is also resolvable*:
$\overline{\cal S} \subset {\cal E} \subset {\cal R}$.

\clause{Theorem}
Regarding expressibility* $\cal E$, resolvability* $\cal R$,
 and solvability $\cal S$,
the topology of the problem space is:
$$\overline{\cal S}\subset{\cal E}\subset{\cal R}\subset{\bb P} \,.$$

\needspace{12\baselineskip}

\comment{Proof}
By Lemmas \ref{ifexpressibleresolvable}
and \ref{ifnotsolvableexpressible}.
As shown in the table,
these lemmas prevent four of the eight combinations,
and the examples show that the other four
do exist.
$$\halign{\hfil$#$\hfil&
          \space\hfil$#$\hfil&
          \space\hfil$#$\hfil&
          \quad#\hfil\cr
\cal E& \cal R& \cal S&  Example \& Comment\cr
\noalign{\hrule\kern3pt}
\top& \top& \top&  $x? [2x = x^2]$\cr
\top& \top& \bot&  $x? [2x = x^2] \land [x > 2]$\cr
\top& \bot& \top&  None, by Lemma~\ref{ifexpressibleresolvable}\cr
\top& \bot& \bot&  None, by Lemma~\ref{ifexpressibleresolvable}\cr
\bot& \top& \top&  $\kappa$, see \ref{cannotresolve}\cr
\bot& \top& \bot&  None, by Lemma~\ref{ifnotsolvableexpressible}\cr
\bot& \bot& \top&  $\Pi\kappa$, see \ref{cannotresolve}\cr
\bot& \bot& \bot&  None, by Lemma~\ref{ifnotsolvableexpressible}\quad{\QED}\cr}$$

\comment{Corollary}
Then,
$\{\, \overline{\cal S},
    {\cal E} \cap {\cal S},
    {\cal R} \cap \overline{\cal E},
    \overline{\cal R} \,\}$
is a partition of ${\bb P}$.

\goodbreak

\comment{Comment}
See below, in \ref{Countability},
that ${\cal E} \T= {\bb P}$, and then ${\bb P}^* \subset \bb P$.
Also ${\cal R} \T= 2^{\bb S}$.

\clause{Definition}
We say that a problem is finite,
if the set of its solutions is finite.
We will refer to the set of finite problems as $\cal F$.
That is,
${\cal F} = \{\, \pi \mid\, |\Sigma_\pi| < \aleph_0 \,\}$.

\clause{Lemma}\label{finiteproblems}
The set of finite problems $\cal F$ is a proper subset of
the set of expressible problems $\cal E$.
The set of not solvable problems $\overline{\cal S}$ is
a proper subset of the set of finite problems $\cal F$.
That is,
$\overline{\cal S} \subset {\cal F} \subset {\cal E}$.

\comment{Proof}
${\cal F}\subset{\cal E}$ because all finite sets are recursive,
but not the converse.
$\overline{\cal S}\subset{\cal F}$ because
 $\forall\nu\in\overline{\cal S},\;
      |\Sigma_\nu| = 0 < \aleph_0$,
but $(x? [2x = x^2]) \in {\cal S}\cap{\cal F}$.
{\QED}

\comment{Proposition}
Including ${\cal F}$, the topology of ${\bb P}$ is:
$\overline{\cal S} \subset {\cal F} \subset
 {\cal E} \subset {\cal R} \subset {\bb P}$.

\comment{Corollary}
The topology
$\overline{\cal S} \subset {\cal F} \subset 
 {\cal E} \subset {\cal R} \subset {\bb P}$
partitions the problem space $\bb P$ into five non-empty places:
$\overline{\cal S}$,
${\cal F} \cap {\cal S}$,
${\cal E} \cap \overline{\cal F}$,
${\cal R} \cap \overline{\cal E}$, and
$\overline{\cal R}$.

\clause{Remark}
The upper part of this topology is further refined by the
so called Turing degrees of unsolvability, that we will
call Turing degrees of inexpressibility.
Turing degree zero, $\bf 0$, corresponds to the first three
places, because ${\cal E} = {\bf 0}$.

\comment{Comment}
Then, $|{\cal E}| = |{\bf 0}| = \aleph_0$,
$|{\cal R}| = \aleph_0$, and
$|{\bb P}| = 2^{\aleph_0} > \aleph_0$.
To complete the cardinalities,
$|\overline{\cal S}| = 1$, so $|{\cal S}| = 2^{\aleph_0}$,
and $|{\cal F}| = \aleph_0$.

\clause{Remark}
Noting ${\cal E}^p$ the set of problems
defined by a condition that can be computed in polynomial time,
and ${\cal R}^p$ the set of problems
that can be resolved in polynomial time, then
${\cal E}^p \subset {\cal E}$ and ${\cal R}^p \subset {\cal R}$.
The so called `P = NP?' question
asks if ${\cal E}^p = {\cal R}^p$,
because ${\rm P} = {\cal E}^p$ and ${\rm NP} = {\cal R}^p$,
% see Cook (2000)
and then it should be called
the `${\cal E}^p = {\cal R}^p?$' question.
See that the general question
`${\cal E} = {\cal R}?$'
was answered negatively by Lemma~\ref{ifexpressibleresolvable},
because ${\cal E} \subset {\cal R}$,
and that ${\cal E}^p \subseteq {\cal R}^p$.

\comment{Comment}
A similar question is
`${\cal E}^p \setminus \{\emptyset\} = {\cal S}^p?$',
where ${\cal S}^p$ is the set of problems that can be solved
in polynomial time, so ${\cal S}^p \subset {\cal S}$.
The corresponding general question is also answered negatively,
because ${\bb P} = {\cal S} \cup \{\emptyset\}$,
so ${\cal E} \setminus \{\emptyset\} \subset {\cal S}$.

\section{Resolvers}

\subsection{Semantics and Syntax}

\clause{Remark}
In this Section~\ref{Resolvers},
we will always be inside a Turing universe,
see \ref{Turinguniverse},
and accordingly we will drop every asterisk.
Though some results do not depend on Turing's thesis,
by now the reader should know when it is the case.

\clause{Definition}
A resolver is a device
that takes problems and returns solutions.

\comment{Comment}
A resolver executes resolutions.

\comment{Comment}
After Theorem~\ref{resolvingiscomputing},
we can equate a resolution $\Re \in \bb R$
to the computing device that
executes the resolution ${\cal P} \in {\frak T}$,
that is, $\Re = {\cal P}$.

\clause{Definition}
We will call the domain of ${\bb S}$ semantics.
We will call the domain of ${\bb S}\to{\bb S}$ syntax.

\comment{Comment}
As $\lambda$-calculus shows,
we only need functions to implement any syntax.

\comment{Comment}
By Theorem~\ref{everythingisin},
everything is in ${\bb S}$, including ${\bb S}\to{\bb S}$.
But this is both mathematically impossible,
by Cantor's theorem,
and practically not interesting.

\comment{Example}
Using a practical example,
if the problem is the survival problem,
so some behaviors keep the resolver alive,
and the rest cause the death of the resolver,
then ${\bb S}$ is the set of behaviors,
and it does not include anything that is not a behavior,
not even predicates on behaviors, nor functions.
Note that the condition of the survival problem,
which is satisfied if the resolver does not die,
is a predicate on behaviors.

\clause{Remark}
In this Section~\ref{Resolvers},
we will assume that ${\bb S}$ is not the set of everything,
and, in particular, we will assume that there is not
any function in ${\bb S}$.
We will focus on the survival problem,
and then assume that ${\bb S}$ is the set of behaviors,
or finite state automata,
but you can think that ${\bb S}={\bb N}$,
or any other countable set,
see \ref{solutionsarecountable}.
Then we will build a series of resolvers,
 from the simplest one implementing one element of ${\bb S}$,
 to more complex resolvers that have to implement functions
 in order to look for resolutions to deal with metaproblems.

\clause{Definition}
A problem type, for example ${\bb P}_{\Psi}$,
is a subset of the set of problems,
that is, ${\bb P}_{\Psi} \subseteq {\bb P}$.
We will note ${\bb S}_{\Psi}$ the set of the solutions
to the type of problems ${\bb P}_{\Psi}$.
That is,
$\forall\pi_{\Psi}\in{\bb P}_{\Psi}$,
$\Sigma_{\pi_{\Psi}} \subseteq {\bb S}_{\Psi} \subseteq {\bb S}$.

\comment{Comment}
The survival problem is not a single problem,
but a type of problems, ${\bb P}_{\Omega}$;
each living being faces a different survival problem.
But, in this case as in many others,
what it is certain is that the solutions to any
of these problems is of a specific kind.
For example, while eating can be a solution,
imagining how to eat is not a solution,
even though it can help us to get something to eat,
because it can be a metasolution.
Then ${\bb S}_{\Omega}$ is the set of behaviors.

\clause{Remark}
Metaproblems $\Pi\pi$ are a type of problem,
 $\Pi{\bb P} = {\bb P}_\Pi$,
and its solutions are resolutions,
 $\Pi{\bb S} = {\bb S}_\Pi = {\bb R}$,
see \ref{metaproblem}.

\clause{Lemma}\label{iffiniteresolvable}
If the set of the solutions to some type of problem is finite,
$0<|{\bb S}_{\Gamma}|<{\aleph_0}$,
then each and every problem of that type is 
expressible and resolvable.

\comment{Proof}
Because those problems are in $\cal F$,
so Lemma~\ref{finiteproblems} apply.
{\QED}

\comment{Comment}
If $0<|{\bb S}_{\Gamma}|=N<{\aleph_0}$, then
$|{\bb P}_{\Gamma}|=2^N<{\aleph_0}$ and
 $|{\bb R}_{\Gamma}|=(2^N)^{2^N}<{\aleph_0}$.
In the finite case,
$|{\bb S}_{\Gamma}| < |{\bb P}_{\Gamma}| <
 |{\bb R}_{\Gamma}| < {\aleph_0}$.

\clause{Definition}\label{constantfunction}
A constant function $K_s:{\bb S}\to{\bb S}$ is: %such as
$\forall s\in{\bb S},\forall x\in{\bb S},\; K_s(x) = s$.

\comment{Comment}
Every constant function $K_s$ is effectively calculable,
see \ref{effectivelycalculable}.
They are $\lambda$-definable;
in $\lambda$-calculus, $K = (\lambda sx.s)$.
This is because our $\lambda$-calculus includes the $K$ combinator,
and so we refer to the $\lambda K$-calculus simply as $\lambda$-calculus.

\comment{Special cases}
Tautology: $K_\top = P_\tau$.
Contradiction: $K_\bot = P_{\bar\tau}$.
See \ref{tautologycontradiction}.

\clause{Definition}
The constant isomorphism is the natural isomorphism
 between ${\bb S}$ and the set of constant functions ${\bb K}$
 that relates each $s\in{\bb S}$ with $K_s\in({\bb S}\to{\bb S})$.
That is, ${\bb S}  \Leftrightarrow {\bb K} : s \leftrightarrow K_s$.

\comment{Comment}
We can extend any operation on $\bb S$ to $\bb K$.
For example, for any binary operation $*$ on $\bb S$, we define
$\forall x,\, [K_a * K_b](x) = K_a(x) * K_b(x)= a*b = K_{a*b}(x)$.

\comment{Comment}
Semantics is included in syntax,
that is, ${\bb S} \cong {\bb K} \subset ({\bb S}\to{\bb S})$.

\clause{Remark}\label{semanticfunction}
A semantic function $f:{\bb S}\to{\bb S}$ is a syntactic element,
$f\in({\bb S}\to{\bb S})$,
but it is not a syntactic function
${\frak f} \in (({\bb S}\to{\bb S})\to({\bb S}\to{\bb S}))$,
because the semantic function $f$ takes semantic elements
and returns semantic elements,
while, using the constant isomorphism,
the syntactic function $\frak f$ is not restricted.
In particular,
a semantic function cannot take a function, and
a semantic function cannot return a function. 

\comment{Comment}
In semantics, literal identity $i$ is the identity,
see \ref{identityequivalence},
because there are not higher order functions in semantics.
But, different syntactic objects can refer to the
same semantic object, as in $f(x)=y$,
which means that $f(x)$ and $y$ are two syntactic objects that
refer to the same semantic object.
Then, there are two identities in syntax:
literal identity $\frak i$, or quotation,
which is the semantic function that just returns what it takes,
and functional identity $\frak u$, or evaluation,
which is the syntactic function that follows the references
and returns the final one,
see \ref{universalprogram}.
Note also that a syntactic object can refer to
no semantic object, and then we say that
the syntactic object is a paradox.

\clause{Definition}\label{range}
The range of a resolver $\Re$, noted $\Xi\Re$,
is the set of the problems for which $\Re$ provides
a non-empty subset of solutions, and only of solutions, that is,
$\Xi\Re = \{\, \pi \mid \Re(\pi) \!\subseteq\! \Sigma_\pi \land
 \Re(\pi) \!\neq\! \emptyset \,\}$.

\comment{Comment}
The range of a resolver is
the set of the problems that the resolver solves.

\clause{Definition}\label{power}
The power of a resolver $\Re$, noted $\Phi{\Re}$,
is the set of the problems that the resolver $\Re$ resolves,
that is, $\Phi{\Re} = \{\, \pi \mid \Re(\pi) = \Sigma_\pi \,\}$.

\comment{Comment}
In practice, if $|\Sigma_\pi| > 1$,
it is not sensible to generate all the solutions, $\Sigma_\pi$,
when just one solution solves the problem.
In these cases the range of the resolver is more
important than its power.

\clause{Theorem}
$\forall \Re$, ${\cal S}\cap \Phi\Re \subseteq \Xi\Re$.

\comment{Proof}
Because $\forall\pi\in \Phi\Re \cap {\cal S}$,
we have that
$\pi\in \Phi\Re$ so $\Re(\pi) = \Sigma_\pi$,
see \ref{power},
and then $\Re(\pi) \subseteq \Sigma_\pi$,
and also that $\pi\in {\cal S}$,
 so $|\Sigma_\pi| > 0 \Leftrightarrow \Sigma_\pi \neq \emptyset$,
see \ref{problemsets} and \ref{solvable},
and then
$\Re(\pi) = \Sigma_\pi \neq \emptyset$,
and therefore $\pi \in \Xi\Re$,
see \ref{range}.
{\QED}

\comment{Comment}
For solvable problems, $\Phi\Re \subseteq \Xi\Re$,
so they are easier to solve than to resolve.
But unsolvable problems, some of them resolved, are impossible to solve!

\clause{Definition}
We will say that
the resources of a resolver are in a set
if the capability implemented in the resolver
belongs to that set.

\comment{Comment}
Now we will construct a series of resolvers $\Re_n$,
from the minimal one that only implements one solution,
and then growing naturally step by step.
Each resolver will implement just one element
out of its resources

\comment{Notation}
We will use $\generic{\Re_n}$ to refer to the set of
all the resolvers of step $n$.

\subsection{Mechanism}

\clause{Definition}
A mechanism $\Re_0$ is any resolver that
 implements one member of $\bb S$.
We will note $\Re_0[s]$, where $s \in {\bb S}$,
the mechanism that implements $s$,
 that is, $\Re_0[s] = s \in {\bb S}$.
Then the mechanisms resources are in $\bb S$, and
$\generic{\Re_0} = {\bb S}$.

\comment{Comment}
Mechanism $\Re_0[s]$ returns $s$ unconditionally.

\comment{Comment}
A mechanism $\Re_0$ implements a semantic unconditional computation.

\clause{Notation}
As resolutions return sets of elements in $\bb S$,
to normalize the situation of mechanisms $\Re_0$
we will use the singleton isomorphism,
see \ref{singletonisomorphism},
and we will write $\Re_0[\{s\}]$ to mean
 the singleton $\{s\}$, that is,
$\Re_0[\{s\}] = \{\Re_0[s]\} = \{s\} \in 2^{\bb S}$.

\clause{Lemma}\label{mechanismrange}
$\forall s \in {\bb S}$,
$\Xi\Re_0[\{s\}] = \{\, \pi \mid P_\pi(s) \,\}$.

\comment{Proof}
Just applying the definition of range, see \ref{range},
to the definition of mechanism, we get:
$\Xi\Re_0[\{s\}] =
 \{\, \pi \mid \Re_0[\{s\}] \!\subseteq\! \Sigma_\pi 
      \land \Re_0[\{s\}] \!\neq\! \emptyset \,\} =
 \{\, \pi \mid \{s\} \!\subseteq\! \Sigma_\pi 
      \land \{s\} \!\neq\! \emptyset \,\} =
 \{\, \pi \mid s \!\in\! \Sigma_\pi \land \top \,\} =
 \{\, \pi \mid s \in \Sigma_\pi \,\} =
 \{\, \pi \mid P_\pi(s) \,\}$.
{\QED}

\comment{Comment}
The range of the mechanism $\Re_0[s]$ is the set
of problems for which $s$ is a solution. 

\clause{Lemma}\label{mechanismpower}
$\forall s \in {\bb S}$,
$\Phi\Re_0[\{s\}] = \{ \delta_s \}$.

\comment{Proof}
Just applying the definition of power, see \ref{power},
to the definition of mechanism, we get:
$\Phi\Re_0[\{s\}] =
 \{\, \pi \mid \Re_0[\{s\}] = \Sigma_\pi \,\} =
 \{\, \pi \mid \{s\} = \Sigma_\pi \,\} =
 \{ \delta_s \}$, the last equation because
$\Sigma_{\delta_s} = \{s\}$,
 see \ref{deltaiscalculable}.
{\QED}

\comment{Comment}
Mechanism $\Re_0[s]$ only resolves problem $\delta_s$.

\clause{Lemma}\label{routinemechanism}
Any singleton routine resolution $R_\pi = \{s\}$ can be implemented by
 the mechanism $\Re_0[R_\pi]$.

\comment{Proof}
If $R_\pi = \{s\}$, then
$R_\pi = \{s\} = \{\Re_0[s]\} = \Re_0[\{s\}] = \Re_0[R_\pi]$.
{\QED}

\comment{Comment}
In theory, we can equal any finite routine resolution
to a union of a finite number of mechanisms,
$R_\pi = \Sigma_\pi =
         \bigcup_{s\in\Sigma_\pi} \{s\} =
         \bigcup_{s\in\Sigma_\pi} \{\Re_0[s]\}$.

\clause{Summary}
In practice, it only makes sense to implement one solution,
as $\Re_0[s]$ does.
Without conditional calculations, the mechanism
could not control when to apply one result or any of the others,
so it would gain nothing implementing more than one.

\comment{Comment}
The mechanism is
a body capable of one behavior.

\comment{Example}
A mechanism can only survive in a specific and very stable environment,
as it is the case of some extremophile archaea.

\subsection{Adapter}

\clause{Definition}
An adapter $\Re_1$ is any resolver that implements
one condition on the members of $\bb S$.
We will note $\Re_1[P_S]$ the adapter that implements $P_S$,
 where $P_S \in ({\bb S}\to{\bb B})$,
 that is, $\Re_1[P_S] = P_S \in ({\bb S}\to{\bb B})$.
Then the adapters resources are in ${\bb S}\to{\bb B}$,
and $\generic{\Re_1}=({\bb S}\to{\bb B})$.

\comment{Comment}
An adapter $\Re_1$ implements a semantic conditional computation.

\clause{Lemma}
Each adapter $\Re_1[P_S]$ implements 
one set of elements of $\bb S$.

\comment{Proof}
Because every predicate $P_S$ defines a set
$S = \{\, s\in{\bb S} \mid P_S(s) \,\} \in 2^{\bb S}$.
The condition $P_S$ is the characteristic function of $S$,
$\forall s\in{\bb S},\; P_S(s) = [s \in S]$.
{\QED}

\comment{Comment}
We will write $\Re_1[P_S] = \Re_1[S] = S \in 2^{\bb S}$.
Only effectively calculable conditions are implementable,
and then adapters can only implement expressible, or recursive,
sets, $\cal E$. Then,
$\Re_1[P_S] = \Re_1[S] = S \in {\cal E}$.

\clause{Lemma}\label{mechanismsareadapters}
Every mechanism $\Re_0$ is an adapter $\Re_1$, that is,
$\generic{\Re_0} \subset \generic{\Re_1}$.

\comment{Proof}
For each mechanism $\Re_0[s]$, which implements $s\in{\bb S}$,
there is an adapter $\Re_1[P_{\delta_s}]$,
see \ref{deltaiscalculable}, that implements the 
singleton $\{s\}\in({\bb S}\to{\bb B})$.
But not every set is a singleton.
Summarizing,
$\generic{\Re_0} = {\bb S} \subset
 ({\bb S}\to{\bb B}) = \generic{\Re_1}$.
{\QED}

\comment{Comment}
In Cantor's paradise, but out of Turing universes,
by the singleton (\ref{singletonisomorphism})
and the set (\ref{setisomorphism}) isomorphisms:
$\generic{\Re_0} = {\bb S} \cong {\bb S}^1 \subset
 2^{\bb S} \cong ({\bb S}\to{\bb B}) = \generic{\Re_1}$.

\clause{Lemma}\label{adapterisunionofmechanisms}
$\Re_1[S] = \bigcup_{s\in S} \{ \Re_0[s] \}$.

\comment{Proof}
Because $\Re_0[s] = s$, so
$\bigcup_{s\in S} \{\Re_0[s]\} = 
 \bigcup_{s\in S} \{s\} = S = \Re_1[S]$.
{\QED}

\comment{Comment}
The results are the same, but not the implementation,
because while
the adapter $\Re_1[S]$ implements a condition,
the union of mechanisms $\bigcup_{s\in S} \{ \Re_0[s] \}$
 works unconditionally.
Thus, the output of the union of mechanisms is independent
of any problem, and then the union cannot implement
$\Re_1[P_S \land P_\pi] = \Re_1[S \cap \Sigma_\pi]$,
for example, so it cannot implement any trial,
see Theorem~\ref{trialisintersection}.

\clause{Lemma}\label{adapterrange}
$\forall S \in 2^{\bb S}$, $\forall \pi \in {\bb P}$,
$\Xi\Re_1[S \cap \Sigma_\pi] =
 \{\, \pi \mid S\cap \Sigma_\pi \neq \emptyset \,\}$.

\comment{Proof}
Because
$\Xi\Re_1[S\cap \Sigma_\pi] =
 \{\, \pi \mid (\Re_1[S\cap \Sigma_\pi] \subseteq \Sigma_\pi) \land
      (\Re_1[S\cap \Sigma_\pi] \neq \emptyset) \,\} =\\
 \{\, \pi \mid (S\cap \Sigma_\pi \subseteq \Sigma_\pi) \land
      (S\cap \Sigma_\pi \neq \emptyset) \,\} =
 \{\, \pi \mid \top \land (S\cap \Sigma_\pi \neq \emptyset) \,\} =\\
 \{\, \pi \mid S\cap \Sigma_\pi \neq \emptyset \,\}$.
{\QED}

\comment{Comment}
The range of the adapter $\Re_1[S \cap \Sigma_\pi]$
is the set of problems that have any solution in $S$.
The adapter $\Re_1[S \cap \Sigma_\pi]$ solves
any problem such that any of its solutions are in $S$.

\comment{Corollary}
If $S \subset S'$, then $\Xi\Re_1[S \cap \Sigma_\pi] \subset
 \Xi\Re_1[S' \cap \Sigma_\pi]$.

\comment{Proof}
In that case,
if a solution to a problem is in $S$, then it is also in $S'$.
But there are also solutions in $S'$ that are not in $S$.
{\QED}

\clause{Lemma}\label{adaptercondition}
If $s \in S$, then $\forall \pi \in {\bb P}$,
 $\Xi\Re_0[\{s\}] \subseteq \Xi\Re_1[S\cap \Sigma_\pi]$.

\comment{Proof}
By Lemma~\ref{mechanismrange},
$\forall \pi \in \Xi\Re_0[\{s\}]$, $P_\pi(s)$,
 that is, $s \in \Sigma_\pi$,
so, if $s \in S$, then $S\cap \Sigma_\pi \neq \emptyset$,
and therefore $\pi \in \Xi\Re_1[S\cap \Sigma_\pi]$,
by Lemma~\ref{adapterrange}.
{\QED}

\comment{Definition}
We will call $s \in S$ the adapter condition.
If the adapter condition holds,
then the adapter $\Re_1[S\cap \Sigma_\pi]$ solves
any problem that the mechanism $\Re_0[s]$ solves.

\comment{Corollary}
If $\{s\} \subset S$,
then $\Xi\Re_0[\{s\}] \subset \Xi\Re_1[S\cap \Sigma_\pi]$.

\comment{Proof}
Because, if $t \in S$ and $t \neq s$, then
$\delta_t \in \Xi\Re_1[S\cap \Sigma_\pi]$ but
$\delta_t \not\in \Xi\Re_0[\{s\}]$.
{\QED}

\comment{Proposition}
If $\{s\} \subset S$, then
$\Xi\Re_0[\{s\}] \not\subset \Xi\Re_1[S]$.\\
Because $\delta_s \in \Xi\Re_0[\{s\}]$,
but $\delta_s \not\in \Xi\Re_1[S]$.

\clause{Lemma}\label{adapterpower}
$\forall S \in 2^{\bb S}$, $\forall \pi \in {\bb P}$,
$\Phi\Re_1[S \cap \Sigma_\pi] = 2^S$.

\comment{Proof}
Because
$\Phi\Re_1[S \cap \Sigma_\pi] = 
 \{\, \pi \mid \Re_1[S \cap \Sigma_\pi] = \Sigma_\pi \,\} =
 \{\, \pi \mid S \cap \Sigma_\pi = \Sigma_\pi \,\} =\\
 \{\, \pi \mid \Sigma_\pi \subseteq S \,\} = 2^S$,
where the last equality uses the set isomorphism,
 see \ref{setisomorphism}.
{\QED}

\comment{Comment}
The power of the adapter $\Re_1[S \cap \Sigma_\pi]$
is the powerset of $S$.
The adapter $\Re_1[S \cap \Sigma_\pi]$ resolves
any problem such that all of its solutions are in $S$.

\comment{Corollary}
If $S \subset S'$, then $\Phi\Re_1[S \cap \Sigma_\pi] \subset
 \Phi\Re_1[S' \cap \Sigma_\pi]$.

\comment{Proof}
Just because, if $S \subset S'$, then $2^S \subset 2^{S'}$.
{\QED}

\clause{Lemma}
If $s \in S$, then $\forall \pi \in {\bb P}$,
$\Phi\Re_0[\{s\}] \subset \Phi\Re_1[S\cap \Sigma_\pi]$.

\comment{Proof}
Using the set isomorphism, see \ref{setisomorphism},
$\delta_s = \Sigma_{\delta_s} = \{s\}$, and then,
if $s \in S$, 
$\Phi\Re_0[\{s\}] = \{ \delta_s \} = \{\{s\}\}
 \subset 2^S = \Phi\Re_1[S \cap \Sigma_\pi]$,
by Lemmas \ref{mechanismpower} and \ref{adapterpower}.
{\QED}

\comment{Comment}
If the adapter condition holds, $s \in S$,
then the adapter $\Re_1[S\cap \Sigma_\pi]$ resolves
any problem that the mechanism $\Re_0[s]$ resolves,
and more.

\comment{Proposition}
If $\{s\} \subset S$, then
$\Phi\Re_0[\{s\}] \not\subset \Phi\Re_1[S]$,
because $\delta_s \not\in \Phi\Re_1[S] = \{S\}$.

\clause{Lemma}\label{trialadapter}
Any effectively calculable trial resolution $T_\pi(S)$
 can be implemented by the adapter $\Re_1[S \cap \Sigma_\pi]$.

\comment{Proof}
$T_\pi(S) = \{\, s\in S \mid s\in\Sigma_\pi \,\} =
            \{\, s \mid s \!\in\! S \land s \!\in\! \Sigma_\pi \,\} =
            \{\, s \mid P_S(s) \land P_\pi(s) \,\}$.
Then
$T_\pi(S) \doteq \Re_1[P_S \land P_\pi] = \Re_1[S \cap \Sigma_\pi]$.
The equality is dotted because,
if the trial is not an effectively calculable function,
then it cannot be implemented.
{\QED}

\clause{Summary}
In practice, an adapter
 $\Re_1[P_S \land P_\pi] = \Re_1[S \cap \Sigma_\pi]$
has a body capable of several
behaviors that provides the set $S$ of behaviors.
If the current behavior were not satisfying
the adapter condition $P_\pi$,
which is interpreted as an error,
then the adapter would change its behavior
 trying another one in $S$.

\comment{Comment}
The adapter is
a body capable of several behaviors, and
a governor that selects the current behavior.

\comment{Example}
A deciduous tree,
 which switches its behavior with seasons,
is an adapter.

\subsection{Perceiver}

\clause{Definition}
A perceiver $\Re_2$ is any resolver that implements
one transformation of the elements in $\bb S$
into the elements in $\bb S$.
We will note $\Re_2[f]$ the perceiver that implements $f$,
 where $f \in ({\bb S} \to {\bb S})$, that is,
$\Re_2[f] = f \in ({\bb S} \to {\bb S})$.
Then the perceiver resources
are in ${\bb S} \to {\bb S}$, and
$\generic{\Re_2}=({\bb S} \to {\bb S})$.

\comment{Comment}
From a semantic point of view,
a perceiver $\Re_2$ implements
 a semantic functional computation.
From a syntactic point of view,
a perceiver $\Re_2$ implements
 a syntactic unconditional computation.

\clause{Remark}
Perceivers are to syntax as mechanisms are to semantics.

\comment{Comment}
When solutions are functions ${\bb S} \to {\bb S}$,
then a perceiver does what a mechanism does,
which is to return a solution unconditionally. That is,
perceivers on metaproblems are as mechanisms on problems.
But, perceivers can go further.

\comment{Comment}
The perceiver $\Re_2[f]$ implements
function $f$ from ${\bb S}$ to ${\bb S}$,
that is, $f: {\bb S} \to {\bb S}$.
Then,
$\forall s\in {\bb S},\; \Re_2[f](s) = f(s) \in {\bb S}$.

\clause{Notation}\label{hardwaresoftware}
By the rewriting rules in \ref{functionset},
$\Re_2[f](S) = \{\, \Re_2[f](s) \mid s \in S \,\} \in 2^{\bb S}$.
Then $\Re_2[f](S)$ returns a set of solutions,
as any well-behaved resolution should do.

\comment{Comment}
The perceiver $\Re_2[f](S)$
implements $f$, meaning that $f$ is hardwired in the perceiver,
while $S$ is just data.
We will call what is implemented hardware,
and what is data software.
We write the hardware between brackets,
and the software between parentheses.
We will assume that
coding software costs less than implementing hardware,
or, in fewer words, that
software is cheaper than hardware

\clause{Lemma}\label{adaptersareperceivers}
Every adapter $\Re_1$ is a perceiver $\Re_2$, that is,
$\generic{\Re_1} \subset \generic{\Re_2}$.

\comment{Proof}
Because ${\bb B} \subset {\bb S}$, and then
 $({\bb S}\to{\bb B}) \subset ({\bb S}\to{\bb S})$. So\\
$\generic{\Re_1} = ({\bb S}\to{\bb B}) \subset
 ({\bb S}\to{\bb S}) = \generic{\Re_2}$.
{\QED}

\comment{Comment}
Each adapter implements one condition
 $P_S \in ({\bb S} \to {\bb B})$.
And any condition $P_S \in ({\bb S} \to{\bb B})$ is also
 a function $P_S \in ({\bb S} \to {\bb S})$,
 because ${\bb B} \subset {\bb S}$.
Therefore, for each adapter $\Re_1[P_S]$,
 which implements condition $P_S$,
there is a perceiver $\Re_2[P_S]$
 that implements the function $P_S$,
and then we write
 $\Re_1[P_S] = P_S = \Re_2[P_S]$.

\comment{Comment}
Again, $\Re_1[P_S] = \Re_2[P_S]$ explains
that the results are the same, but not the implementation.

\clause{Lemma}\label{identityperceiver}
$\forall S\in2^{\bb S},\; \Re_1[S] = \Re_2[i](S)$.

\comment{Proof}
Function $i: {\bb S} \to {\bb S}$ is the semantic identity,
$i={\frak i}$ see \ref{semanticfunction},
 so $\forall s \in {\bb S},\; i(s) = s$, and
$\Re_2[i](S) =
 \{\, \Re_2[i](s) \mid s \in S \,\} =
 \{\, i(s) \mid s \in S \,\} =
 \{\, s \mid s \in S \,\} =
 S = \Re_1[S]$.
{\QED}

\comment{Comment}
The same perceiver hardware $\Re_2[i]$,
just by changing its software, 
can emulate different adapters:
$\Re_2[i](S) = \Re_1[S]$, and $\Re_2[i](S') = \Re_1[S']$.
Then the perceiver $\Re_2[i](S)$
is more flexible than the adapter $\Re_1[S]$,
because $S$ is hardwired in the adapter,
while it is easily replaceable data for the perceiver.

\clause{Lemma}\label{perceiverrange}
$\forall S\in2^{\bb S}$,
 $\Xi\Re_1[S] = \Xi\Re_2[i](S)$ and
 $\Phi\Re_1[S] = \Phi\Re_2[i](S)$.

\comment{Proof}
Because, by Lemma~\ref{identityperceiver},
$\Re_1[S] = \Re_2[i](S)$.
{\QED}

\comment{Definition}
The perceiver condition is satisfied if
it implements the semantic identity~$i$.

\comment{Comment}
If the perceiver condition holds,
then the perceiver $\Re_2[i](S)$ solves
any problem solved by the adapter $\Re_1[S]$, and
the perceiver $\Re_2[i](S)$ resolves
any problem resolved by the adapter $\Re_1[S]$.

\comment{Remark}
Semantic identity $i$ is the ideal for perception.

\clause{Corollary}
$\Xi\Re_2[i](S \cap \Sigma_\pi) = \Xi\Re_1[S \cap \Sigma_\pi]$ and
$\Phi\Re_2[i](S \cap \Sigma_\pi) = \Phi\Re_1[S \cap \Sigma_\pi]$.

\comment{Proof}
By Lemma~\ref{perceiverrange}.
{\QED}

\comment{Comment}
The same perceiver hardware $\Re_2[i]$ can be
tuned to a different trial just by changing its software,
from $\Re_2[i](S \cap \Sigma_\pi)$
to $\Re_2[i](S' \cap \Sigma_\rho)$,
for example.

\comment{Proposition}
If $S \subset S'$, then
$\Xi\Re_2[i](S \cap \Sigma_\pi) \subset
 \Xi\Re_2[i](S' \cap \Sigma_\pi)$, and
$\Phi\Re_2[i](S \cap \Sigma_\pi) \subset
 \Phi\Re_2[i](S' \cap \Sigma_\pi)$,
by Lemma~\ref{identityperceiver} and
corollaries to Lemmas
 \ref{adapterrange} and \ref{adapterpower}.

\clause{Definition}\label{elementable}
A function on sets $F: 2^{\bb S} \to  2^{\bb S}$ is elementable
if it exists an effectively calculable function
 $f: {\bb S} \to  {\bb S}$
such that $\forall S,\; F(S) = \{\, f(s) \mid s \in S \,\}$.

\comment{Comment}
We write $F(S) = f(S)$, by the rules in \ref{functionset}.
Note the three requirements:
 that $f$ is a semantic function, $f: {\bb S} \to  {\bb S}$,
 that $f$ is effectively calculable, and
 that $F(S) = f(S)$.

\comment{Proposition}
Set identity $I: 2^{\bb S} \to  2^{\bb S} \mid
\forall S \in 2^{\bb S},\; I(S)=S$,
is elementable by semantic identity $i$, because
$\forall S \in 2^{\bb S},\; 
i(S) = \{\, i(s) \mid s \in S \,\} = \{\, s \mid s \in S \,\} =
S = I(S)$.

\clause{Lemma}\label{analogyperceiver}
Any analogy resolution
 $A\circ T_{A\pi}(S)\circ {\cal T}_A$
can be implemented by the tri-perceiver
 $\Re_2[{\cal T}_a](\Re_2[i](S \cap \Re_2[a](\Sigma_\pi)))$,
if $A$ is elementable by $a$, and
${\cal T}_A$ by ${\cal T}_a$.

\comment{Proof}
An analogy resolution is
 $A\circ {T\!}_{A\pi}(S)\circ {\cal T}_A$.
Both $A$ and ${\cal T}_A$ are functions from sets to sets,
${\cal T}_A : 2^{\bb S} \to  2^{\bb S}$ and
$A : ({\bb P} \to  {\bb P}) = (2^{\bb S} \to  2^{\bb S})$, so
if both $A$ and ${\cal T}_A$ are elementable,
then a perceiver can implement them.
We have $a$ and ${\cal T}_a$, which are both semantic functions
such that $A(S) = a(S)$, and ${\cal T}_A(S) = {\cal T}_a(S)$.
Then $P_{a\pi} = a(P_\pi) = \Re_2[a](P_\pi) = \Re_2[a](\Sigma_\pi)$
 implements the first third,
${T\!}_{a\pi}(S) = \Re_1[P_S \land P_{a\pi}] =
 \Re_1[P_S \land \Re_2[a](P_\pi)] = \Re_1[S \cap \Re_2[a](\Sigma_\pi)]$
 implements the second third, see \ref{trialadapter},
and using \ref{identityperceiver},
 $\Re_2[{\cal T}_a](\Re_2[i](S \cap \Re_2[a](\Sigma_\pi)))$
implements the whole analogy resolution
 $a\circ T_{a\pi}(S)\circ {\cal T}_a$.
{\QED}

\comment{Corollary}
Identity analogy $I\circ T_{I\pi}(S)\circ {\cal T}_I$
can be implemented by the bi-perceiver
$\Re_2[i](S \cap \Re_2[i](\Sigma_\pi))$,
which uses the identity perceiver $\Re_2[i]$ twice.

\comment{Proof}
$\Re_2[{\cal T}_i](\Re_2[i](S \cap \Re_2[i](\Sigma_\pi))) =
 \Re_2[i](\Re_2[i](S \cap \Re_2[i](\Sigma_\pi))) =
 \Re_2[i](S \cap \Re_2[i](\Sigma_\pi))$,
because set identity $I$ is elementable by semantic identity $i$,
and ${\cal T}_I = I$, so it is also elementable by ${\cal T}_i = i$,
and $i \circ i = i$.
{\QED}

\clause{Summary}
While an adapter uses a trial and error resolution,
and this means that error is part of the usual procedure,
a perceiver executes the trial and error inside itself.
If the analogy provides a good model, then
the internal trial is as good as the external one,
with the advantage that the errors are only simulated errors.
More to the point,
if the problem the resolver faces is the survival problem,
then the adapter errors are literally death errors,
or at least pain,
while the perceiver errors are just mental previsions
of what not to do.
See that, if the perceiver implements the
identity analogy, as $\Re_2[i]$ does,
then the model is good,
because the internal problem is equal to the external one.
And the perceiver $\Re_2[i]$ is more
flexible than the adapter.

\comment{Comment}
The perceiver is
a body capable of several behaviors,
a governor that selects the current behavior, and
a simulator that internalizes behaviors.

\comment{Example}
The perceiver governor determines what to do
based upon an internal interpretation.
According to \cite Lettvin et al. (1959),
a frog is a perceiver that uses an internal routine.
Frog's $i$ is such that any dark point that moves
rapidly in its field of vision is a fly which it will
try to eat.

\subsection{Learner}

\clause{Definition}
A learner $\Re_3$ is any resolver that implements
one condition on the members of ${\bb S} \to {\bb S}$.
We will note $\Re_3[P_F]$ the learner that implements $P_F$,
 where $P_F \in (({\bb S} \to {\bb S})\to{\bb B})$,
 that is, $\Re_1[P_F] = P_F \in (({\bb S} \to {\bb S})\to{\bb B})$.
Then the learners resources are in
 $({\bb S} \to {\bb S})\to{\bb B}$, and
$\generic{\Re_3} = (({\bb S} \to {\bb S})\to{\bb B})$.

\comment{Comment}
A learner $\Re_3$ implements a syntactic conditional computation.

\clause{Remark}
Learners are to syntax as adapters are to semantics.

\comment{Comment}
When solutions are functions ${\bb S} \to {\bb S}$,
then a learner does what an adapter does,
which is to return a predicate on solutions.
That is, learners on metaproblems are as adapters on problems.
But, learners can go further.

\clause{Lemma}
Each learner $\Re_3[P_F]$ implements 
one set of members of $({\bb S} \to {\bb S})$.

\comment{Proof}
Because every predicate
$P_F: ({\bb S} \to {\bb S}) \to {\bb B}$
defines a set\\
$F = \{\, f\in({\bb S} \to {\bb S}) \mid P_F(f) \,\}
 \in 2^{{\bb S} \to {\bb S}}$.
The condition $P_F$ is the characteristic function of $F$,
$\forall f\in({\bb S} \to {\bb S}),\; P_F(f) = [f \in F]$.
{\QED}

\comment{Comment}
We will write
$\Re_3[P_F] = \Re_3[F] = F  \in 2^{{\bb S} \to {\bb S}}$.

\comment{Comment}
The learner $\Re_3[F]$ implements $F \in 2^{{\bb S} \to {\bb S}}$.
So $\forall s\in{\bb S},\, \Re_3[F](s) = F(s) \in 2^{\bb S}$,
because $F(s) =\{\, f(s) \mid f\in F\,\}$,
by the rewriting rules in \ref{functionset},
and then $\Re_3[F](s)$ returns a set of solutions,
as any well-behaved resolution should do.
Also, by the same rules,
$\Re_3[F](S) = F(S) = 
\{\, f(s) \mid s \!\in\! S \times f \!\in\! F \,\} \in 2^{\bb S}$.

\clause{Lemma}\label{perceiversarelearners}
Every perceiver $\Re_2$ is a learner $\Re_3$, that is,
$\generic{\Re_2} \subset  \generic{\Re_3}$.

\comment{Proof}
For each perceiver $\Re_2[f]$,
 which implements $f\in({\bb S}\to{\bb S})$,
there is a learner $\Re_3[P_{\delta_f}]$,
see \ref{deltaiscalculable}, that implements the 
singleton $\{f\}\in(({\bb S}\to{\bb S})\to{\bb B})$.
But not every set is a singleton. Then,
$\generic{\Re_2} = ({\bb S}\to{\bb S}) \subset
 (({\bb S}\to{\bb S})\to{\bb B}) = \generic{\Re_3}$.
{\QED}

\clause{Lemma}\label{learnerisunionofperceivers}
$\Re_3[F] = \bigcup_{f\in F} \{\Re_2[f]\}$.

\comment{Proof}
Because $\Re_2[f] = f$,
so $\bigcup_{f\in F} \{\Re_2[f]\} = \bigcup_{f\in F} \{f\} =
 F = \Re_3[F]$.
{\QED}

\comment{Comment}
Again, the results are the same, but not the implementation.
The union of perceivers $\bigcup_{f\in F} \{\Re_2[f]\}$
cannot select a function to use,
so it cannot implement any meta-trial,
as $\Re_3[R\cap\Sigma_{\Pi\pi}]$ does,
see \ref{metatriallearner}.

\comment{Comment}
These are not sets of solutions, but sets of semantic functions.

\clause{Lemma}
$\forall f\in F$, $\forall S\in2^{\bb S}$,
$\Re_2[f](S) \subseteq \Re_3[F](S)$.

\comment{Proof}
If $f \in F$, then
$f(S) = \{\, f(s) \mid s \in S \,\}
\subseteq
\{\, f'(s) \mid s \!\in\! S \times f' \!\in\! F \,\} = F(S)$.
{\QED}

\clause{Notation}
We will rewrite $\Re_1[\Re_3[F](S) \cap \Sigma_\pi]$ as
$\Re_3[F](S \Cap \Sigma_\pi)$.

\comment{Comment}
We can write
$\Re_1[\Re_3[F](S) \cap \Sigma_\pi] = \Re_3[F](S \Cap\Sigma_\pi)$,
because any learner can implement semantic conditions,
that is, because $\generic{\Re_1} \subset \generic{\Re_3}$.

\clause{Lemma}\label{learnerrange}
If $f \in F$, 
then $\forall S\in2^{\bb S}$, $\forall \pi \in {\bb P}$,
$\Xi\Re_2[f](S) \subseteq \Xi\Re_3[F](S \Cap\Sigma_\pi)$.

\comment{Proof}
Firstly see that, if $f \in F$, then $f(S) \subseteq F(S)$,
 by \ref{functionset}, so,
$f(S)\cap\Sigma_\pi \subseteq F(S)\cap\Sigma_\pi$.
Secondly see that
$\forall \pi \in \Xi\Re_2[f](S)$,
$f(S) \cap \Sigma_\pi = f(S) \neq \emptyset$.
This is because
$\Xi\Re_2[f](S) = 
 %\{\, \pi \mid \Re_2[f](S) \subseteq \Sigma_\pi 
 %\land \Re_2[f](S) \neq \emptyset \,\} =
 \{\, \pi \mid f(S) \subseteq \Sigma_\pi
 \land f(S) \neq \emptyset \,\}$.
Now, taking both together,
$\forall \pi \in \Xi\Re_2[f](S)$,
$\emptyset \neq f(S) = f(S) \cap \Sigma_\pi \subseteq
 F(S) \cap \Sigma_\pi$,
so for these $\pi$,
$F(S)\cap\Sigma_\pi \neq \emptyset$.
Then these
$\pi \in \{\, \pi \mid
 F(S)\cap\Sigma_\pi \subseteq \Sigma_\pi \land
 F(S)\cap\Sigma_\pi \neq \emptyset \,\} =
 \Xi\Re_1[\Re_3[F](S) \cap \Sigma_\pi]$,
because $F(S)\cap\Sigma_\pi \subseteq \Sigma_\pi$
is always true.
{\QED}

\comment{Definition}
We will call $f \in F$ the learner condition.
If the learner condition holds,
then the learner $\Re_3[F](S \Cap\Sigma_\pi)$ solves
any problem that the perceiver $\Re_2[f](S)$ solves.

\clause{Lemma}\label{learnerpower}
If $f \in F$,
then $\forall S \in 2^{\bb S}$, $\forall \pi \in {\bb P}$,
$\Phi\Re_2[f](S) \subseteq \Phi\Re_3[F](S \Cap \Sigma_\pi)$.

\comment{Proof}
If $f \in F$, then $f(S) \subseteq F(S)$,
 see \ref{functionset}, so,
$f(S)\cap\Sigma_\pi \subseteq F(S)\cap\Sigma_\pi$.
Now $\forall \pi \in \Phi\Re_2[f](S)$,
 $f(S)=\Sigma_\pi$, and then for these $\pi$,
 $\Sigma_\pi = f(S)\cap \Sigma_\pi \subseteq 
  F(S)\cap\Sigma_\pi \subseteq \Sigma_\pi$.
Therefore, for these $\pi$,
$F(S)\cap\Sigma_\pi = \Sigma_\pi$, and
$\Phi\Re_2[f](S) \subseteq
 \{\, \pi \mid F(S)\cap\Sigma_\pi = \Sigma_\pi \,\} =
 \Phi\Re_1[\Re_3[F](S) \cap \Sigma_\pi]$.
{\QED}

\comment{Comment}
If the learner condition holds, $f \in F$,
then the learner $\Re_3[F](S \Cap \Sigma_\pi)$ resolves
any problem that the perceiver $\Re_2[f](S)$ resolves.

\clause{Corollary}
In particular, if $i\in R$, then
$\Xi\Re_2[i](S\cap\Sigma_\pi) \subseteq
 \Xi\Re_3[R](S \Cap \Sigma_\pi)$ and
$\Phi\Re_2[i](S\cap\Sigma_\pi) \subseteq
 \Phi\Re_3[R](S \Cap \Sigma_\pi)$.

\comment{Proof}
By Lemmas \ref{learnerrange} and \ref{learnerpower}.
See that
$\Xi\Re_3[R](S \cap \Sigma_\pi \Cap \Sigma_\pi) \subseteq
 \Xi\Re_3[R](S \Cap \Sigma_\pi)$, and
$\Phi\Re_3[R](S \cap \Sigma_\pi \Cap \Sigma_\pi) \subseteq
 \Phi\Re_3[R](S \Cap \Sigma_\pi)$,
because
$\Re_3[R](S \cap \Sigma_\pi) \subseteq \Re_3[R](S)$,
so corollaries to Lemmas
 \ref{adapterrange} and \ref{adapterpower}
apply (the equal case is trivial).
{\QED}

% \clause{Remark}
% A learner solves metaproblems by trial,
% as an adapter solves problems by trial.
% The following correlations stand:
% $\Re_3 \leftrightarrow \Re_1$,
% $\Pi\pi \leftrightarrow \pi$,
% $R \leftrightarrow S$, and then
% $\Re_3[R \cap \Sigma_{\Pi\pi}] \leftrightarrow
%  \Re_1[S \cap \Sigma_\pi]$.
% Therefore, $\Re_3[R \cap \Sigma_{\Pi\pi}]$
% compares to $\Re_2[i]$, where $i \in R$,
% as $\Re_1[S \cap \Sigma_\pi]$
% compares to $\Re_0[s]$, where $s \in S$.

\clause{Lemma}\label{metatriallearner}
Any meta-trial resolution $T_{\Pi\pi}(R)$ can be
 implemented by the learner $\Re_3[R \cap \Sigma_{\Pi\pi}]$,
if $P_R$ and $P_{\Pi\pi}$ are elementable.

\comment{Comment}
The diagram for the meta-trial, or trial of the metaproblem,
see \ref{metadiagrams}, is:
$${\pi} \into\Pi {\Pi\pi}
  \into{T_{\Pi\pi}(R)} {\Sigma_{\Pi\pi}}
  \into{{\cal T}_\Pi} {\Sigma_\pi} \;.
$$

\comment{Comment}
A learner solves metaproblems by trial,
as an adapter solves problems by trial.
The following correlations stand:
$\Re_3 \leftrightarrow \Re_1$,
$\Pi\pi \leftrightarrow \pi$,
$R \leftrightarrow S$, and then
$\Re_3[R \cap \Sigma_{\Pi\pi}] \leftrightarrow
 \Re_1[S \cap \Sigma_\pi]$.
Therefore, $\Re_3[R \cap \Sigma_{\Pi\pi}]$
compares to $\Re_2[i]$, where $i \in R$,
as $\Re_1[S \cap \Sigma_\pi]$
compares to $\Re_0[s]$, where $s \in S$.

\comment{Proof}
$T_{\Pi\pi}(R) = \{\, r\in R \mid r\in\Sigma_{\Pi\pi} \,\} =
             \{\, r \mid r\!\in\! R \,\land\, r\!\in\!\Sigma_{\Pi\pi} \,\} =
             \{\, r \mid P_R(r) \land P_{\Pi\pi}(r) \,\}$.
In the meta-trial $T_{\Pi\pi}(R)$,
$R$ is a set of resolutions, where
 ${\bb R} = ({\bb P}\to2^{\bb S}) = (2^{\bb S}\to2^{\bb S})$,
that is, $P_R: (2^{\bb S}\to2^{\bb S}) \to {\bb B}$,
and the condition of the metaproblem $\Pi\pi$ is
also $P_{\Pi\pi} : {\bb R} \to {\bb B} =
 ({\bb P}\to2^{\bb S}) \to {\bb B} =
 (2^{\bb S}\to2^{\bb S}) \to {\bb B}$.
So if both $P_R$ and $P_{\Pi\pi}$ are elementable
by $\wp_R$ and $\wp_{\Pi\pi}$, then
both of them, and its conjunction, are implementable. Then  
$T_{\Pi\pi}(R) = \{\, r \mid P_R(r) \land P_{\Pi\pi}(r) \,\} \doteq
 \Re_3[\wp_R \land \wp_{\Pi\pi}] = \Re_3[R \cap \Sigma_{\Pi\pi}]$.
{\QED}

\clause{Summary}
Perceiver success depends crucially on the analogy,
that is, on how much the analogy resembles the identity $i$.
And a learner can adapt the analogy
 to the problem it is facing,
because the learner $\Re_3[R]$
implements a set of functions $R$
from which it can select another analogy
when the current one fails.
Adapting the analogy is also known as modeling.
So a learner can apply different analogies,
but a learner can also apply a routine
if it knows a solution, because the routine is more efficient,
or a trial, when the model is not good enough
or too pessimistic.

\comment{Comment}
The learner is
a body capable of several behaviors,
a governor that selects the current behavior,
a simulator that internalizes behaviors, and
a modeler that adjusts the model used by the simulator.

\comment{Example}
Where there is modeling and simulation there is learning,
because enhancing the model prevents repeating errors.
A dog is a learner.

\subsection{Subject}

\clause{Definition}
A subject $\Re_4$ is any resolver that implements
one transformation of the elements in ${\bb S}\to{\bb S}$
into the elements in ${\bb S}\to{\bb S}$.
We will note $\Re_4[{\frak f}]$
 the subject that implements $\frak f$, where
 ${\frak f} \in (({\bb S} \to {\bb S}) \to ({\bb S} \to {\bb S}))$,
 that is, $\Re_4[{\frak f}] = {\frak f}
 \in (({\bb S} \to {\bb S}) \to ({\bb S} \to {\bb S}))$.
Then the subject resources are in
 $({\bb S}\to{\bb S}) \to ({\bb S}\to{\bb S})$, and
$\generic{\Re_4} = (({\bb S}\to{\bb S}) \to ({\bb S}\to{\bb S}))$.

\comment{Comment}
A subject $\Re_4$ implements a syntactic functional computation.

\clause{Remark}
Subjects are to syntax as perceivers are to semantics.

\comment{Comment}
When solutions are functions ${\bb S} \to {\bb S}$,
then a subject does what a perceiver does,
which is to return a function on solutions to solutions.
That is, subjects on metaproblems are as perceivers on problems.
But, subjects can go further.

\comment{Comment}
The subject $\Re_4[{\frak f}]$ implements function $\frak f$
 from ${\bb S}\to{\bb S}$ to ${\bb S} \to {\bb S}$,
that is, ${\frak f}: ({\bb S}\to{\bb S}) \to ({\bb S} \to {\bb S})$.
Then, $\forall f\in({\bb S}\to{\bb S}),\;
 \Re_4[{\frak f}](f) = {\frak f}(f) \in ({\bb S} \to {\bb S})$.

\clause{Notation}
As resolutions return sets of elements in $\bb S$,
to normalize the situation of subjects,
for which $\Re_4[{\frak f}](f)(s) \in {\bb S}$,
we will use the rewriting rules in \ref{functionset}
 to get
 $\Re_4[{\frak f}](F)(S) = 
 \{\, \Re_4[{\frak f}](f)(s) \mid s \!\in\! S \times f \!\in\! F \,\}
 \in 2^{\bb S}$.

\comment{Comment}
Subject $\Re_4[{\frak f}](F)(S)$
has two software levels:
semantics~($S$) and syntax~($F$).

\clause{Lemma}\label{learnersaresubjects}
Every learner $\Re_3$ is a subject $\Re_4$, that is,
$\generic{\Re_3} \subset \generic{\Re_4}$.

\comment{Proof}
First we define set ${\frak B} = \{ K_\top, K_\bot \}$,
both functions ${\bb S} \to {\bb S}$,
 see \ref{constantfunction}.
Next we define the the natural isomorphism between $\bb B$
and ${\frak B}$, mapping
$\top$ to the function that always returns $\top$,
 which is $K_\top=P_\tau$, and
$\bot$ to the function that always returns $\bot$,
 which is $K_\bot=P_{\bar\tau}$,
 see \ref{tautologycontradiction}.
And so ${\bb B} \Leftrightarrow {\frak B}:
 \top \leftrightarrow K_\top,
 \bot \leftrightarrow K_\bot$, and
${\bb B} \cong {\frak B} \subset ({\bb S} \to {\bb S})$.
Then,
$(({\bb S} \to {\bb S}) \to {\bb B}) \subset
 (({\bb S} \to {\bb S}) \to ({\bb S} \to {\bb S}))$,
and therefore
$\generic{\Re_3} =
 (({\bb S} \to {\bb S}) \to {\bb B}) \subset
 (({\bb S} \to {\bb S}) \to ({\bb S} \to {\bb S})) =
\generic{\Re_4}$.
{\QED}

\comment{Comment}
Each learner implements one condition 
 $P_F \in (({\bb S} \to {\bb S}) \to {\bb B})$.
And any condition on functions
 $P_F \in (({\bb S} \to {\bb S}) \to {\bb B})$
is also a function on functions to functions
 $P_F \in (({\bb S} \to {\bb S}) \to ({\bb S} \to {\bb S}))$,
because ${\bb B} \subset ({\bb S} \to {\bb S})$.
Therefore, for each learner $\Re_3[P_F]$,
 which implements condition $P_F$,
there is a subject $\Re_4[P_F]$ that implements
the function $P_F$, and then we write
$\Re_3[P_F]= P_F = \Re_4[P_F]$.

\comment{Comment}
Again, $\Re_3[P_F] = \Re_4[P_F]$ explains
that the results are the same, but not the implementation.

\clause{Lemma}\label{identitysubject}
$\forall F\in2^{\bb S\to S}$, $\forall S\in2^{\bb S}$,
$\Re_3[F](S) = \Re_4[{\frak u}](F)(S)$.

\comment{Comment}
Function $\frak u$ is the identity for programs,
 see \ref{universalprogram},
 or functional identity, or evaluation,
 see \ref{identityequivalence}.
Function ${\frak u}$ is syntactic because it is
not restricted to semantic objects,
see \ref{semanticfunction}.
Syntactic function ${\frak u}$
is equivalent to $\lambda$-calculus $I=(\lambda x.x)$,
 see \ref{identityfunction}:
$\forall f\in{\bb S}\to{\bb S}$, ${\frak u}(f) = f$ and
$\forall s\in{\bb S}$, ${\frak u}(f)(s) = f(s)$.

\comment{Proof}
By the rewriting rules in \ref{functionset},
$\Re_4[{\frak u}](F)(S) =
 \{\, \Re_4[{\frak u}](f)(s) \mid s \!\in\! S \times f \!\in\! F \,\} =
 \{\, {\frak u}(f)(s) \mid s \!\in\! S \times f \!\in\! F \,\} =
 \{\, f(s) \mid s \!\in\! S \times f \!\in\! F \,\} = F(S) = \Re_3[F](S)$.
{\QED}

\comment{Corollary}
$\forall F\in2^{\bb S\to S}$, $\Re_3[F] = \Re_4[{\frak u}](F)$.

\clause{Lemma}\label{identitysubjectrange}\label{subjectrange}
$\forall F\in2^{\bb S\to S}$, $\forall S\in2^{\bb S}$,
 $\Xi\Re_3[F](S) = \Xi\Re_4[{\frak u}](F)(S)$ and\\
 $\Phi\Re_3[F](S) = \Phi\Re_4[{\frak u}](F)(S)$.

\comment{Proof}
Because, by Lemma~\ref{identitysubject},
$\Re_3[F](S) = \Re_4[{\frak u}](F)(S)$.
{\QED}

\comment{Definition}
The subject condition is satisfied if it implements
the functional identity~${\frak u}$.

\comment{Comment}
If the subject condition holds, then
the subject $\Re_4[{\frak u}](F)(S)$ solves
any problem solved by the learner $\Re_3[F](S)$, and also
the subject $\Re_4[{\frak u}](F)(S)$ resolves
any problem resolved by the learner $\Re_3[F](S)$.

\comment{Remark}
Functional identity $\frak u$ is the ideal for reason.

\clause{Corollary}
In particular,
$\Xi\Re_4[{\frak u}](R)(S \Cap\Sigma_\pi) =
 \Xi\Re_3[R](S \Cap\Sigma_\pi)$ and\\
$\Phi\Re_4[{\frak u}](R)(S \Cap\Sigma_\pi) =
 \Phi\Re_3[R](S \Cap\Sigma_\pi)$.

\comment{Proof}
By Lemma~\ref{subjectrange}.
{\QED}

\comment{Comment}
Subject $\Re_4[{\frak u}](R)(S \Cap\Sigma_\pi)$
is more flexible than
learner $\Re_3[R](S \Cap\Sigma_\pi)$, because
$R$ is software for the subject while
it is hardware in the learner,
and software is cheaper than hardware,
see \ref{hardwaresoftware}.

\clause{Theorem}\label{subjectisaresolutionmachine}
Subject $\Re_4[{\frak u}]$ is a full resolution machine.

\comment{Proof}
By Theorem~\ref{universalTuringmachine}
and Lemma~\ref{universalprogram},
$\Re_4[{\frak u}] = {\frak u} = {\frak c}({\cal U})$,
so using the program isomorphism,
see \ref{programisomorphism},
$\Re_4[{\frak u}] = {\cal U}$,
which is a Turing complete device,
and therefore is a full resolution machine,
by Theorem~\ref{syntaxengineisuniversal}.
{\QED}

\clause{Lemma}\label{metaanalogysubject}
Any effectively calculable resolution $\Re$
 can be implemented by the subject $\Re_4[{\frak u}]$.

\comment{Proof}
By Theorem~\ref{subjectisaresolutionmachine}.
{\QED}

\comment{Corollary}
Any effectively calculable meta-analogy resolution
${\cal A}\circ{T\!}_{{\cal A}\Pi\pi}(R)\circ{\cal T}_{\cal A}$
can be implemented by the subject $\Re_4[{\frak u}]$,
including metaresolving, see \ref{metaresolving}.
% Otherwise we could not write nor read this paper.

\clause{Summary}
The subject, by internalizing metaproblems,
prevents meta-errors, that is,
the subject can test internally a resolution before executing it.
The subject is also more flexible than the learner, because
subject modeling is done in software, instead of in hardware. 
And subject $\Re_4[{\frak u}]$ can reason about any model.
This means that subject $\Re_4[{\frak u}]$
is a resolver that can calculate solutions,
but also problems and resolutions without limits;
it can represent the problem it is facing to itself,
and it can represent itself to itself.
In this sense, the subject $\Re_4[{\frak u}]$ is conscious.

\comment{Comment}
The subject is
a body capable of several behaviors,
a governor that selects the current behavior,
a simulator that internalizes behaviors,
a modeler that adjusts the model used by the simulator, and
a reason that internalizes resolutions.

\comment{Example}
It seems that only our species, {\it Homo sapiens},
is Turing complete.
We deal with the evolution to Turing completeness
and its relation to language in \cite Casares (2016b).

\subsection{Resolvers Hierarchy}

\clause{Theorem}\label{resolvers hierarchy}
There is a hierarchy of resolvers:\hfil\break
$\generic{\Re_0} \subset \generic{\Re_1} \subset
 \generic{\Re_2} \subset \generic{\Re_3} \subset
 \generic{\Re_4}$.

\comment{Proof}
Because ${\bb S}\subset({\bb S}\to{\bb B})\subset
 ({\bb S}\to{\bb S})\subset
 (({\bb S}\to{\bb S})\to{\bb B})\subset
 (({\bb S}\to{\bb S})\to({\bb S}\to{\bb S}))$,
by Lemmas
\ref{mechanismsareadapters},
\ref{adaptersareperceivers},
\ref{perceiversarelearners}, and
\ref{learnersaresubjects}.
There are not more types of resolvers, because
there is not a resolver more capable than
 $\Re_4[{\frak u}]={\cal U}$,
by Theorems \ref{subjectisaresolutionmachine}
and \ref{maximumcomputing}.
{\QED}

\needspace{14\baselineskip}

\clause{Summary}\label{summary}
This table groups concepts closely related from
 problem theory, as trial,
 computing theory, as adapter $\Re_1$, and
 set theory, as $S\in{\bb S}\to{\bb B}$.
$$\abovedisplayskip=0pt plus 12pt
\halign{\strut\hfil#\quad\vrule&
 \space\hfil#\hfil\space\vrule&
 \space\hfil#\hfil\space&\vrule\quad#\hfil\cr
\hphantom{function}&
\hphantom{${\frak f} \in ({\bb S}\to{\bb S})\to({\bb S}\to{\bb S})$}&&\cr
 & {\bf Semantics}& {\bf Syntax}& \cr
 \noalign{\hrule}
 & Routine& Meta-routine& \cr
one & Mechanism $\Re_0$& Perceiver $\Re_2$& element\cr
 & $s \in {\bb S}$& $f \in ({\bb S}\to{\bb S})$& \cr
 \noalign{\hrule}
 & Trial& Meta-trial& \cr
some & Adapter $\Re_1$& Learner $\Re_3$& set\cr
 & $S \in {\bb S}\to{\bb B}$& $F \in ({\bb S}\to{\bb S})\to{\bb B}$& \cr
 \noalign{\hrule}
 & Analogy& Meta-analogy& \cr
any & Perceiver $\Re_2$& Subject $\Re_4$& function\cr
 & $f \in {\bb S}\to{\bb S}$& ${\frak f} \in ({\bb S}\to{\bb S})\to({\bb S}\to{\bb S})$& \cr
 \noalign{\hrule}
 & ${\bb S}$& ${\bb S}\to{\bb S}$& \cr
 & Elements& Functions& \cr
}$$

\comment{Comment}
A perceiver is a syntactic mechanism.
A learner is a syntactic adapter.
A subject is a syntactic perceiver.
A subject is a syntactic$^2$ mechanism.

\clause{Theorem}\label{problemtheoryiscomplete}
The problem theory is complete.

\comment{Proof}
Aside from definitions,
the problem theory posits that
there are three ways to resolve a problem:
routine, trial, and analogy; see \ref{Resolution}.
Adding the metaproblem of the problem,
we get five ways to resolve a problem and its metaproblem,
which are the basic three plus meta-trial and meta-analogy,
see \ref{fivetypesofresolution}.
For each way there is a resolver,
see Lemmas \ref{routinemechanism}, \ref{trialadapter},
\ref{analogyperceiver}, \ref{metatriallearner}, and
\ref{metaanalogysubject},
and the resources of each resolver are in a
series of mathematical objects of increasing generality
that covers everything until syntactic functions,
 see \ref{resolvers hierarchy} and \ref{summary}.
Now, to execute meta-analogies, 
\lower 2pt\hbox{$2^{2^{\bb S}\to2^{\bb S}}\to2^{2^{\bb S}\to2^{\bb S}}\!\!$},
 see \ref{metadiagrams},
or at least the elementable ones,
 see \ref{elementable},
we need subjects,
 which implement syntactic functions
 $({\bb S}\to{\bb S})\to({\bb S}\to{\bb S})$.
And there is a subject that
is a Turing complete device, $\Re_4[{\frak u}]$,
 see \ref{subjectisaresolutionmachine},
so it has the maximum computing power, 
 see \ref{maximumcomputing},
and then the maximum resolving power,
 see \ref{syntaxengineisuniversal}.
This means that
there are not more resolvers beyond the subject,
and therefore that the series is complete, and then that 
the problem theory covers everything and is complete.
{\QED}

\comment{Comment}
It also means that no more resolutions are needed,
although we could do without routine, for example,
by using Theorem~\ref{routineastrial},
and then reducing routines to trials.
Nevertheless, a routine is not a trial,
because a semantic element is not a semantic set, or
because a mechanism implementing a routine
is not an adapter implementing a trial,
see comment to Lemma~\ref{adapterisunionofmechanisms}.

\comment{Comment}
This theorem is true if the Turing's thesis is true,
 see \ref{Turingsthesis}.
Conversely, if this theorem is true, then
`what is effectively calculable to resolve problems is computable'.

\clause{Remark}\label{evolutionofresolvers}
Provided
that a bigger range means more survival opportunities,
that software is cheaper than hardware,
that the adapter, the perceiver, the learner,
 and the subject conditions
 are satisfied in some environments, and
that in each step the increasing of complexity
 was overcome by its fitness,
then an evolution of resolvers
---mechanism to adapter to perceiver to learner to subject---
should follow.

\comment{Comment}
Although depending on conditions,
see Lemmas \ref{adaptercondition}, \ref{perceiverrange},
\ref{learnerrange}, and \ref{subjectrange},
the evolution of resolvers is directed,
and its final singularity is the
Turing complete subject $\Re_4[{\frak u}]$.

\comment{Comment}
In detail, the strictest evolution of resolvers is:
% $\Re_0[s]$,
% $\Re_1[S \cap \Sigma_\pi]$ (with $s \in S$),
% $\Re_2[i](S \cap \Sigma_\pi)$,
% $\Re_3[R](S \Cap \Sigma_\pi)$ (with $i \in R$), and
% $\Re_4[{\frak u}](R)(S \Cap \Sigma_\pi)$.
$
\Xi\Re_0[s]
 \mathrel{\smash{\numeq{\{\!s\!\}\subset S}\subset}}
\Xi\Re_1[S\cap\Sigma_\pi]
 \mathrel{\smash{\numeq{S\subset S'\!\!}\subset}}
\Xi\Re_2[i](S'\!\cap\Sigma_\pi)
 \numeq{\{\!i\!\}\subset R}\subseteq
\Xi\Re_3[R](S'\!\Cap\Sigma_\pi)
 \numeq{R\subset R'\!\!}\subseteq
\Xi\Re_4[{\frak u}](R')(S'\!\Cap\Sigma_\pi)$.

\clause{Thesis}\label{weareresolvers}
We are the result of an evolution of resolvers
of the survival problem.

\comment{Argument}
The resolvers hierarchy suggests an
evolution of resolvers of the survival problem,
 see \ref{evolutionofresolvers}.
And lacking of better explanations,
that we are Turing complete resolvers,
that is, subjects $\Re_4[{\frak u}] = {\cal U}$,
 see \ref{Turingcomplete},
suggests that we are indeed the
result of an evolution of resolvers
of the survival problem.

\comment{Comment}
Our species is Turing complete.
Therefore we must explain the evolution of Turing completeness.

\section{Conclusion}

\subsection{Purpose}

The problem theory is the union of
set theory and computing theory.
The integration of the two theories
is achieved by using a new vocabulary to refer to old concepts,
but mainly by giving the old theories a purpose that they
did not have: to resolve problems.
For example,
a set defined by intension is named a problem, and
the same set defined by extension is named its set of solutions.
While both still refer to the same set,
as it is the case in set theory,
the status of each of them is now very different:
one is a question and the other is an answer.
And when the problem theory states that
computing is resolving,
it is calling a set resolvable if it is recursively enumerable,
but mainly
it is saying that the transition from intension to extension
has to be calculated, because it
is not written magically in ``The Book''; % Paul Erd\H{o}s
someone has to write it.

The purpose of resolving problems is not final,
but the main conclusion of the paper,
the Thesis~\ref{weareresolvers},
is nearly ultimate:
We are Turing complete subjects because
we are the result of an evolution of resolvers
of the survival problem.
In other words,
we resolve problems to survive.
So, if survival is indeed the ultimate purpose,
then the problem theory provides purpose
and meaning to set theory and to computing theory.

The final Thesis~\ref{weareresolvers} also closes a loop,
because a Turing complete resolver $\Re_4[{\frak u}]$
can model {\it everything}, and then 
{\it everything\/} can be a solution,
as it is stated in Theorem \ref{everythingisin}.
But those {\it everythings\/} are not absolute,
but limited to what is computable, see \ref{threelimits}.
That is, if Turing's thesis stands, see \ref{Turingsthesis},
then {\it everything\/} is everything that is computable.
This way a restriction of computing theory, countability,
is inherited by problem theory and transferred to set theory;
see the details below in Subsection~\ref{Countability}.
The other question that requires some more
elaboration is the status of the Turing's thesis itself,
which we will postpone until Subsection~\ref{Intuition}.

Nevertheless, besides that main Thesis~\ref{weareresolvers},
the problem theory concepts presented in this paper
can be used to model, understand, and classify
both natural and artificial resolvers,
because the paper provides definitions, theorems, and taxonomies
for resolvers, and also for problems.
And, by the way, the paper defines
 adaptation, perception, and learning,
and it shows that there are just three ways to resolve
any problem: routine, trial, and analogy.

\subsection{Countability}

In computing everything is countable,
see \ref{computingiscounting},
and the problem theory in Turing universes
inherited countability from computing theory,
see \ref{resolvingiscounting}.
In a Turing universe, see \ref{Turinguniverse},
the limits of calculation are the limits of computing,
and then there are only computable functions,
 including predicates, see \ref{computable},
and computable sets, see \ref{recursivelyenumerable}.
Then the problem theory in Turing universes
is consistent if and only if computing is consistent.
And computing is consistent,
as a corollary to Church-Rosser theorem in $\lambda$-calculus,
 see \cite Curry \& Feys (1958) Chapter 4.

Therefore, our way to control paradoxes
 in set theory, and then in this paper,
is to confine ourselves to Turing universes.
But don't worry; if this is a Turing universe,
as it seems to be,
then we are only excluding imaginary universes.

For example,
the mathematical theorem that 
states that everything is a solution is proved,
and it makes sense, see~\ref{everythingisin}.
But it also causes paradoxes,
because from it we derive ${\bb P} \subset {\bb S}$, but
${\bb P} \numeq1= ({\bb S}\to{\bb B}) \numeq2= 2^{\bb S}$, and then
$|{\bb P}| \numeq3= |2^{\bb S}| \numeq4= 2^{|\bb S|} > |{\bb S}|$,
by Cantor's theorem.
It is not a paradox in a Turing universe
 because the forth equality is false in it.
The second equality is false in a Turing universe because,
as we saw in Lemma~\ref{cannotexpress},
there are resolvable problems that are not expressible,
so $({\bb S}\to{\bb B}) \T\subset 2^{\bb S}$.
The third equality is true,
though it follows the second one!
And the forth equality is false in a Turing universe
because the number of computable sets is countable,
so, if $|{\bb S}^*| = {\aleph_0}$,
then $|2^{\bb S}|^* = {\aleph_0} < 2^{\aleph_0} = 2^{|{\bb S}^*|}$,
that is, $|2^{\bb S}| \T< 2^{|{\bb S}|}$.
Therefore, ${\bb P}^*\! \subset {\bb S}^*$ is true, but
${\bb P}^*$ is the set of computable predicates,
that is, ${\bb P}^* = {\cal E}$ of \ref{problemsets}, and
$[2^{\bb S}]^* \subset {\bb S}^*$ is also true, but
$[2^{\bb S}]^*$ is the set of computable sets,
that is, $[2^{\bb S}]^* = {\cal R}$ of \ref{problemsets}.
The conclusion is that ${\bb S}^*$, the set of solutions,
is the set of everything that is computable.

We have just rejected 
the uncountable case, where
$|{\bb P}| > {\aleph_0}$,
but there are two other possibilities:
the (infinite) countable case, where
$|{\bb S}^*| = |{\bb P}^*| = |{\bb R}^*| = {\aleph_0}$,
see \ref{resolvingiscounting};
and the finite case, where
$|{\bb S}_{\Gamma}| < |{\bb P}_{\Gamma}| <
 |{\bb R}_{\Gamma}| < {\aleph_0}$,
see \ref{iffiniteresolvable}.

We are finite, so it would be natural
to restrict our investigations to the finite case,
calling for finiteness instead of
calling for countableness.
But the finite case is trivial,
and more importantly,
the difference between an unrestricted universal computer
and a finite universal computer is
not qualitative but quantitative.
There is not any step of any calculation
that an unrestricted universal computer can compute
and a finite universal computer cannot compute,
see \ref{proviso}.
So in the limit, that is,
without time nor memory restrictions,
we are universal computers.
And note that those restrictions are variable,
and that they can be relaxed nearly as desired
just spending some more time,
 or building a faster computer machine,
or using some more external memory.
In the case of a Turing machine,
the external memory is the tape,
and the internal memory is where
its processor keeps the internal state,
see \ref{Turingmachine}.
Note also that we can 
code a program to generate every natural number,
although we cannot follow the computation till its end.
Summarizing:
we are better defined saying that
we are qualitatively universal Turing machines,
but with some unspecified quantitative limitations,
than saying that we are
qualitatively finite state automata,
because finite state automata are not expandable.

Finally,
the rejections of finiteness and uncountableness imply that 
countableness is the golden mean.
This is Pythagorean heaven revisited,
{\it everything is countable},
but this time we have rescued the terrifying $\sqrt 2$,
and other irrational numbers.
As Kronecker said:
``God made counting numbers; all else is the work of man''.

\subsection{Intuition}

Is it possible to resolve a non-computable problem?
A problem is computable if, by definition,
see \ref{recursivelyenumerable}
and \ref{resolvableisrecursivelyenumerable},
a Turing machine can execute a valid resolution of the problem,
so the non-computable problem would not be resolved by computing,
but by other means.
My answer to the question is `no', because I think that
a problem is resolvable if, and only if,
the problem is computable,
see \ref{resolvingiscomputing}.

Nevertheless you may think otherwise, and say that
there is another way of resolving,
let us call it `intuition', that is not computable.
If that were the case, then
the problem theory with its mathematical formulation,
as presented in this paper,
would capture the concept of `computable problem',
but not the whole concept of `problem'.
In order to see this,
please consider the following two statements:
\begingroup\everypar{}\parindent=20pt\parskip=0pt
 \item{$\circ$} Some problems are computable.
 \item{$\circ$} A universal computer can execute
   any computable resolution.
\par\noindent\endgroup
Even if you believe that
there are resolvable problems that are not computable,
you can still decide easily that both are true;
the first is a fact, and the second is a theorem.
And then everything in this paper
would still be true of computable problems,
computable resolutions, and computable solutions.

The key point in this discussion is that
`intuition' would refute Turing's thesis,
see \ref{Turingsthesis},
because if there were `intuitive' resolutions, then
we could effectively calculate what is not computable.
Turing's thesis is not a theorem, and
we follow \cite Post (1936)
in considering Turing's thesis to be
a law of nature that states a limitation
of our own species calculating capacity,
see \cite Casares (2016a),
by which we are bound to
see ourselves as a final singularity.
Summarizing:
If Turing's thesis were eventually false, then
this problem theory would be about computable problems.
But, while Turing's thesis remains valid,
the problem theory is about problems,
the set of effectively calculable functions is countable
 (\ref{Turingsthesis}),
universal computers are the most capable computing devices
 (\ref{maximumcomputing}),
everything is an expression
 (\ref{allisexpression}),
resolving is computing
 (\ref{resolvingiscomputing}),
and the problem theory is complete
 (\ref{problemtheoryiscomplete}).

\vfill\break
\xsection{References}
\nobreak\medskip

\reference Cantor (1895):
Georg Cantor,
``Contributions to the Founding of the Theory of
  Transfinite Numbers (First Article)'';
in \periodical{Mathematische Annalen},
vol.\ xlvi, pp.\ 481--512, 1895.
Translated by P.E.B.\ Jourdain in
\book{Contributions to the Founding of the Theory of
  Transfinite Numbers},
Dover, New York, 1955.
\ISBN 978-0-486-60045-1.

\reference Casares (2016a):
Ram\'on Casares,
``Proof of Church's Thesis'';
\URL{\tt arXiv:1209.5036}<http://arxiv.org/abs/1209.5036>.

\reference Casares (2016b):
Ram\'on Casares,
``Syntax Evolution: Problems and Recursion'';\\
\URL{\tt arXiv:1508.03040}<http://arxiv.org/abs/arXiv:1508.03040>.

\reference Curry \& Feys (1958):
Haskell B.\ Curry, and
Robert Feys, with
William  Craig,
\book{Combinatory Logic}, Vol.\ I.
North-Holland Publishing Company,
Amsterdam, 1958.
\ISBN 978-0-7204-2207-8.

\reference Gandy (1980):
Robin Gandy,
``Church's Thesis and Principles for Mechanisms'';
\DOI{10.1016/s0049-237x(08)71257-6}.
In \book{The Kleene Symposium}
(editors: J.\ Barwise, H.J.\ Keisler \& K.\ Kunen),
Volume 101 of
Studies in Logic and the Foundations of Mathematics;
North-Holland, Amsterdam, 1980, pp.~123--148;
\ISBN 978-0-444-55719-3.

\reference Kleene (1936):
Stephen C.\ Kleene,
``$\lambda$-Definability and Recursiveness''; 
in \periodical{Duke Mathematical Journal},
vol.\ 2, pp.\ 340--353, 1936,
\DOI{10.1215/s0012-7094-36-00227-2}.

\reference Lettvin et al. (1959):
Jerome Y.\ Lettvin, Humberto R.\ Maturana,
Warren S.\ McCulloch, and Walter H.\ Pitts,
``What the Frog's Eye Tells the Frog's Brain'';
in \periodical{Proceedings of the IRE},
vol.\ 47, no.\ 11, pp.\ 1940--1951, November 1959,\\
\DOI{10.1109/jrproc.1959.287207}.

\reference Post (1936):
Emil L.\ Post,
``Finite Combinatory Processes --- Formulation 1'';
in \periodical{The Journal of Symbolic Logic},
vol.\ 1, no.\ 3, pp.\ 103--105, September 1936,\\
\DOI{10.2307/2269031}.
Received October 7, 1936.

\reference Post (1944):
Emil L.\ Post,
``Recursively Enumerable Sets of Positive Integers
  and their Decision Problems'';
in \periodical{Bulletin of the American Mathematical Society},
vol.\ 50, no.~5, pp.\ 284--316, 1944,
\DOI{10.1090/s0002-9904-1944-08111-1}.

\reference Turing (1936):
Alan M.\ Turing,
``On Computable Numbers,
 with an Application to the Entscheidungsproblem'';
in \periodical{Proceedings of the London Mathematical Society},
vol.\ s2-42, issue 1, pp.\ 230--265, 1937,
\DOI{10.1112/plms/s2-42.1.230}.
Received 28 May, 1936. Read 12 November, 1936.

\reference Turing (1937):
Alan M.\ Turing,
``Computability and $\lambda$-Definability''; in
\periodical{The Journal of Symbolic Logic},
vol.\ 2, no.\ 4, pp.\ 153--163, December 1937,
\DOI{10.2307/2268280}.
Received September 11, 1937.

\reference Turing (1938):
Alan M.\ Turing,
``Systems of Logic Based on Ordinals'';
Princeton University PhD dissertation.
Submitted 17 May, 1938. Oral examination 31 May, 1938.
Printed in \periodical{Proceedings of the London Mathematical Society},
vol.\ s2-45, issue 1, pp.\ 161--228, 1939,
\DOI{10.1112/plms/s2-45.1.161}.
Received 31 May, 1938. Read 16 June, 1938.

\bye